\documentclass[table]{article}
\usepackage{anyfontsize}
\usepackage{tutorial}

\newcommand{\hf}{\raisebox{.28em}{\hspace{.05em}\includegraphics[height=1em]{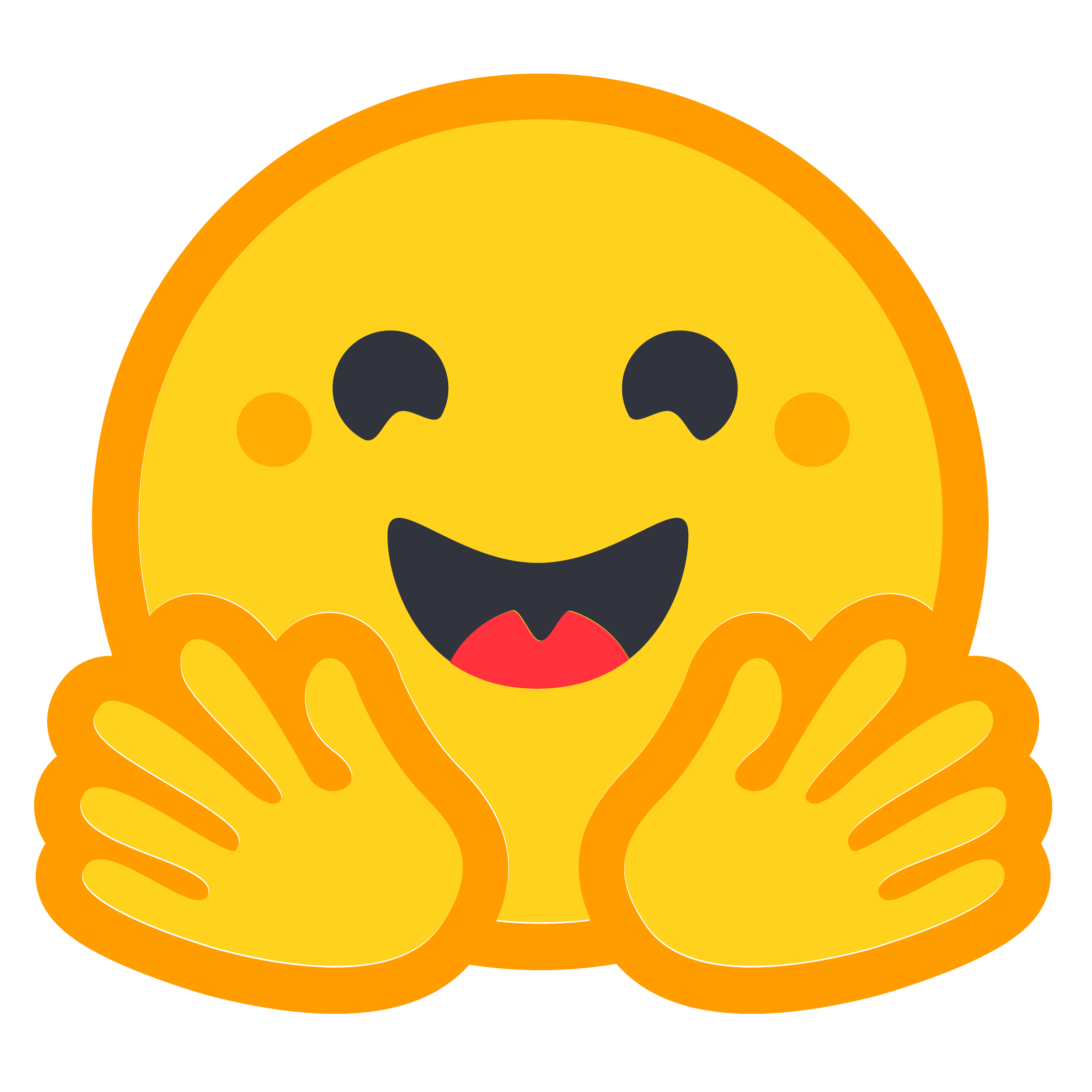}}\xspace}

\newcommand{\oxford}{\raisebox{.3em}{\hspace{.05em}\includegraphics[height=1em]{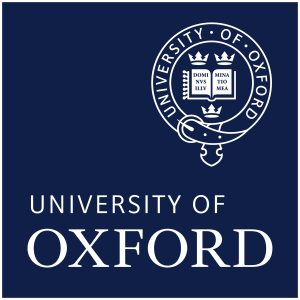}}\xspace}

\title{
Robot Learning: A Tutorial
}

\authorOne[]{Francesco Capuano \oxford \hf}
\authorOne[]{Caroline Pascal\hf}
\authorOne[]{Adil Zouitine\hf}
\authorOne[]{Thomas Wolf\hf}
\authorOne[]{Michel Aractingi\hf}

\contribution[]{\oxford University of Oxford, \hf Hugging Face}

\abstract{
Robot learning is at an inflection point, driven by rapid advancements in machine learning and the growing availability of large-scale robotics data. 
This shift from classical, model-based methods to data-driven, learning-based paradigms is unlocking unprecedented capabilities in autonomous systems. 
This tutorial navigates the landscape of modern robot learning, charting a course from the foundational principles of Reinforcement Learning and Behavioral Cloning to generalist, language-conditioned models capable of operating across diverse tasks and even robot embodiments.
This work is intended as a guide for researchers and practitioners, and our goal is to equip the reader with the conceptual understanding and practical tools necessary to contribute to developments in robot learning, with ready-to-use examples implemented in~\lerobot.
\newline

Code: \textbf{\url{https://github.com/huggingface/lerobot}}
\newline
Date: \textbf{\today}
}

\begin{document}

\maketitle

\tableofcontents
\section*{Foreword}

Robotics is an inherently multidisciplinary field, which is witnessing unprecedented advancements since its inception in the 1960s.
Yet, more than sixty years after the debut of Unimate, robots have still not fully integrated into the rich, unstructured, and dynamic world we humans inhabit.
Over the decades, numerous disciplines have shown immense promise in tackling the challenges of creating autonomous robotic systems.
This tutorial takes a clear stance in the debate on whether modern Machine 
Learning can play a pivotal role in the development of 
autonomous robots: we believe this to be the case.

Nonetheless, we also hold that the wealth of research from both academia and industry in classical robotics over the past six decades is, simply put, too valuable to be cast aside in favor of purely learning-based methods.
However, the interplay between classical robotics and modern machine learning is still in its nascent stages, and the path to integration yet to be clearly defined.
In turn our goal here is to present what we consider to be the most relevant approaches within robot learning today, while warmly extending an invite to collaborate to expand the breadth of this work! Start contributing today \href{https://github.com/fracapuano/robot-learning-tutorial}{here}.

This tutorial\dots
\begin{itemize}
    \item Does \emph{not} aim to be a comprehensive guide to general field of robotics, manipulation or underactuated systems:~\citet{sicilianoSpringerHandbookRobotics2016} and~\citet{tedrakeRoboticManipulationPerception,tedrakeUnderactuatedRoboticsAlgorithms} do this better than we ever could.
    \item Does \emph{not} aim to be an introduction to statistical or deep learning:~\citet{shalev-shwartzUnderstandingMachineLearning2014} and~\citet{prince2023understanding} cover these subjects better than we ever could.
    \item Does \emph{not} aim to be a deep dive into Reinforcement Learning, Diffusion Models, or Flow Matching: invaluable works such as~\citet{suttonReinforcementLearningIntroduction2018},~\citet{nakkiranStepbyStepDiffusionElementary2024}, and~\citet{lipmanFlowMatchingGuide2024} do this better than we ever could.
\end{itemize}

Instead, our goal here is to provide an intuitive explanation as per why these disparate ideas have converged to form the exciting field of modern robot learning, driving the unprecedented progress we see today. 
In this spirit, we follow the adage: "a jack of all trades is a master of none, \emph{but oftentimes better than a master of one}."

We sincerely hope this tutorial serves as a valuable starting point for your journey into robot learning.

\newpage
\section{Introduction}

\begin{figure}
    \centering
    \includegraphics[width=\linewidth]{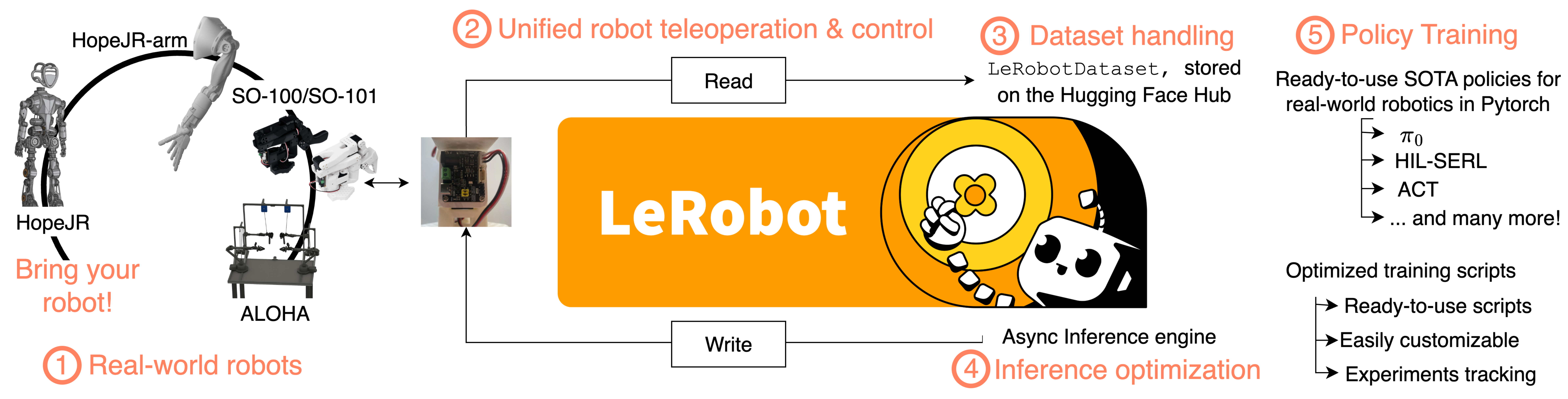}
    \caption{\lerobot~is the open-source library for end-to-end robotics developed by Hugging Face. The library is vertically integrated on the entire robotics stack, supporting low-level control of real-world robot devices, advanced data and inference optimizations, as well as  SOTA robot learning methods with simple implementations in pure Pytorch.}
    \label{fig:figure1}
\end{figure}

Autonomous robotics holds the premise of relieving humans from repetitive, tiring or dangerous manual tasks.
Consequently, the field of robotics has been widely studied since its first inception in the 1950s.
Lately, advancements in Machine Learning (ML) have sparked the development of a relatively new class of methods used to tackle robotics problems, leveraging large amounts of data and computation rather than human expertise and modeling skills to develop autonomous systems.

The frontier of robotics research is indeed increasingly moving away from classical model-based control paradigm, embracing the advancements made in ML, aiming to unlock (1) monolithic perception-to-action control pipelines and (2) multi-modal data-driven feature extraction strategies, together with (3) reduced reliance on precise models of the world and (4) a better positioning to benefit from the growing availability of open robotics data.
While central problems in manipulation, locomotion and whole-body control demand knowledge of rigid-body dynamics, contact modeling, planning under uncertainty, recent results seem to indicate learning can prove just as effective as explicit modeling, sparking interest in the field of \emph{robot learning}.
This interest can be largely justified considering the significant challenges related to deriving accurate models of robot-environment interactions.

Moreover, since end-to-end learning on ever-growing collections of text and image data has historically been at the core of the development of \emph{foundation models} capable of semantic reasoning across multiple modalities (images, text, audio, etc.), deriving robotics methods grounded in learning appears particularly consequential, especially as the number of openly available datasets continues to grow.

Robotics is, at its core, an inherently multidisciplinary field, requiring a wide range of expertise in both \emph{software} and \emph{hardware}.
The integration of learning-based techniques further broadens this spectrum of skills, raising the bar for both research and practical applications.
\lerobot~is an open-source library designed to integrate end-to-end with the entire robotics stack.
With a strong focus on accessible, real-world robots \highlight{(1) \lerobot~supports many, openly available, robotic platforms} for manipulation, locomotion and even whole-body control.
\lerobot also implements a \highlight{(2) unified, low-level approach to reading/writing robot configurations} to extend support for other robot platforms with relatively low effort. 
The library introduces \lerobotdataset, \highlight{(3) a native robotics dataset's format} currently being used by the community to efficiently record and share datasets.
\lerobot~also supports many state-of-the-art (SOTA) algorithms in robot learning---mainly based on Reinforcement Learning (RL) and Behavioral Cloning (BC) techniques---with efficient implementations in Pytorch, and extended support to experimentation and experiments tracking.
Lastly, \lerobot~defines a custom, optimized inference stack for robotic policies decoupling action planning from action execution, proving effective in guaranteeing more adaptability at runtime.

This tutorial serves the double purpose of providing useful references for the Science behind---and practical use of---common robot learning techniques.
To this aim, we strike to provide a rigorous yet concise overview of the core concepts behind the techniques presented, paired with practical examples of how to use such techniques concretely, with code examples in \lerobot, for researchers and practitioners interested in the field of robot learning.
This tutorial is structured as follows:
\begin{itemize}
\item Section~\ref{sec:classical} reviews classical robotics foundations, introducing the limitations of dynamics-based approaches to robotics.
\item Section~\ref{sec:learning-rl} elaborates on the limitations of dynamics-based methods, and introduce RL as a practical approach to solve robotics problems, considering its upsides and potential limitations.
\item Section~\ref{sec:learning-imitation} further describes robot learning techniques that aim at solving single-tasks learning, leveraging BC techniques to autonomously reproduce specific expert demonstrations.
\item Section~\ref{sec:learning-foundation} presents recent contributions on developing generalist models for robotics applications, by learning from large corpora of multi-task \& multi-robot data (\emph{robotics foundation models}).
\end{itemize}

Our goal with this tutorial is to provide an intuitive explanation of the reasons various disparate ideas from Machine Learning (ML) have converged and are powering the current evolution of Robotics, driving the unprecedented progress we see today.
We complement our presentation of the most common and recent approaches in robot learning with practical code implementations using \lerobot, and start here by presenting the dataset format introduced with \lerobot.

\subsection{\lerobotdataset}

\lerobotdataset~is one of the most impactful features of \lerobot, developed in keeping with the observation that robotics data is increasingly central in robot learning. 
Thus, \lerobot~defines a standardized dataset format designed to address the specific needs of robot learning research, providing a unified and convenient access to robotics data across modalities, including sensorimotor readings, multiple camera feeds and teleoperation status.
\lerobotdataset~also accommodates for storing general information regarding the data being collected, including textual descriptions of the task being performed by the teleoperator, the kind of robot used, and relevant measurement specifics like the frames per second at which the recording of both image and robot state's streams are proceeding.

In this, \lerobotdataset~provides a unified interface for handling multi-modal, time-series data, and it is designed to seamlessly integrate with the PyTorch and Hugging Face ecosystems.
\lerobotdataset~can be easily extended by users and it is highly customizable by users, and it already supports openly available data coming from a variety of embodiments supported in \lerobot, ranging from manipulator platforms like the SO-100 arm and ALOHA-2 setup, to real-world humanoid arm and hands, as well as entirely simulation-based datasets, and self-driving cars.
This dataset format is built to be both efficient for training and flexible enough to accommodate the diverse data types encountered in robotics, while promoting reproducibility and ease of use for users. 

\subsubsection{The dataset class design}

A core design choice behind \lerobotdataset~is separating the underlying data storage from the user-facing API.
This allows for efficient storage while presenting the data in an intuitive, ready-to-use format.

Datasets are always organized into three main components:
\begin{itemize}
\item \textbf{Tabular Data}: Low-dimensional, high-frequency data such as joint states, and actions are stored in efficient memory-mapped files, and typically offloaded to the more mature \texttt{datasets} library by Hugging Face, providing fast with limited memory consumption.
\item \textbf{Visual Data}: To handle large volumes of camera data, frames are concatenated and encoded into MP4 files. Frames from the same episode are always grouped together into the same video, and multiple videos are grouped together by camera. To reduce stress on the file system, groups of videos for the same camera view are also broke into multiple sub-directories, after a given threshold number.
\item \textbf{Metadata} A collection of JSON files which describes the dataset's structure in terms of its metadata, serving as the relational counterpart to both the tabular and visual dimensions of data. Metadata include the different feature schema, frame rates, normalization statistics, and episode boundaries.
\end{itemize}

For scalability, and to support datasets with potentially millions of trajectories (resulting in hundreds of millions or billions of individual camera frames), we merge data from different episodes into the same high-level structure.
Concretely, this means that any given tabular collection and video will not typically contain information about one episode only, but rather a concatenation of the information available in multiple episodes.
This keeps the pressure on the file system limited, both locally and on remote storage providers like Hugging Face, though at the expense of leveraging more heavily relational-like, metadata parts of the dataset, which are used to reconstruct information such as at which position, in a given file, an episode starts or ends.
An example struture for a given \lerobotdataset~would appear as follows:
\begin{itemize}
\item \texttt{meta/info.json}: This metadata is a central metadata file. It contains the complete dataset schema, defining all features (e.g., \texttt{observation.state}, \texttt{action}), their shapes, and data types. It also stores crucial information like the dataset's frames-per-second (\texttt{fps}), \lerobot's version at the time of capture, and the path templates used to locate data and video files.
\item \texttt{meta/stats.json}: This file stores aggregated statistics (mean, std, min, max) for each feature across the entire dataset, used for data normalization for most policy models and accessible externally via \texttt{dataset.meta.stats}.
\item \texttt{meta/tasks.jsonl}: This file contains the mapping from natural language task descriptions to integer task indices, which are useful for task-conditioned policy training.
\item \texttt{meta/episodes/*} This directory contains metadata about each individual episode, such as its length, the corresponding task, and pointers to where its data is stored in the dataset's files. For scalability, this information is stored in files rather than a single large JSON file.
\item \texttt{data/*}: Contains the core frame-by-frame tabular data, using parquet files to allow for fast, memory-mapped access. To improve performance and handle large datasets, data from multiple episodes are concatenated into larger files. These files are organized into chunked subdirectories to keep the size of directories manageable. A single file typically contains data for more than one single episode.
\item \texttt{videos/*}: Contains the MP4 video files for all visual observation streams. Similar to the \texttt{data/} directory, the video footage from multiple episodes is concatenated into single MP4 files. This strategy significantly reduces the number of files in the dataset, which is more efficient for modern filesystems.
\end{itemize}

\subsection{Code Example: Batching a (Streaming) Dataset}

This section provides an overview of how to access datasets hosted on Hugging Face using the \lerobotdataset~class.
Every dataset on the Hugging Face Hub containing the three main pillars presented above (Tabular, Visual and relational Metadata), and can be assessed with a single instruction.

In practice, most reinforcement learning (RL) and behavioral cloning (BC) algorithms tend to operate on stack of observation and actions.
For the sake of brevity, we will refer to joint spaces, and camera frames with the single term of \emph{frame}.
For instance, RL algorithms may use a history of previous frames \(o_{t-H_o:t} \) to mitigate partial observability, and BC algorithms are in practice trained to regress chunks of multiple actions (\(a_{t+t+H_a} \)) rather than single controls.
To accommodate for these specifics of robot learning training, \lerobotdataset~provides a native windowing operation, whereby users can define the \emph{seconds} of a given window (before and after) around any given frame, by using the \texttt{delta\_timestemps} functionality.
Unavailable frames are opportunely padded, and a padding mask is also returned to filter out the padded frames.
Notably, this all happens within the \lerobotdataset, and is entirely transparent to higher level wrappers commonly used in training ML models such as \texttt{torch.utils.data.DataLoader}.

Conveniently, by using \lerobotdataset~with a Pytorch \texttt{DataLoader} one can automatically collate the individual sample dictionaries from the dataset into a single dictionary of batched tensors for downstream training or inference.
\lerobotdataset~also natively supports streaming mode for datasets.
Users can stream data of a large dataset hosted on the Hugging Face Hub, with a one-line change in their implementation.
Streaming datasets supports high-performance batch processing (ca. 80-100 it/s, varying on connectivity) and high levels of frames randomization, key features for practical BC algorithms which otherwise may be slow or operating on highly non-i.i.d. data.
This feature is designed to improve on accessibility so that large datasets can be processed by users without requiring large amounts of memory and storage.

\begin{pbox}[label={ex:dataset-batching}]{Batching a (Streaming) Dataset \\ \url{https://github.com/fracapuano/robot-learning-tutorial/blob/main/snippets/ch1/01_datasets.py}}
\begin{lstlisting}[language=python]
import torch
from lerobot.datasets.lerobot_dataset import LeRobotDataset
from lerobot.datasets.streaming_dataset import StreamingLeRobotDataset

delta_timestamps = {
    # 0.2, and 0.1 seconds *before* each frame
    "observation.images.wrist_camera": [-0.2, -0.1, 0.0]
}

# Optionally, use StreamingLeRobotDataset to avoid downloading the dataset
dataset = LeRobotDataset(
    "lerobot/svla_so101_pickplace",
    delta_timestamps=delta_timestamps
)

# Streams frames from the Hugging Face Hub without loading into memory
streaming_dataset = StreamingLeRobotDataset(
    "lerobot/svla_so101_pickplace",
    delta_timestamps=delta_timestamps
)

# Get the 100th frame in the dataset by 
sample = dataset[100]
print(sample)
# {
# 'observation.state': tensor([...]), 
# 'action': tensor([...]), 
# 'observation.images.wrist_camera': tensor([3, C, H, W]), for delta timesteps
# ...
# }

batch_size=16
# wrap the dataset in a DataLoader to use process it batches for training purposes
data_loader = torch.utils.data.DataLoader(
    dataset,
    batch_size=batch_size
)

# Iterate over the DataLoader in a training loop
num_epochs = 1
device = "cuda" if torch.cuda.is_available() else "cpu"

for epoch in range(num_epochs):
    for batch in data_loader:
        # Move data to the appropriate device (e.g., GPU)
        observations = batch["observation.state"].to(device)
        actions = batch["action"].to(device)
        images = batch["observation.images.wrist_camera"].to(device)

        # Next, you can do amazing_model.forward(batch)
        ...
\end{lstlisting}
\end{pbox}

\subsection{Code Example: Collecting Data}
\label{paragraph:collecting-data}

\begin{pbox}[label={ex:record-dataset}]{Record a Dataset \\ \url{https://github.com/fracapuano/robot-learning-tutorial/blob/main/snippets/ch1/02_record_data.py}}
\begin{lstlisting}[language=python]
"""
You can also use the CLI to record data. To see the required arguments, run:
lerobot-record --help
"""
from lerobot.cameras.opencv.configuration_opencv import OpenCVCameraConfig
from lerobot.datasets.lerobot_dataset import LeRobotDataset
from lerobot.datasets.utils import hw_to_dataset_features
from lerobot.robots.so100_follower import SO100Follower, SO100FollowerConfig
from lerobot.teleoperators.so100_leader.config_so100_leader import SO100LeaderConfig
from lerobot.teleoperators.so100_leader.so100_leader import SO100Leader
from lerobot.utils.control_utils import init_keyboard_listener
from lerobot.utils.utils import log_say
from lerobot.utils.visualization_utils import init_rerun
from lerobot.scripts.lerobot_record import record_loop

NUM_EPISODES = 5
FPS = 30
EPISODE_TIME_SEC = 60
RESET_TIME_SEC = 10
TASK_DESCRIPTION = ...  # provide a task description

HF_USER = ...  # provide your Hugging Face username

follower_port = ...  # find your ports running: lerobot-find-port
leader_port = ...
follower_id = ...  # to load the calibration file
leader_id = ...

# Create the robot and teleoperator configurations
camera_config = {"front": OpenCVCameraConfig(
    index_or_path=0, width=640, height=480, fps=FPS)
}
robot_config = SO100FollowerConfig(
    port=follower_port,
    id=follower_id,
    cameras=camera_config
)
teleop_config = SO100LeaderConfig(
    port=leader_port, 
id=leader_id
)

# Initialize the robot and teleoperator
robot = SO100Follower(robot_config)
teleop = SO100Leader(teleop_config)

# Configure the dataset features
action_features = hw_to_dataset_features(robot.action_features, "action")
obs_features = hw_to_dataset_features(robot.observation_features, "observation")
dataset_features = {**action_features, **obs_features}

# Create the dataset where to store the data
dataset = LeRobotDataset.create(
    repo_id=f"{HF_USER}/robot-learning-tutorial-data",
    fps=FPS,
    features=dataset_features,
    robot_type=robot.name,
    use_videos=True,
    image_writer_threads=4,
)

# Initialize the keyboard listener and rerun visualization
_, events = init_keyboard_listener()
init_rerun(session_name="recording")

# Connect the robot and teleoperator
robot.connect()
teleop.connect()

episode_idx = 0
while episode_idx < NUM_EPISODES and not events["stop_recording"]:
    log_say(f"Recording episode {episode_idx + 1} of {NUM_EPISODES}")

    record_loop(
        robot=robot,
        events=events,
        fps=FPS,
        teleop=teleop,
        dataset=dataset,
        control_time_s=EPISODE_TIME_SEC,
        single_task=TASK_DESCRIPTION,
        display_data=True,
    )

    # Reset the environment if not stopping or re-recording
    if (not events["stop_recording"]) and \
        (episode_idx < NUM_EPISODES - 1 or events["rerecord_episode"]):
        log_say("Reset the environment")
        record_loop(
            robot=robot,
            events=events,
            fps=FPS,
            teleop=teleop,
            control_time_s=RESET_TIME_SEC,
            single_task=TASK_DESCRIPTION,
            display_data=True,
        )

    if events["rerecord_episode"]:
        log_say("Re-recording episode")
        events["rerecord_episode"] = False
        events["exit_early"] = False
        dataset.clear_episode_buffer()
        continue

    dataset.save_episode()
    episode_idx += 1

# Clean up
log_say("Stop recording")
robot.disconnect()
teleop.disconnect()
dataset.push_to_hub()
\end{lstlisting}
\end{pbox}

\newpage
\section{Classical Robotics}
\label{sec:classical}

\epigraph{\textit{Know your enemy} [...]}{Sun Tzu}

\begin{tldr}
Learning-based approaches to robotics are motivated by the need to (1) generalize across tasks and embodiments (2) reduce dependency on human expertise (3) leverage historical trends on the production of data---all traditionally overlooked by dynamics-based techniques.
\end{tldr}

\subsection{Explicit and Implicit Models}

\begin{figure}
    \centering
    \includegraphics[width=0.5\linewidth]{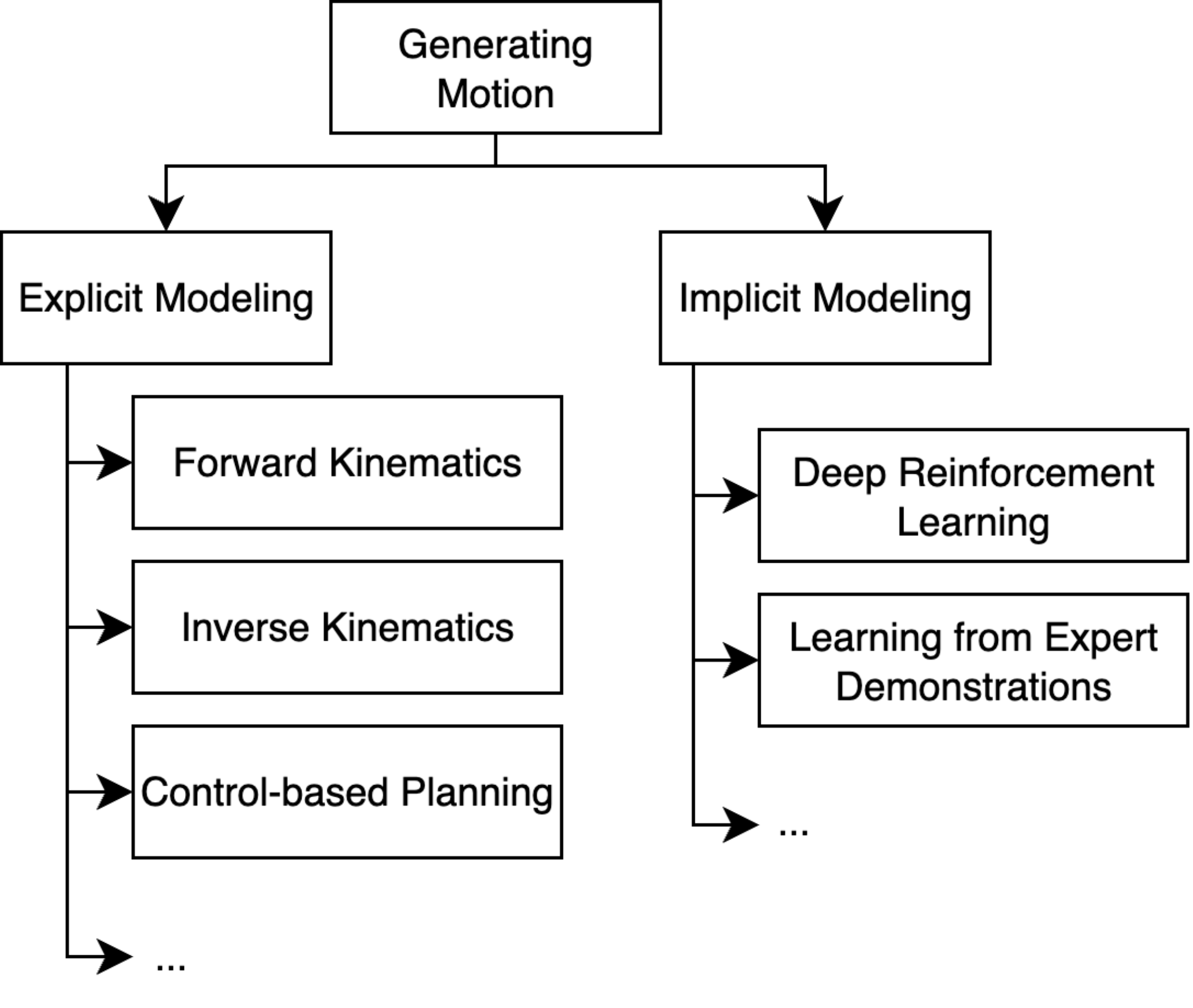}
    \caption{Overview of methods to generate motion (clearly non-exhausitve, see~\citet{bekrisStateRobotMotion2024}). The different methods can be grouped based on whether they explicitly (\emph{dynamics-based}) or implicitly (\emph{learning-based}) model robot-environment interactions.}
    \label{fig:generating-motion-atlas}
\end{figure}

Robotics is concerned with producing artificial motion in the physical world in useful, reliable and safe fashion.
Thus, robotics is an inherently multi-disciplinar domain: producing autonomous motion in the physical world requires, to the very least, interfacing different software (motion planners) and hardware (motion executioners) components.
Further, knowledge of mechanical, electrical, and software engineering, as well as rigid-body mechanics and control theory have therefore proven quintessential in robotics since the field first developed in the 1950s.
More recently, Machine Learning (ML) has also proved effective in robotics, complementing these more traditional disciplines~\citep{connellRobotLearning1993}.
As a direct consequence of its multi-disciplinar nature, robotics has developed as a rather wide array of methods, all concerned with the main purpose of \highlight{producing artificial motion in the physical world}.

Methods to produce robotics motion range from traditional \emph{explicit} models---\highlight{dynamics-based}\footnote{In here, we refer to both \emph{kinematics} and \emph{dynamics}-based control.} methods, leveraging precise descriptions of the mechanics of robots' rigid bodies and their interactions with eventual obstacles in the environment---to \emph{implicit} models---\highlight{learning-based} methods, treating artificial motion as a statistical pattern to learn given multiple sensorimotor readings~\citep{agrawalComputationalSensorimotorLearning,bekrisStateRobotMotion2024}.
A variety of methods have been developed between these two extrema.
For instance, ~\citet{hansenTemporalDifferenceLearning2022} show how learning-based systems can benefit from information on the physics of problems, complementing a traditional learning method such as Temporal Difference (TD)-learning~\citet{suttonReinforcementLearningIntroduction2018} with Model-Predictive Control (MPC).
Conversely, as explicit models may be relying on assumptions proving overly simplistic---or even unrealistic---in practice, learning can prove effective to improve modeling of complex phenomena or complement perception~\citep{mccormacSemanticFusionDense3D2016}.
Such examples aim at demonstrating the richness of approaches to robotics, and Figure~\ref{fig:generating-motion-atlas} graphically illustrates some of the most relevant techniques.
Such a list is clearly far from being exhaustive, and we refer to~\citet{bekrisStateRobotMotion2024} for a more comprehensive overview of both general and application-specific methods for motion generation.
In this section, we wish to introduce the inherent benefits of \highlight{learning-based approaches to robotics}---the core focus on this tutorial.

\subsection{Different Types of Motion}

\begin{figure}
    \centering
    \includegraphics[width=0.7\linewidth]{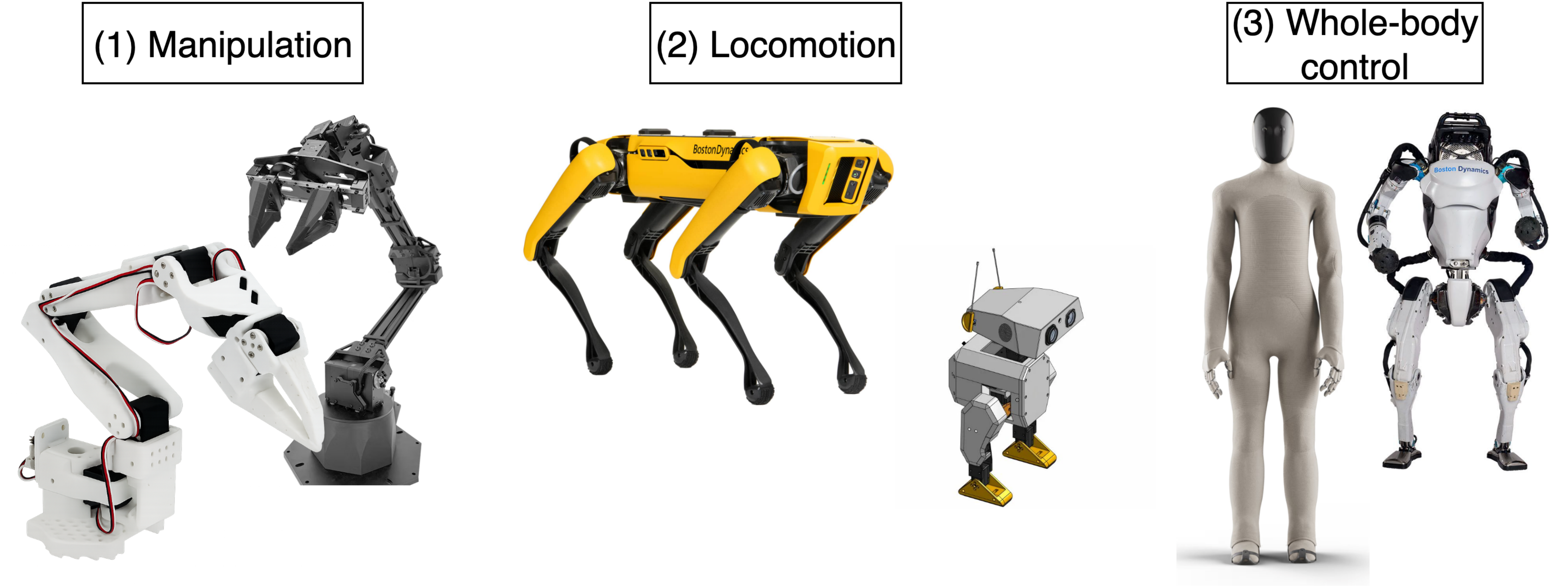}
    \caption{Different kinds of motions are achieved with potentially very different robotic platforms. From left to right, top to bottom: ViperX, SO-100, Boston Dynamics' Spot, Open-Duck, 1X's NEO, Boston Dynamics' Atlas. This is an example list of robotic platforms and is (very) far from being exhaustive.}
    \label{fig:robotics-platforms-atlas}
\end{figure}

In the vast majority of instances, robotics deals with producing motion via actuating joints connecting nearly entirely-rigid links.
A key distinction between focus areas in robotics is based on whether the generated motion modifies (1) the absolute state of the environment (via dexterity), (2) the relative state of the robot with respect to its environment (exercising mobility skills), or (3) a combination of the two (Figure~\ref{fig:robotics-platforms-atlas}).

Effects such as (1) are typically achieved \emph{through} the robot, i.e. generating motion to perform an action inducing a desirable modification, effectively \emph{manipulating} the environment (manipulation).
Motions like (2) may result in changes in the robot's physical location within its environment.
Generally, modifications to a robot's location within its environment may be considered instances of the general \emph{locomotion} problem, further specified as \emph{wheeled} or \emph{legged} locomotion based on whenever a robot makes use of wheels or leg(s) to move in the environment.
Lastly, an increased level of dynamism in the robot-environment interactions can be obtained combining (1) and (2), thus designing systems capable to interact with \emph{and} move within their environment.
This category is problems is typically termed \emph{mobile manipulation}, and is characterized by a typically much larger set of control variables compared to either locomotion or manipulation alone.

The traditional body of work developed since the very inception of robotics is increasingly complemented by learning-based approaches.
ML has indeed proven particularly transformative across the entire robotics stack, first empowering planning-based techniques with improved state estimation used for traditional planning~\citep{tangPerceptionNavigationAutonomous2023} and then end-to-end replacing controllers, effectively yielding perception-to-action methods~\citep{koberReinforcementLearningRobotics}.
Work in producing robots capable of navigating a diverse set of terrains demonstrated the premise of both dynamics and learning-based approaches for locomotion~\citep{griffinWalkingStabilizationUsing2017,jiDribbleBotDynamicLegged2023,leeLearningQuadrupedalLocomotion2020,margolisRapidLocomotionReinforcement2022}, and recent works on whole-body control indicated the premise of learning-based approaches to generate rich motion on complex robots, including humanoids~\citep{zhangWoCoCoLearningWholeBody2024,bjorckGR00TN1Open2025}.
Manipulation has also been widely studied, particularly considering its relevance for many impactful use-cases ranging from high-risk applications for humans~\citep{fujitaDevelopmentRobotsNuclear2020,alizadehComprehensiveSurveySpace2024} to manufacturing~\citep{sannemanStateIndustrialRobotics2020}.
While explicit models have proven fundamental in achieving important milestones towards the development of modern robotics, recent works leveraging implicit models proved particularly promising in surpassing scalability and applicability challenges via learning~\citep{koberReinforcementLearningRobotics}.

\subsection{Example: Planar Manipulation}
Robot manipulators typically consist of a series of links and joints, articulated in a chain finally connected to an \emph{end-effector}.
Actuated joints are considered responsible for generating motion of the links, while the end effector is instead used to perform specific actions at the target location (e.g., grasping/releasing objects via closing/opening a gripper end-effector, using a specialized tool like a screwdriver, etc.).

\begin{figure}
    \centering
    \includegraphics[width=0.4\linewidth]{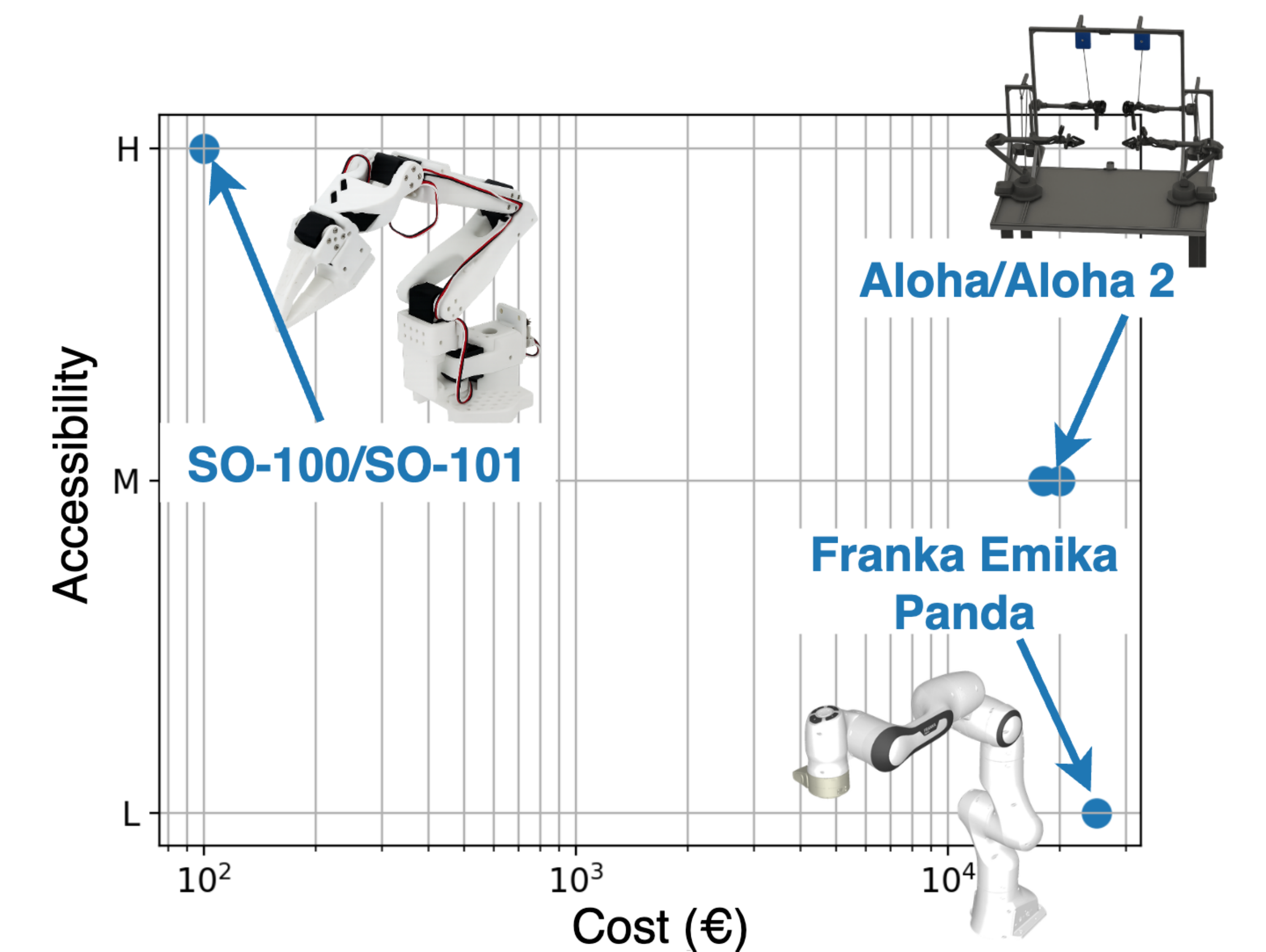}
    \caption{Cheaper, more accessible robots are starting to rival traditional platforms like the Panda arm platforms in adoption in resource-constrained scenarios. The SO-100, in particular, has a cost in the 100s of Euros, and can be entirely 3D-printed in hours, while the industrially-manufactured Panda arm costs tens of thousands of Euros and is not openly available.}
    \label{fig:robotic-platforms-costs}
\end{figure}

Recently, the development of low-cost manipulators like the ALOHA~\citep{zhaoLearningFineGrainedBimanual2023} ALOHA-2~\citep{aldacoALOHA2Enhanced} and SO-100/SO-101~\citep{knightStandardOpenSO100} platforms significantly lowered the barrier to entry to robotics, considering the increased accessibility of these robots compared to more traditional platforms like the Franka Emika Panda arm (Figure~\ref{fig:robotic-platforms-costs}).

Deriving an intuition as per why learning-based approaches are gaining popularity in the robotics community requires briefly analyzing traditional approaches for manipulation, leveraging tools like forward and inverse kinematics (FK, IK) and control theory.
Providing a detailed overview of these methods falls (well) out of the scope of this tutorial, and we refer the reader to works including~\citet{sicilianoSpringerHandbookRobotics2016, lynchModernRoboticsMechanics2017, tedrakeRoboticManipulationPerception, tedrakeUnderactuatedRoboticsAlgorithms} for a much more comprehensive description of these techniques.
Here, we mostly wish to highlight the benefits of ML over these traditional techniques

\begin{figure}
    \centering
    \includegraphics[width=0.7\linewidth]{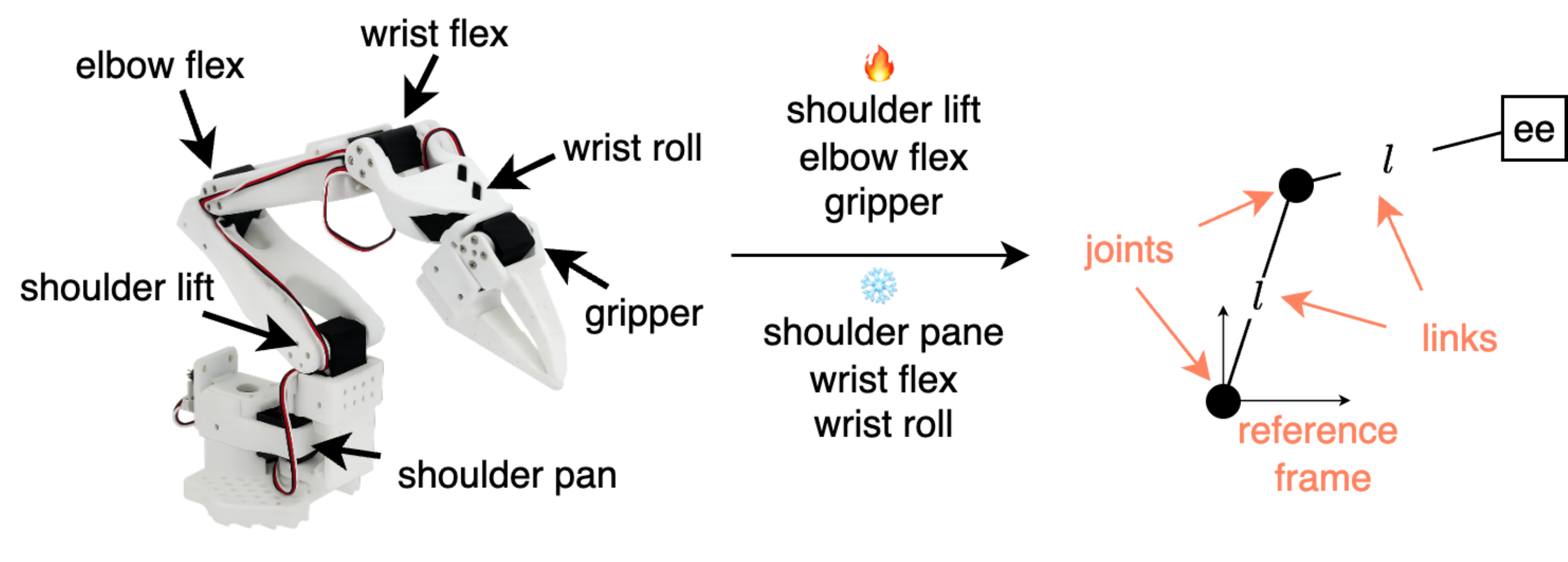}
    \caption{The SO-100 arm is a 6-dof manipulator arm. Preventing some of its joints (shoulder pane, wrist flex and wrist roll) from actuating, it can be represented as a traditional 2-dof planar manipulator (the gripper joint in the end-effector is not considered towards the count of the degrees of freedom used to produce motion).}
    \label{fig:make-so100-planar-manipulator}
\end{figure}

Consider the (simple) case where a SO-100 is restrained from actuating (1) the shoulder pane and (2) the wrist flex and roll motors.
This effectively reduces the degrees of freedom of the SO-100 from the original 5+1 (5 joints + 1 gripper) to 2+1 (shoulder lift, elbow flex + gripper).
As the end-effector does not impact motion in this model, the SO-100 is effectively reduced to the planar manipulator robot presented in Figure~\ref{fig:make-so100-planar-manipulator}, where spheres represent actuators, and solid lines indicate length-\(l\) links from the base of the SO-100 to the end-effector (\emph{ee}).

Further, let us make the simplifying assumption that actuators can produce rotations up to \( 2 \pi \) radians.
In practice, this is seldom the case due to movement obstructions caused by the robot body itself (for instance, the shoulder lift cannot produce counter-clockwise movement due to the presence of the robot's base used to secure the SO-100 to its support and host the robot bus), but we will introduce movement obstruction at a later stage.

All these simplifying assumptions leave us with the planar manipulator of Figure~\ref{fig:planar-manipulation-simple}, free of moving its end-effector by controlling the angles \( \theta_1 \) and \( \theta_2 \), jointly referred to as the robot's \emph{configuration}, and indicated with \( q = [\theta_1, \theta_2 ] \in [-\pi, +\pi]^2 \).
The axis attached to the joints indicate the associated reference frame, whereas circular arrows indicate the maximal feasible rotation allowed at each joint. 
In this tutorial, we do not cover topics related to spatial algebra, and we instead refer the reader to \citet[Chapter~2]{lynchModernRoboticsMechanics2017} and \citet[Chapter~3]{tedrakeRoboticManipulationPerception} for excellent explanations of the mechanics and theoretical foundations of producing motion on rigid bodies.

\newcommand{\panelheight}{3.2cm}  

\begin{figure}
    \centering
    \begin{subfigure}[t]{0.32\linewidth}
        \centering
        \includegraphics[width=\linewidth,height=\panelheight,keepaspectratio]{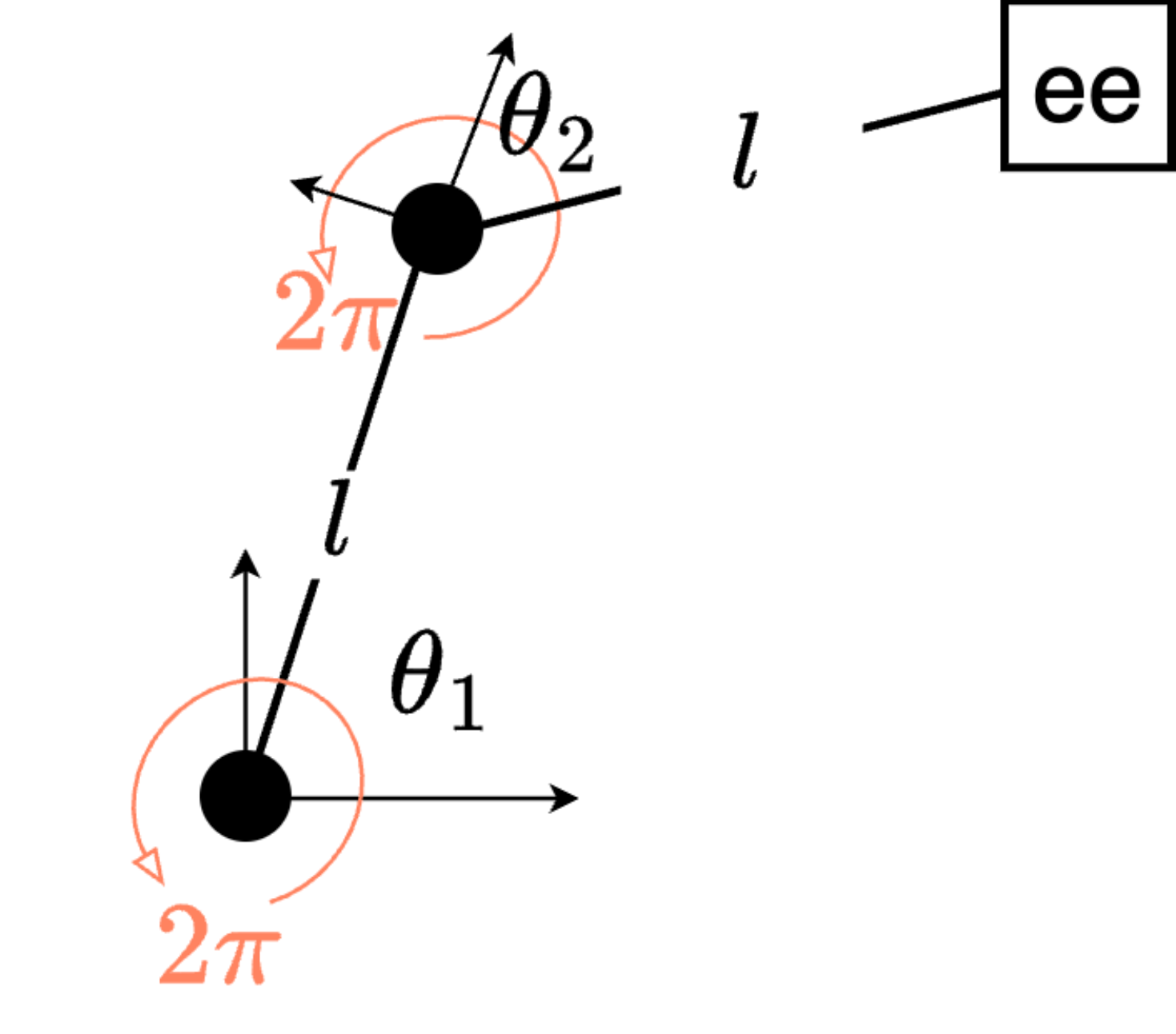}
        \caption{Free to move}
        \label{fig:planar-manipulation-simple}
    \end{subfigure}\hfill
    \begin{subfigure}[t]{0.32\linewidth}
        \centering
        \includegraphics[width=\linewidth,height=\panelheight,keepaspectratio]{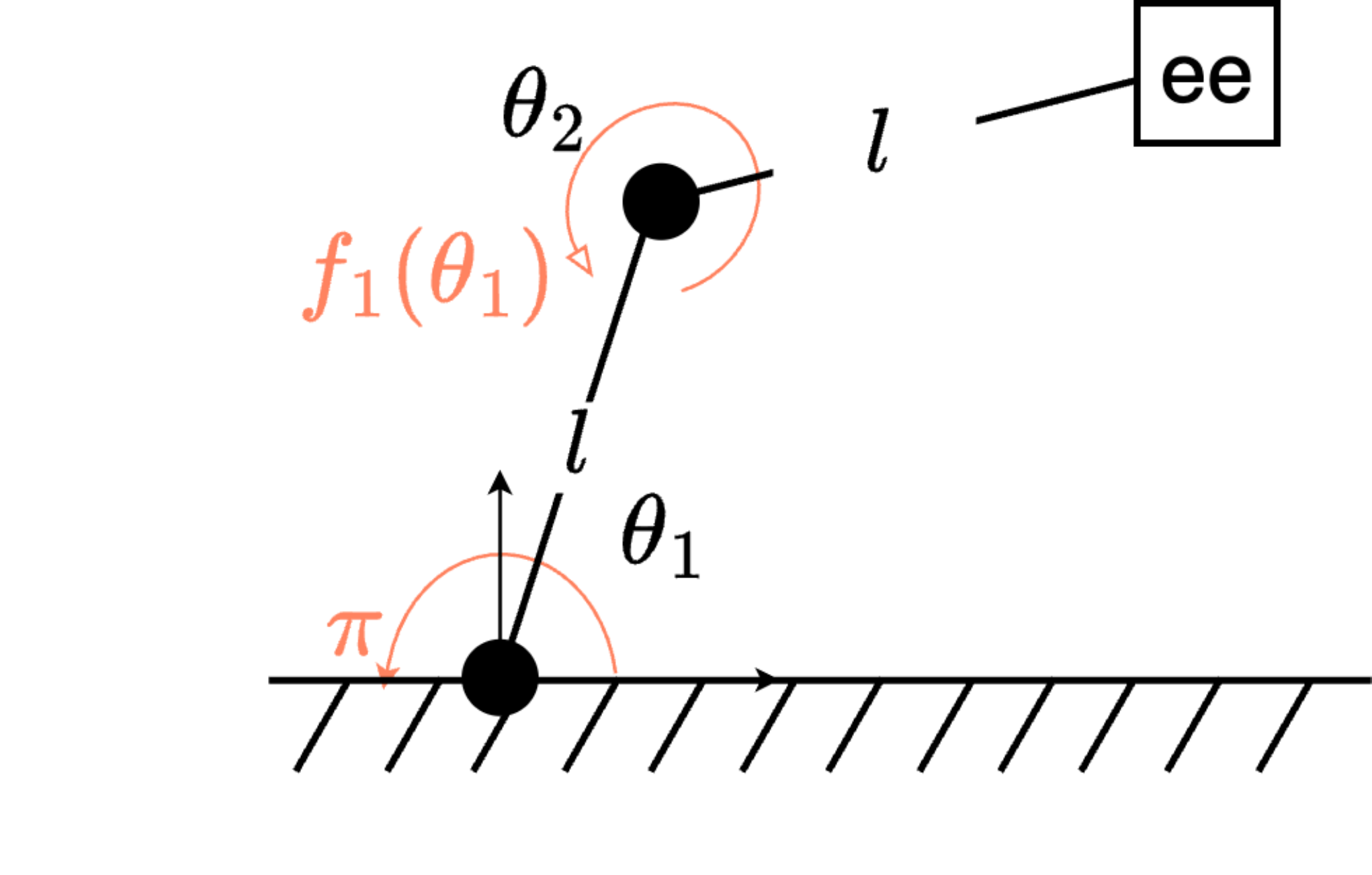}
        \caption{Constrained by the surface}
        \label{fig:planar-manipulator-floor}
    \end{subfigure}\hfill
    \begin{subfigure}[t]{0.32\linewidth}
        \centering
        \includegraphics[width=\linewidth,height=\panelheight,keepaspectratio]{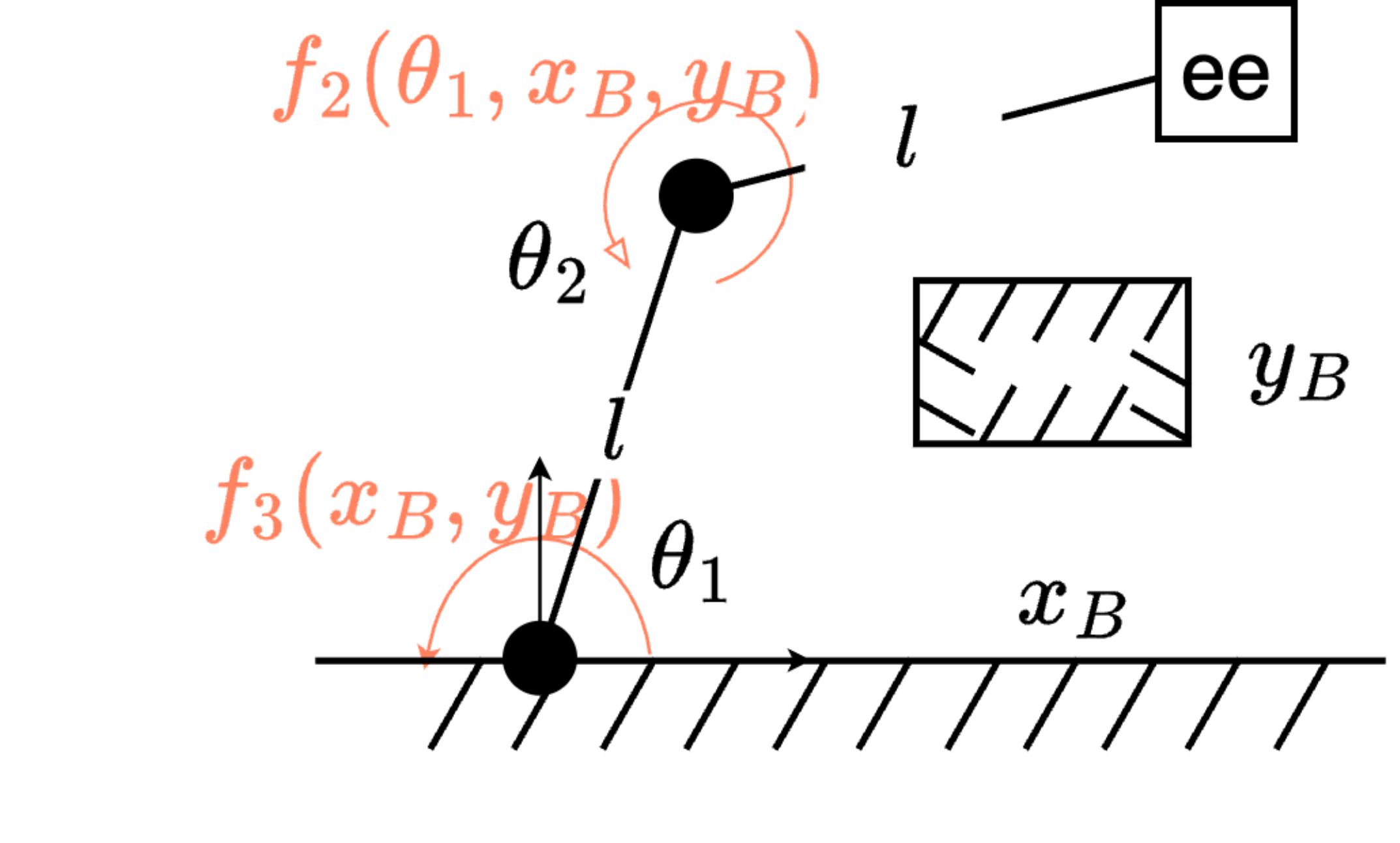}
        \caption{Constrained by surface and (fixed) obstacle}
        \label{fig:planar-manipulator-floor-shelf}
    \end{subfigure}
    \caption{Planar, 2-dof schematic representation of the SO-100 manipulator under diverse deployment settings. From left to right: completely free of moving; constrained by the presence of the surface; constrained by the surface and presence of obstacles. Circular arrows around each joint indicate the maximal rotation feasible at that joint.}
\end{figure}

Considering the (toy) example presented in Figure~\ref{fig:planar-manipulation-simple}, then we can analytically write the end-effector's position \( p \in \mathbb R^2 \) as a function of the robot's configuration, \( p = p(q), p: \mathcal Q \mapsto \mathbb R^2 \). 
In particular, we have:
\begin{equation*}
p(q) = 
\begin{pmatrix}
p_x(\theta_1, \theta_2) \\  
p_y(\theta_1, \theta_2)
\end{pmatrix}
=
\begin{pmatrix}
l \cos(\theta_1) + l \cos(\theta_1 + \theta_2) \\
l \sin(\theta_1) + l \sin(\theta_1 + \theta_2)
\end{pmatrix}
\in S^{n=2}_{l_1+l_2} = \{ p(q) \in \mathbb R^2: \Vert p(q) \Vert_2^2 \leq (2l)^2, \ \forall q \in \mathcal Q \}
\end{equation*}

Deriving the end-effector's \emph{pose}---position \emph{and} orientation---in some \(m\)-dimensional space \( \vec{p} \in \mathcal{P} \subset \mathbb{R}^{m} \) starting from the configuration \( \q \in \mathcal Q \subset \mathbb R^n \) of a \( n \)-joints robot is referred to as \emph{forward kinematics} (FK), whereas identifying the configuration corresponding to any given target pose is termed \emph{inverse kinematics} (IK).
In that, FK is used to map a robot configuration into the corresponding end-effector pose, whereas IK is used to reconstruct the configuration(s) given an end-effector pose.

In the simplified case here considered (for which \( \vec{p} \equiv p \), as the orientation of the end-effector is disregarded for simplicity), one can solve the problem of controlling the end-effector's location to reach a goal position \( p^* \) by solving analytically for \( q: p(q) = f_{\FK}(q) = p^*\).
However, in the general case, one might not be able to solve this problem analytically, and can typically resort to iterative optimization methods comparing candidate solutions using a loss function (in the simplest case, \( \Vert p(q) - p^* \Vert_2^2 \) is a natural candidate), yielding:

\begin{align}
\min_{q \in \mathcal Q} \Vert p(q) - p^* \Vert_2^2 \, .
\label{eq:ik_problem}
\end{align}

Exact analytical solutions to IK are even less appealing when one considers the presence of obstacles in the robot's workspace, resulting in constraints on the possible values of \( q \in \mathcal Q \subseteq [-\pi, +\pi]^n \subset \mathbb R^n \) in the general case of \(n\)-links robots.

For instance, the robot in Figure~\ref{fig:planar-manipulator-floor} is (very naturally) obstacled by the presence of the surface upon which it rests: \( \theta_1 \) can now exclusively vary within \([0,  \pi] \), while possible variations in \( \theta_2 \) depend on \( \theta_1 \) (when \( \theta_1 \to 0 \) or \( \theta_1 \to \pi \), further downwards movements are restricted).
Even for a simplified kinematic model, developing techniques to solve~eq.~\ref{eq:ik_problem} is in general non-trivial in the presence of constraints, particularly considering that the feasible set of solutions \( \mathcal Q \) may change across problems.
Figure~\ref{fig:planar-manipulator-floor-shelf} provides an example of how the environment influences the feasible set considered, with a new set of constraints deriving from the position of a new obstacle.

However, IK---solving eq.~\ref{eq:ik_problem} for a feasible \( q \)---only proves useful in determining information regarding the robot's configuration in the goal pose, and crucially does not provide information on the \emph{trajectory} to follow over time to reach a target pose.
Expert-defined trajectories obviate to this problem providing a length-\(K\) succession of goal poses \( \tau_K = [p^*_0, p^*_1, \dots p^*_K] \) for tracking.
In practice, trajectories can also be obtained automatically through \emph{motion planning} algorithms, thus avoiding expensive trajectory definition from human experts.
However, tracking \( \tau_K \) via IK can prove prohibitively expensive, as tracking would require \( K \) resolutions of eq.~\ref{eq:ik_problem} (one for each target pose).
\emph{Differential} inverse kinematics (diff-IK) complements IK via closed-form solution of a variant of eq.~\ref{eq:ik_problem}. 
Let \( J(q) \) denote the Jacobian matrix of (partial) derivatives of the FK-function \( f_\FK: \mathcal Q \mapsto \mathcal P \), such that \( J(q) = \frac{\partial f_{FK}(q)}{\partial q } \).
Then, one can apply the chain rule to any \( p(q) = f_{\FK}(q) \), deriving \( \dot p = J(q) \dot q \), and thus finally relating variations in the robot configurations to variations in pose, thereby providing a platform for control.

Given a desired end-effector trajectory \( \targetvel(t) \) (1) indicating anchor regions in space and (2) how much time to spend in each region, diff-IK finds \( \dot q(t) \) solving for joints' \emph{velocities} instead of \emph{configurations},
\begin{align}
\dot q(t) = \arg\min_\nu \; \lVert J(q(t)) \nu - \targetvel (t) \rVert_2^2
\label{eq:reg_ik_velocity}
\end{align}

Unlike~eq.~\ref{eq:ik_problem}, solving for \( \dot q \) is much less dependent on the environment (typically, variations in velocity are constrained by physical limits on the actuators).
Conveniently, eq.~\ref{eq:reg_ik_velocity} also often admits the closed-form solution \( \dot q = J(q)^+ \targetvel \), where \( J^+(q) \) denotes the Moore-Penrose pseudo-inverse of \( J(q) \).
Finally, discrete-time joint configurations \( q \) can be reconstructed from joint velocities \( \dot q \) using forward-integration on the continuous-time joint velocity , \( q_{t+1} = q_t + \Delta t\,\dot q_t \) for a given \( \Delta t \), resulting in tracking via diff-IK.

Following trajectories with diff-IK is a valid option in well-controlled and static environments (e.g., industrial manipulators in controlled manufacturing settings), and relies on the ability to define a set of target velocities to track \( [\targetvel_0, \targetvel_1, \dots, \targetvel_k ] \)---an error-prone task largely requiring human expertise.
Furthermore, diff-IK relies on the ability to (1) access \( J(q) \, \forall q \in \mathcal Q \) and (2) compute its pseudo-inverse at every iteration of a given control cycle---a challenging assumption in highly dynamical settings, or for complex kinematic chains.

\subsubsection{Adding Feedback Loops}
While very effective when a goal trajectory has been well specified, the performance of diff-IK can degrade significantly in the presence of modeling/tracking errors, or in the presence of non-modeled dynamics in the environment.

\begin{wrapfigure}[12]{r}{0.3\textwidth}
    \vspace{-\intextsep}
    \centering
    \includegraphics[width=\linewidth]{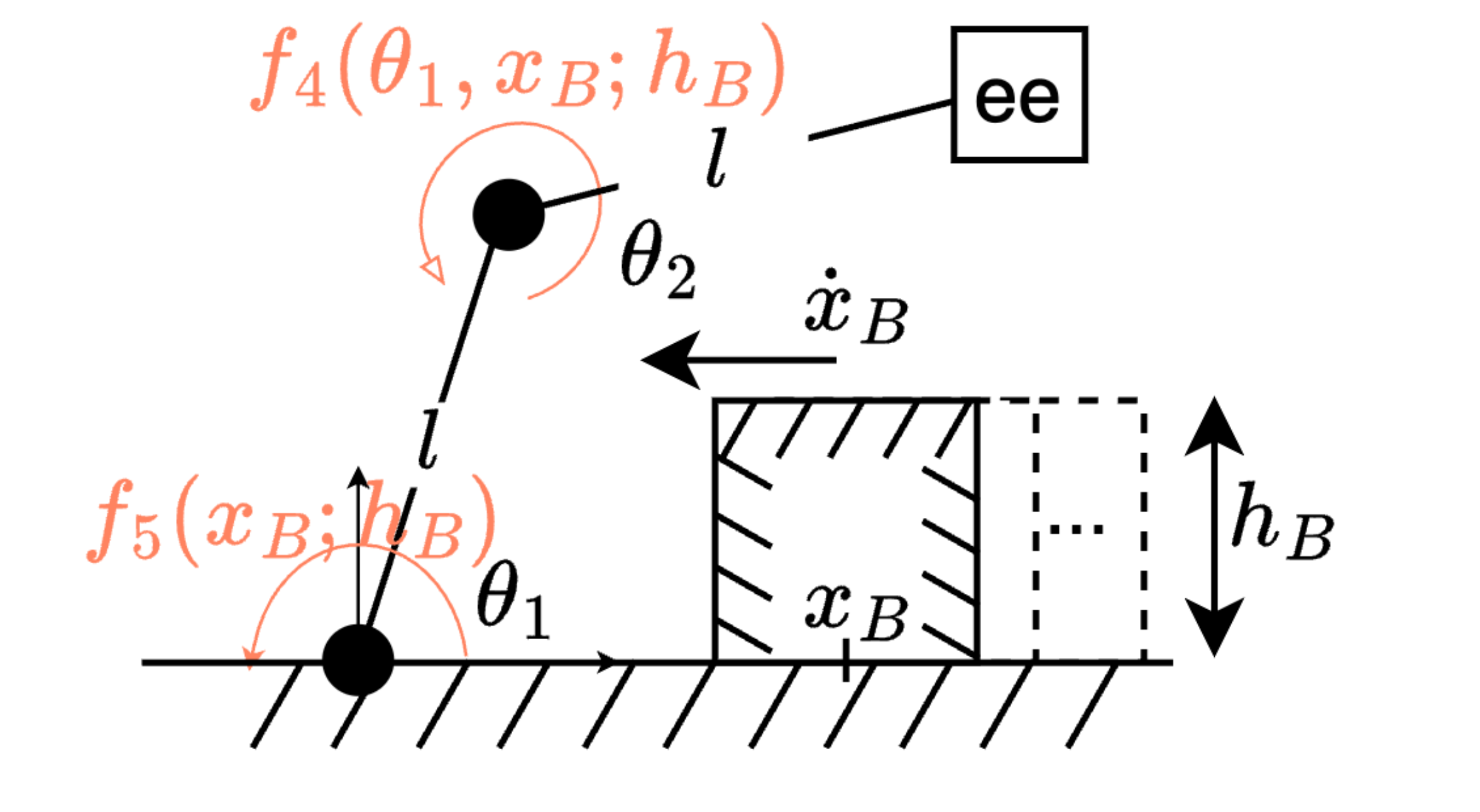}
    \caption{Planar manipulator robot in the presence of a moving obstacle.}
    \label{fig:planar-manipulator-box-velocity}
\end{wrapfigure}

One such case is presented in Figure~\ref{fig:planar-manipulator-box-velocity}, where another rigid body other than the manipulator is moving in the environment along the horizontal axis, with velocity \( \dot x_B \).
Accounting analytically for the presence of this disturbance---for instance, to prevent the midpoint of the link from ever colliding with the object---requires access to \( \dot x_B \) at least, to derive the equation characterizing the motion of the environment.

Less predictable disturbances however (e.g., \( \dot x_B \leftarrow \dot x_B + \eps, \eps \sim N(0,1) \)) may prove challenging to model analytically, and one could attain the same result of preventing link-object collision by adding a condition on the distance between the midpoint of \( l \) and \( x_B \), enforced through a feedback loop on the position of the robot and object at each control cycle.

To mitigate the effect of modeling errors, sensing noise and other disturbances, classical pipelines indeed do augment diff-IK with feedback control looping back quantities of interest.
In practice, following a trajectory with a closed feedback loop might consist in backwarding the error between the target and measured pose, \( \Delta p = \targetpos - p(q) \), hereby modifying the control applied to \( \dot q = J(q)^+ (\targetvel + k_p \Delta p ) \), with \( k_p \) defined as the (proportional) gain.

More advanced techniques for control consisting in feedback linearization, PID control, Linear Quatratic Regulator (LQR) or Model-Predictive Control (MPC) can be employed to stabilize tracking and reject moderate perturbations, and we refer to \citet[Chapter~8]{sicilianoSpringerHandbookRobotics2016} for in-detail explanation of these concepts, or \citep[Chapter~8]{tedrakeRoboticManipulationPerception} for a simple, intuitive example in the case of a point-mass system.
Nonetheless, feedback control presents its challenges as well: tuning gains remains laborious and system-specific. 
Further, manipulation tasks present intermittent contacts inducing hybrid dynamics (mode switches) and discontinuities in the Jacobian, challenging the stability guarantees of the controller and thus often necessitating rather conservative gains and substantial hand-tuning.

We point the interested reader to~\citet[Chapter~2,7,8]{sicilianoSpringerHandbookRobotics2016}, \citet[Chapter~6,11]{lynchModernRoboticsMechanics2017}, and~\citet[Chapter~3,8]{tedrakeRoboticManipulationPerception} for extended coverage of FK, IK, diff-IK and control for (diff-)IK.

\subsection{Limitations of Dynamics-based Robotics}
Despite the last 60+ years of robotics research, autonomous robots are still largely incapable of performing tasks at human-level performance in the physical world generalizing across (1) robot embodiments (different manipulators, different locomotion platforms, etc.) and (2) tasks (tying shoe-laces, manipulating a diverse set of objects).
While essential in the early development of robotics, the aforementioned methods require significant human expertise to be used in practice, and are typically specific to a particular applicative problem.

\begin{figure}
    \centering
    \includegraphics[width=0.9\linewidth]{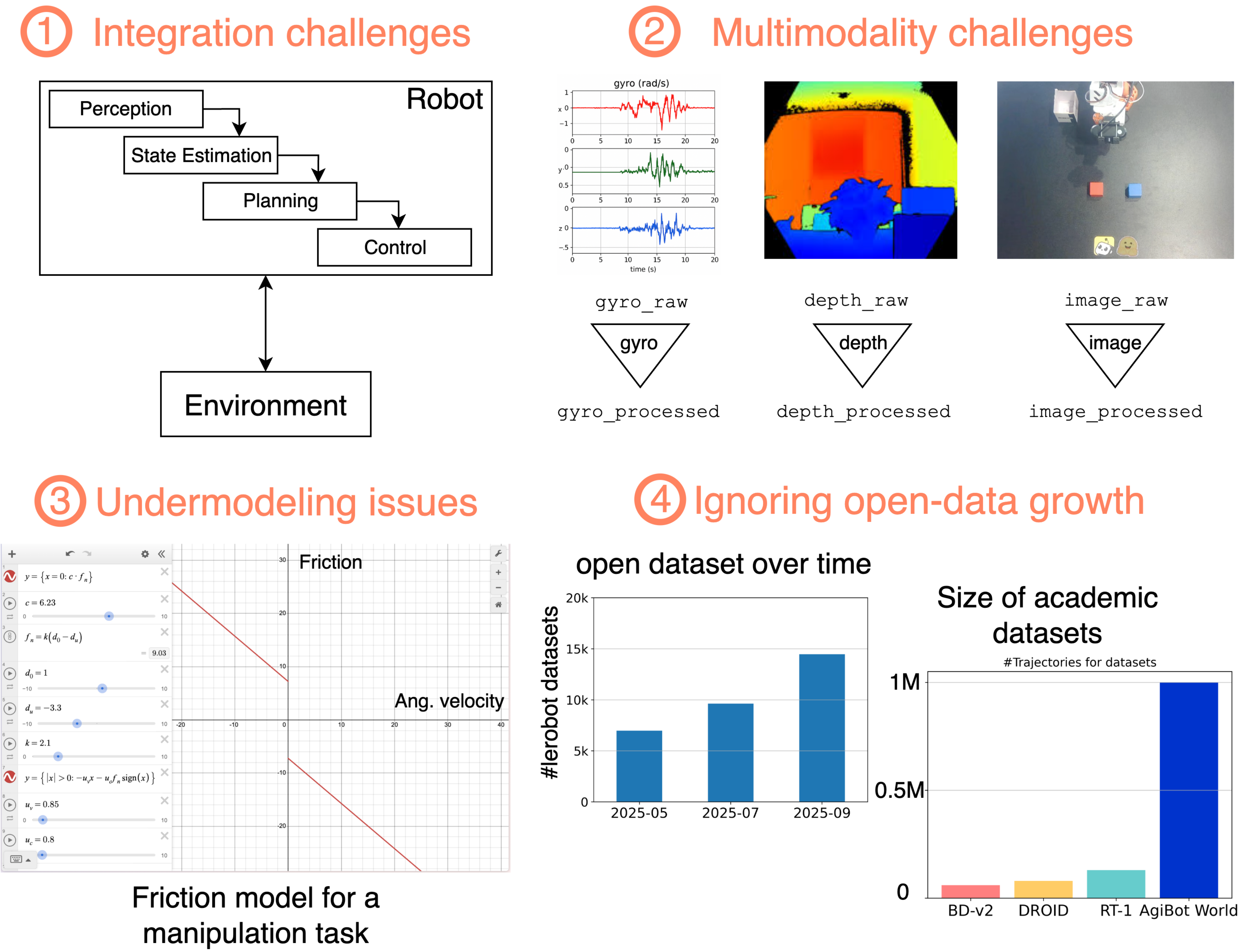}
    \caption{Dynamics-based approaches to robotics suffer from several limitations: (1) orchestrating multiple components poses integration challenges; (2) the need to develop custom processing pipelines for the sensing modalities and tasks considered hinders scalability; (3) simplified analytical models of physical phenomena (here friction at the gripper; credits to~\citet{antonovaReinforcementLearningPivoting2017}) limit real-world performance. Lastly, (4) dynamics-based methods overlook trends in the availability and growth of robotics data.}
    \label{fig:classical-limitations}
\end{figure}

Dynamics-based robotics pipelines have historically been \highlight{developed sequentially, engineering the different blocks} now within most architectures for specific purposes.
That is, sensing, state estimation, mapping, planning, (diff-)IK, and low-level control have been traditionally developed as distinct modules with fixed interfaces.
Pipelining these specific modules proved error-prone, and brittleness emerges---alongside compounding errors---whenever changes incur (e.g., changes in lighting for sensing, occlusion/failure of sensors, control failures).
Adapting such a stack to new tasks or robotic platforms often entails re-specifying objectives, constraints, and heuristics at multiple stages, incurring significant engineering overhead.

Moreover, classical planners operate on compact, assumed-sufficient state representations; extending them to reason directly over raw, heterogeneous and noisy data streams is non-trivial.
This results in a \highlight{limited scalability to multimodal data and multitask settings}, as incorporating high-dimensional perceptual inputs (RGB, depth, tactile, audio) traditionally required extensive engineering efforts to extract meaningful features for control. 
Also, the large number of tasks, coupled with the adoption of \emph{per-task} planners, goal parameterizations, and safety constraints, results in an explosion in design and validation options, with little opportunity to reuse solutions across tasks.

Setting aside integration and scalability challenges: developing accurate modeling of contact, friction, and compliance for complicated systems remains difficult.
Rigid-body approximations are often insufficient in the presence of deformable objects, and \highlight{relying on approximated models hinders real-world applicability} of the methods developed.
In the case of complex, time-dependent and/or non-linear dynamics, even moderate mismatches in parameters, unmodeled evolutions, or grasp-induced couplings can qualitatively affect the observed dynamics.

Lastly, dynamics-based methods (naturally) overlook the rather recent \highlight{increase in availability of openly-available robotics datasets}. 
The curation of academic datasets by large centralized groups of human experts in robotics~\citep{oneillOpenXEmbodimentRobotic2025, khazatskyDROIDLargeScaleInTheWild2025} is now increasingly complemented by a \highlight{growing number of robotics datasets contributed in a decentralized fashion} by individuals with varied expertise.
If not tangentially, dynamics-based approaches are not posed to maximally benefit from this trend, which holds the premise of allowing generalization in the space of tasks and embodiments, like data was the cornerstone for advancements in vision~\citep{alayracFlamingoVisualLanguage2022} and natural-language understanding~\citep{brownLanguageModelsAre2020}.

Taken together, these limitations (Figure~\ref{fig:classical-limitations}) motivate the exploration of learning-based approaches that can (1) integrate perception and control more tightly, (2) adapt across tasks and embodiments with reduced expert modeling interventions and (3) scale gracefully in performance as more robotics data becomes available.

\newpage
\section{Robot (Reinforcement) Learning}
\label{sec:learning-rl}

\epigraph{\textit{Approximate the solution, not the problem} [...]}{Richard Sutton}

\begin{tldr}
The need for expensive, high-fidelity simulators can be obviated learning from real-world data, using sample-efficient algorithms that can safely train directly on hardware.
\end{tldr}

\begin{figure}
    \centering
    \includegraphics[width=0.9\linewidth]{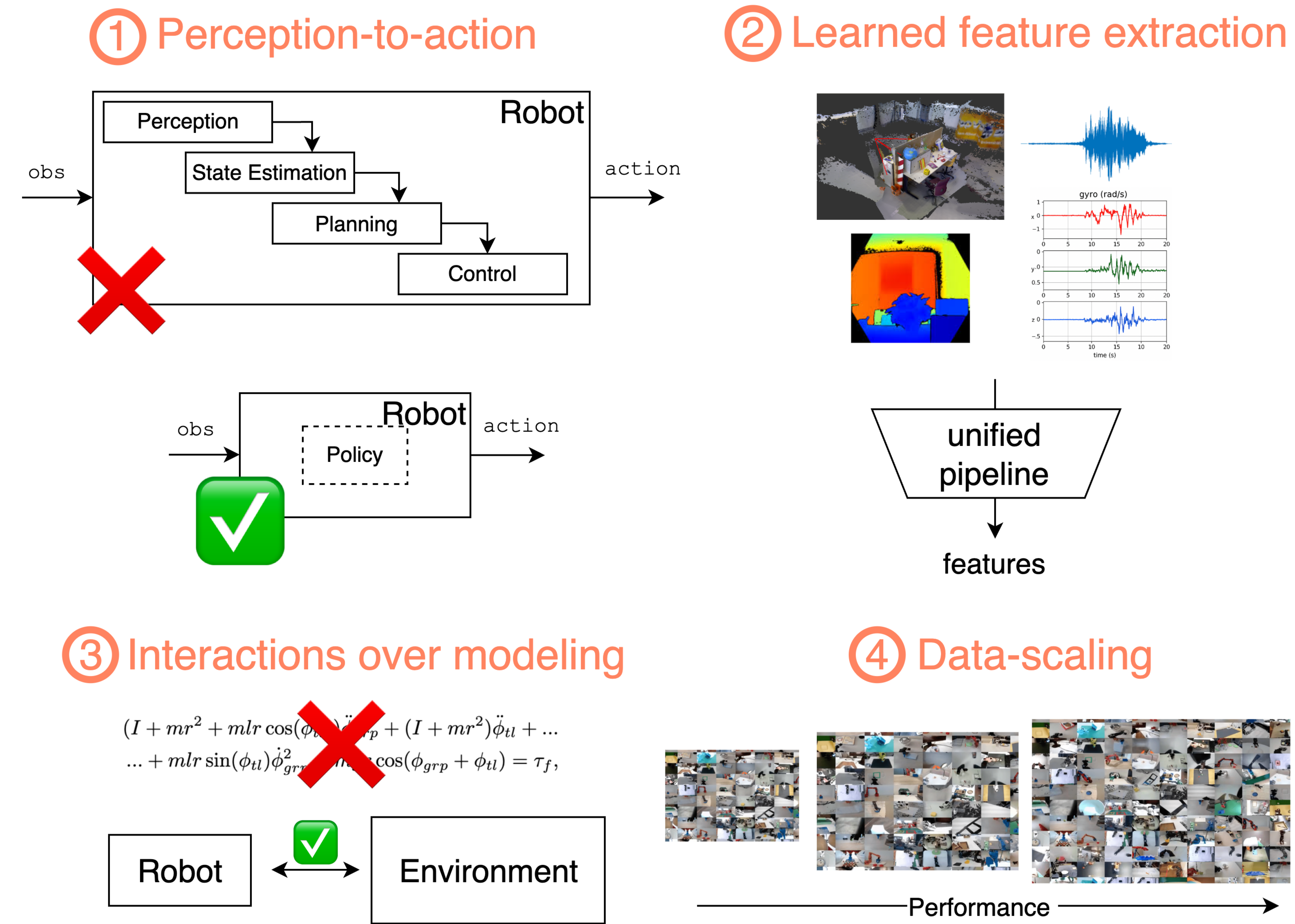}
    \caption{Learning-based robotics streamlines perception-to-action by learning a (1) unified high-level controller capable to take (2) high-dimensional, unstructured sensorimotor information. Learning (3) does not require a dynamics model and instead focuses on interaction data, and (4) empirically correlates with
    the scale of the data used.
    }
    \label{fig:robot-learning-upsides}
\end{figure}

Learning-based techniques for robotics naturally address the limitations presented in Section~\ref{sec:classical} (Figure~\ref{fig:robot-learning-upsides}).
In particular, learning-based techniques typically rely on monolithich prediction-to-action pipelines (\emph{visuomotor policies}) which do directly map sensorimotor inputs to predicted actions, streamlining control policies by removing the need to interface multiple components.
Mapping sensory inputs to actions also makes it possible to incorporate diverse input modalities, leveraging the automatic feature extraction capabilities of modern learning systems. 
Moreover, learning-based approaches can, in principle, bypass explicit modeling altogether and instead rely solely on interaction data---an advantage that proves transformative when dynamics are difficult to model or entirely unknown.
Lastly, learning for robotics (\emph{robot learning}) is naturally well posed to leverage the growing amount of robotics data openly available, just as computer vision and natural language processing did historically benefit from large-scale corpora of data, in great part overlooked by dynamics-based approaches.

\begin{wrapfigure}[23]{r}{0.3\textwidth}
    \vspace{-\intextsep}
    \centering
    \includegraphics[width=\linewidth]{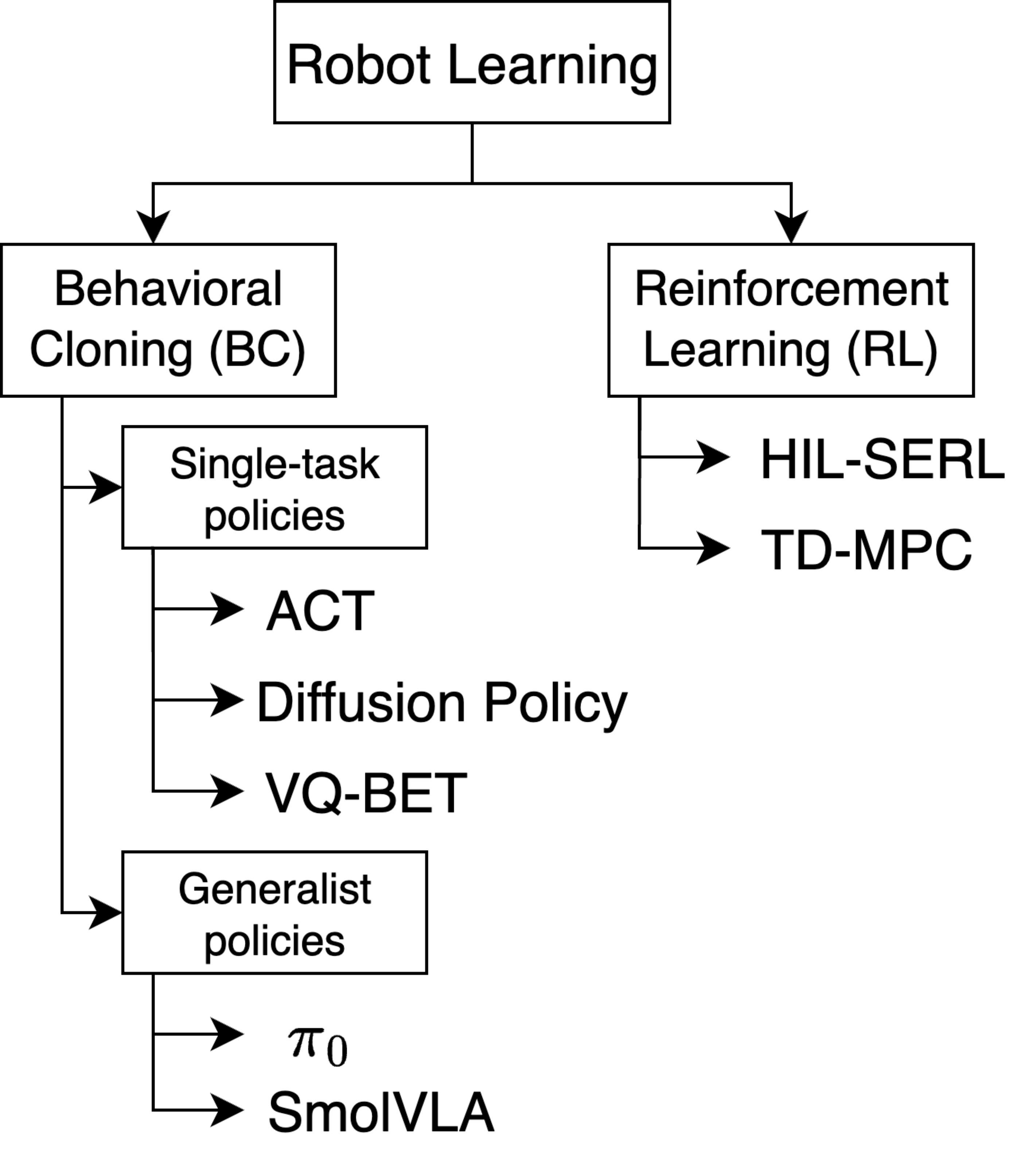}
    \caption{Overview of the robot learning methods implemented in \lerobot. All algorithms are implemented in Pytorch. References:~\citet{zhaoLearningFineGrainedBimanual2023,chiDiffusionPolicyVisuomotor2024,leeBehaviorGenerationLatent2024,black$p_0$VisionLanguageActionFlow2024,shukorSmolVLAVisionLanguageActionModel2025,luoPreciseDexterousRobotic2024,hansenTemporalDifferenceLearning2022} (top-to-bottom, left-to-right).}
    \label{fig:robot-learning-atlas}
\end{wrapfigure}

Being a field at its relative nascent stages, no prevalent technique(s) proves distinctly better than any other in the domain of robot learning.
Still, two major classes of methods gained prominence: \highlight{Reinforcement Learning (RL)} and \highlight{Behavioral Cloning (BC)} (Figure~\ref{fig:robot-learning-atlas}).
In this section, we provide a conceptual overview of applications of RL to robotics, as well as introduce practical examples of how to use RL within \lerobot.
We then introduce the major limitations RL suffers from, to introduce BC techniques in Section~\ref{sec:learning-imitation} and Section~{sec:learning-foundation}.

In Figure~\ref{fig:robot-learning-atlas} we deliberately include generalist robot models~\citep{black$p_0$VisionLanguageActionFlow2024,shukorSmolVLAVisionLanguageActionModel2025} alongside task-specific BC methods.
While significantly different in spirit---\emph{generalist} models are language-conditioned and use instructions to generate motion valid across many tasks, while \emph{task-specific} models are typically not language-conditioned and used to perform a single task---\emph{foundation} models are still largely trained to reproduce trajectories contained in a (large) training set of input demonstrations.
Thus, we argue generalist policies can indeed be grouped alongside other task-specific BC methods, as they both leverage similar training data and schemas.
Figure~\ref{fig:robot-learning-atlas} illustrates this categorization graphically, explicitly listing all the robot learning policies currently available in \lerobot: Action Chunking with Transformers (ACT)~\citep{zhaoLearningFineGrainedBimanual2023}, Diffusion Policy~\citep{chiDiffusionPolicyVisuomotor2024}, Vector-Quantized Behavior Transformer (VQ-BeT)~\citep{leeBehaviorGenerationLatent2024}, \( \pi_0 \)~\citep{black$p_0$VisionLanguageActionFlow2024}, SmolVLA~\citep{shukorSmolVLAVisionLanguageActionModel2025}, Human-in-the-loop Sample-efficient RL (HIL-SERL)~\citep{luoPreciseDexterousRobotic2024} and TD-MPC~\citep{hansenTemporalDifferenceLearning2022}.

\begin{figure}
    \centering
    \includegraphics[width=0.8\linewidth]{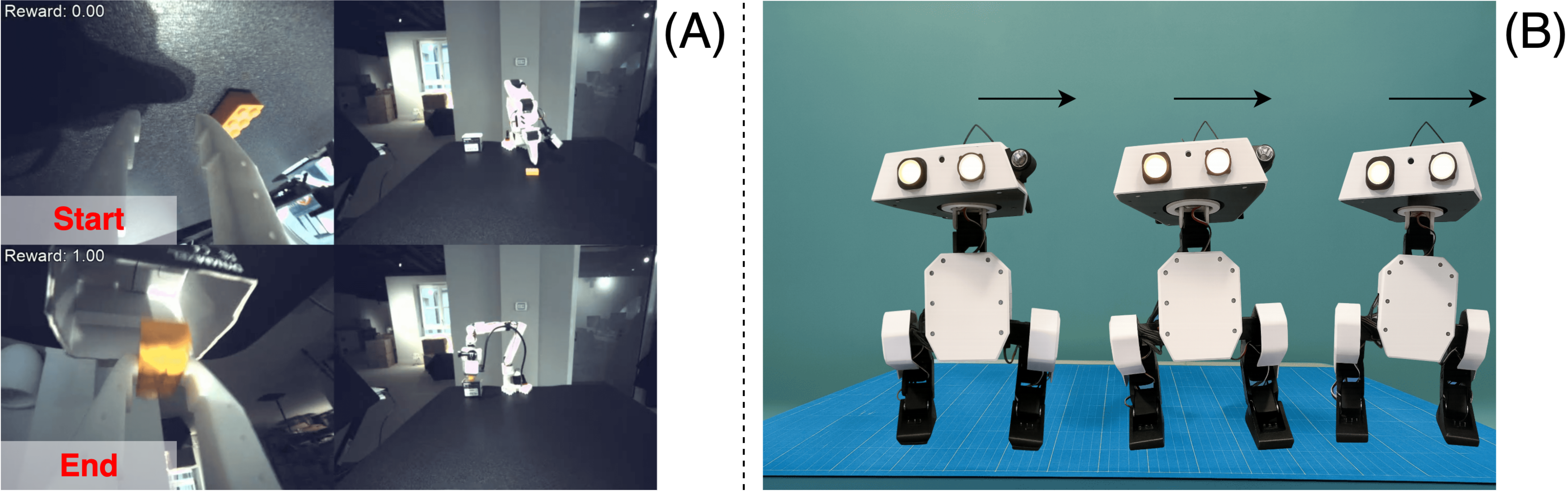}
    \caption{Examples of two different robotics tasks performed using RL. In the manipulation task (A) an agent learns to reach for a yellow plastic block in its environment, and to put it inside of a box. In the locomotion task (B) an agent learns to move its center of mass sideways without falling.}
    \label{fig:robotics-with-rl-examples}
\end{figure}

Applications of RL to robotics have been studied long enough that the relationship between these two disciplines has been compared to that of physics and matematics~\citep{koberReinforcementLearningRobotics}.
Indeed, due to their inherently interactive and sequential nature, robotics control problems can be directly cast as RL problems.
Figure~\ref{fig:robotics-with-rl-examples} presents two of such cases.
Reaching for an object to then move it somewhere else in the scene is a sequential problem where over time the controller needs to adjust the position of the robot arm based on the current configuration and the (possibly varying) position of the object.
Figure~\ref{fig:robotics-with-rl-examples} also shows an example of a locomotion problem, where sequentiality is inherent in the problem formulation: while sliding to the side, the controller needs to keep adjusting to the robot's to avoid failure (falling).

\subsection{A (Concise) Introduction to RL}
\begin{figure}
    \centering
    \includegraphics[width=0.5\linewidth]{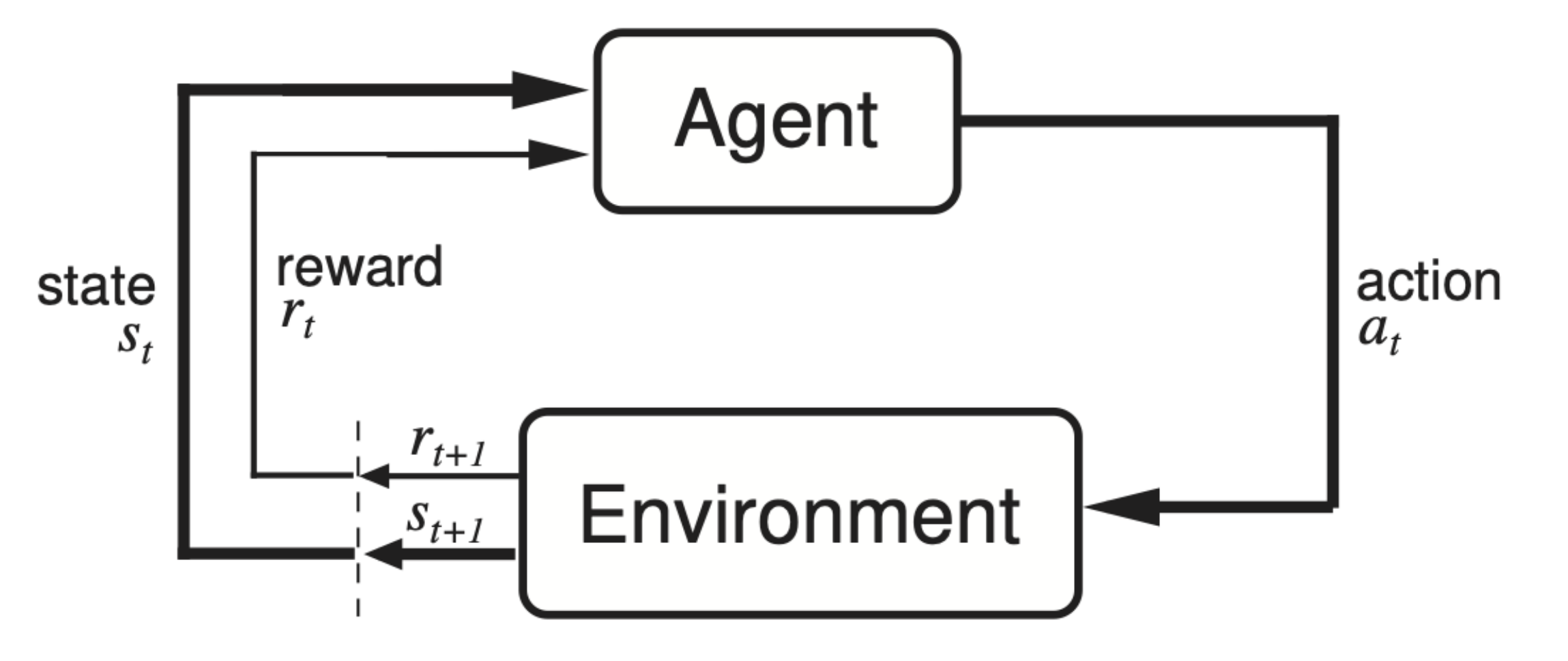}
    \caption{Agent-Environment interaction diagram (image credits to~\citet{suttonReinforcementLearningIntroduction2018}).}
    \label{fig:rl-most-famous-pic}
\end{figure}

The RL framework~\citep{suttonReinforcementLearningIntroduction2018}, which we briefly introduce here, has often been used to tackle robotics problems~\citep{koberReinforcementLearningRobotics}.
RL is a subfield within ML fundamentally concerned with the development of autonomous systems (\emph{agents}) capable to \emph{continuously behave} in an evolving environment, developing (ideally, well-performing) control strategies (\emph{policies}).
Crucially for robotics, RL agents improve through trial and error, bypassing explicit models of the problem dynamics in favor of interaction data.
In RL, this feedback loop between actions and outcomes (Figure~\ref{fig:rl-most-famous-pic}) is established through the agent sensing a scalar quantity (\emph{reward}) measuring how desirable a given \emph{transition} is for the accomplishment of its goal.

Formally, interactions between an agent and its environment are typically modeled via a Markov Decision Process (MDP)~\citep{bellmanMarkovianDecisionProcess1957}.
Representing robotics problems via MDPs offers several advantages, including (1) incorporating uncertainty through MDP's inherently stochastic formulation and (2) providing a theoretically-sound framework for learning \emph{without} an explicit model of the environment dynamics.
While accommodating a continuous time formulation too, MDPs are typically considered in discrete time in RL, assuming interactions to atomically take place at discrete \emph{timestep} \( t=0,1,2,3, \dots, T \).
MDPs allowing for an unbounded number of interactions (\( T \to + \infty \)) are termed \emph{infinite-horizon}, and opposed to \emph{finite-horizon} MDPs in which \( T \) is finite.
Unless diversely specified, we will only be referring to discrete-time finite-horizon (\emph{episodic}) MDPs.

Formally, a lenght-\(T\) Markov Decision Process (MDP) is a tuple \( \mathcal M = \langle \statespace, \actionspace, \dynamics, r, \gamma, \rho, T \rangle \), where:
\begin{itemize}
    \item \(\statespace\) is the \emph{state space}; \(\state \in \statespace\) denotes the (possibly non-directly observable) environment state at time \(t\). In robotics, states often comprise robot configuration and velocities (\(q_t, \dot q_t\)), and can also accomodate sensor readings such as camera or audio streams.
    \item \(\actionspace\) is the \emph{action space}; \(\action \in \actionspace\) may represent joint torques, joint velocities, or even end-effector commands at timestep \( t \). In general, actions correspond to commands intervenings on the configuration of the robot. 
    \item \(\dynamics\) represents the (possibly non-deterministic) environment dynamics, with \(\dynamics: \statespace \times \actionspace \times \statespace \mapsto [0, 1] \), \( \dynamics \, \transition = \transitionprob \). For instance, for a planar manipulator dynamics could be considered deterministic when the environment is fully described (Figure~\ref{fig:planar-manipulation-simple}), and stochastic when unmodeled disturbances depending on non-observable parameters intervene (Figure~\ref{fig:planar-manipulator-box-velocity}).
    \item \(r: \statespace \times \actionspace \times \statespace \to \mathbb R\) is the \emph{reward function}, weighing the transition \( \transition \) in the context of the achievement of an arbitrary goal. For instance, a simple reward function for quickly moving along the \( x \) axis (Figure~\ref{fig:robotics-with-rl-examples}) could be based on the absolute position of the robot along the \( x \) axis~(\(p_{x_t}\)), present negative penalties for falling over (measured from \( p_{z_t} \)) and a introduce bonuses \( \dot p_{x_t} \) for speed, \(r \transition \equiv r(\state) = p_{x_t} \cdot \dot p_{x_t} - \tfrac{1}{p_{z_t}} \).
\end{itemize}
Lastly, \(\gamma \in [0,1) \) represent the discount factor regulating preference for immediate versus long-term reward (with an effective horizon equal to \( \tfrac{1}{1-\gamma} \)), and \( \rho \) is the distribution over \(\statespace \) for the MDP's \emph{initial}, \( s_0 \sim \rho \).

Therefore, a length-\(T\) \emph{trajectory} is the (random) sequence
\begin{equation}\label{eq:trajectory_definition}
    \tau = \trajectory,
\end{equation}
with per-step rewards defined as \(r_t = r \transition \) for ease of notation.
Interestingly, assuming both the environment dynamics and conditional distribution over actions given states---i.e., the \emph{policy}---to be \emph{Markovian}:
\begin{align}
\mathbb P(\stateplusone \vert s_t, a_t, s_{t-1}, a_{t-1}, \dots s_0, a_0 ) &= \mathbb P \transitiongiven \label{eq:dynamics_markovian} \\
\mathbb P(\action \vert \state, a_{t-1}, s_{t-1}, s_0, a_0) &= \mathbb P(\action \vert \state), \label{eq:policy_markovian}
\end{align}
the probability of observing a given trajectory \( \tau \) factorizes into:
\begin{equation}\label{eq:traj_prob}
    \mathbb P(\tau) = \mathbb P (s_0) \prod_{t=0}^{T-1} \mathbb P \transitiongiven \ \mathbb P(\action \vert \state).
\end{equation}

Policies \( \mathbb P(\action \vert \state) \) are typically indicated as \( \pi(\action \vert \state) \), often parametrized via \( \theta \), yielding \( \pi_\theta (\action \vert \state )\), and are traine by optimizing the (discounted) \emph{return} associated to a given \( \tau \), i.e. the (random) sum of measured rewards over an arbitrary trajectory, 
\[
    G(\tau) = \sum_{t=0}^{T-1} \gamma^{t} r_t.
\]
In that, agents seek to learn control strategies (\emph{policies}, \( \pi_\theta \)) maximizing the expected return \( \mathbb E_{\tau \sim \pi_\theta} G(\tau) \). 
For a given dynamics \( \mathcal D \)---i.e., for a given problem---taking the expectation over the (possibly random) trajectories resulting from acting according to a certain policy provides a direct, goal-conditioned ordering in the space of all the possible policies \( \Pi \), yielding the (maximization) target \( J : \Pi \mapsto \mathbb R \)
\begin{align}
    J(\pi_\theta) &= \mathbb E_{\tau \sim \mathbb P_{\theta; \mathcal D}} \left[ G(\tau) \right], \label{eq:RL-j-function} \\
    \mathbb P_{\theta; \mathcal D} (\tau) &= \rho \prod_{t=0}^{T-1} \mathcal D \transition \ \pi_\theta (\action \vert \state).\label{eq:traj-probabilities-for-policies}
\end{align}

\begin{figure}
    \centering
    \includegraphics[width=0.4\textwidth]{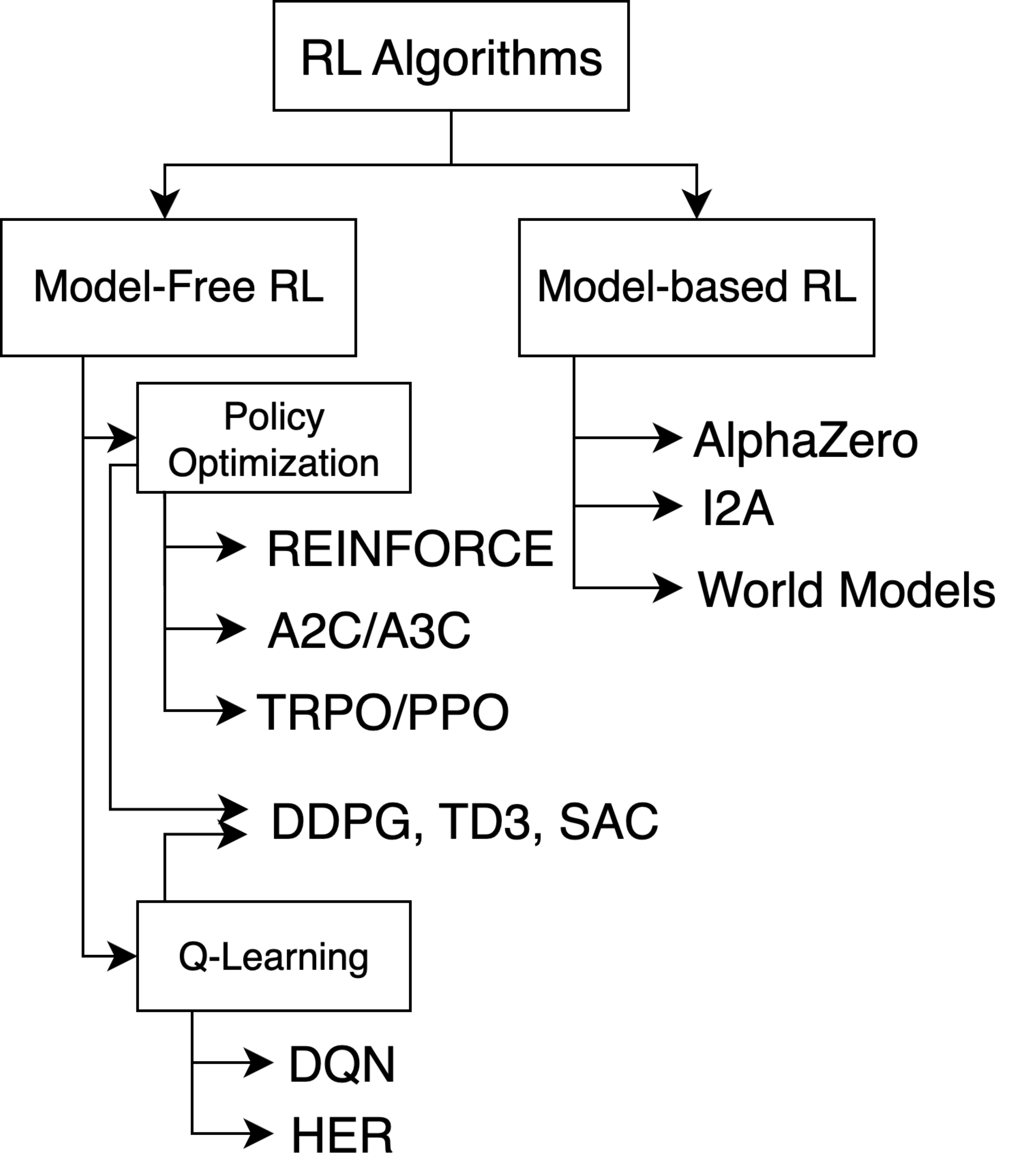}
    \caption{Popular RL algorithms. See~\citet{SpinningUp2018} for a complete list of citations.}
    \label{fig:rl-algos-atlas}
\end{figure}

Crucially, in the RL framework the agent is assumed to only \emph{observe} the environment dynamics and not to intervene on them, and thus eq.~\ref{eq:RL-j-function} varies exclusively with the policy followed.
In turn, MDPs naturally provide a framework to optimize over the space of the possible behaviors an agent might enact (\( \pi \in \Pi \)), searching for the \emph{optimal policy} \( \pi^* = \arg \max_{\theta} J(\pi_\theta) \), where \( \theta \) is the parametrization adopted by the policy set \( \Pi: \pi_\theta \in \Pi, \ \forall \theta \).
Besides providing a target for policy search, \( G(\tau) \) can also be used to discriminate between states \( s_t \) and \(\state, \action\) pairs.
Given any state \( s \in \statespace \)---e.g., given a configuration \( q \) of a robot---the \emph{state-value} function
\[
    V_\pi(s) = \mathbb E_{\tau \sim \pi} \left[ G(\tau) \big \vert s_0 = s \right]
\]
can be used to discriminate between desirable and undesirable state in terms of long-term (discounted) reward maximization, under a given policy \(\pi\).
Similarily, the \emph{state-action} value function also conditions the cumulative discounted reward on selecting action \( a \) when in \( s \), and thereafter act according to \( \pi \),
\[
    Q_\pi(s,a) = \mathbb E_{\tau \sim \pi} \left[ G (\tau) \big \vert s_0 = s, a_0=a \right].
\]
Importantly, value functions are interrelated:
\begin{align}
Q_\pi(s_t, a_t) &= \mathbb{E}_{\stateplusone \sim \mathbb P(\bullet \vert \state, \action)} \left[ r_t + \gamma V_\pi(\stateplusone) \right] \label{eq:q-as-v} \\
V_\pi(\state) &= \mathbb E_{\action \sim \pi(\bullet \vert \state)} \left[ Q_\pi (\state, \action) \right],
\label{eq:v-as-q}
\end{align}
inducing an ordering over states and state-action pairs under \( \pi \), and value functions are thus central to most RL algorithms.
A variety of algorithms have been developed in RL attempting to find (approximate) solutions to the problem of maximizing cumulative reward (we report some in Figure~\ref{fig:rl-algos-atlas}).

Popular approaches to continuous state and action space---such as those studied within robotics---include~\citet[TRPO]{schulmanTrustRegionPolicy2017},~\citet[PPO]{ schulmanProximalPolicyOptimization2017} and~\citet[SAC]{ haarnojaSoftActorCriticOffPolicy2018}.
Across manipulation~\citep{akkayaSolvingRubiksCube2019} and locomotion problems~\citep{leeLearningQuadrupedalLocomotion2020}, RL proved extremely effective in providing a platform to (1) leverage a unified, streamlined perception-to-action pipeline, (2) natively integrate propioperception with multi-modal high-dimensional sensory streams  (3) disregard a description of the environment dynamics, by focusing on observed interaction data rather than modeling, and (4) anchor policies in the experience collected and stored in datasets.
For a more complete survey of applications of RL to robotics, we refer the reader to~\citet{koberReinforcementLearningRobotics,tangDeepReinforcementLearning2025}.

\subsection{Real-world RL for Robotics}
Streamlined end-to-end control pipelines, data-driven feature extraction and a disregard for explicit modeling in favor of interaction data are all features of RL for robotics.
However, RL still suffers from limitations concerning safety and learning efficiency, particularly pressing for real-world robotics applications.

First, especially early in training, \highlight{actions are typically explorative, and thus may be erractic}.
On physical systems, untrained policies may command high velocities, self-collisiding configurations, or torques exceeding joint limits, leading to wear and potential hardware damage.
Mitigating these risks requires external safeguards (e.g., watchdogs, safety monitors, emergency stops), often incuring in a high degree of human supervision.
Further, in the typical episodic setting considered in most robotics problems, experimentation is substantially slowed down by the need to manually reset the environment over the course of training, a time-consuming and error-prone process.
Second, learning efficiently remains problematic in RL, \highlight{limiting the applicability of RL in real-world robotics due to consequently prohibitive timescales of training}.
Even strong algorithms such as SAC~\citep{haarnojaSoftActorCriticOffPolicy2018} typically require a large numbers of transitions \( \{ \sars \}_{t=1}^N \).
On real-world hardware, generating this data is time-consuming.

\begin{figure}
    \centering
    \includegraphics[width=0.7\linewidth]{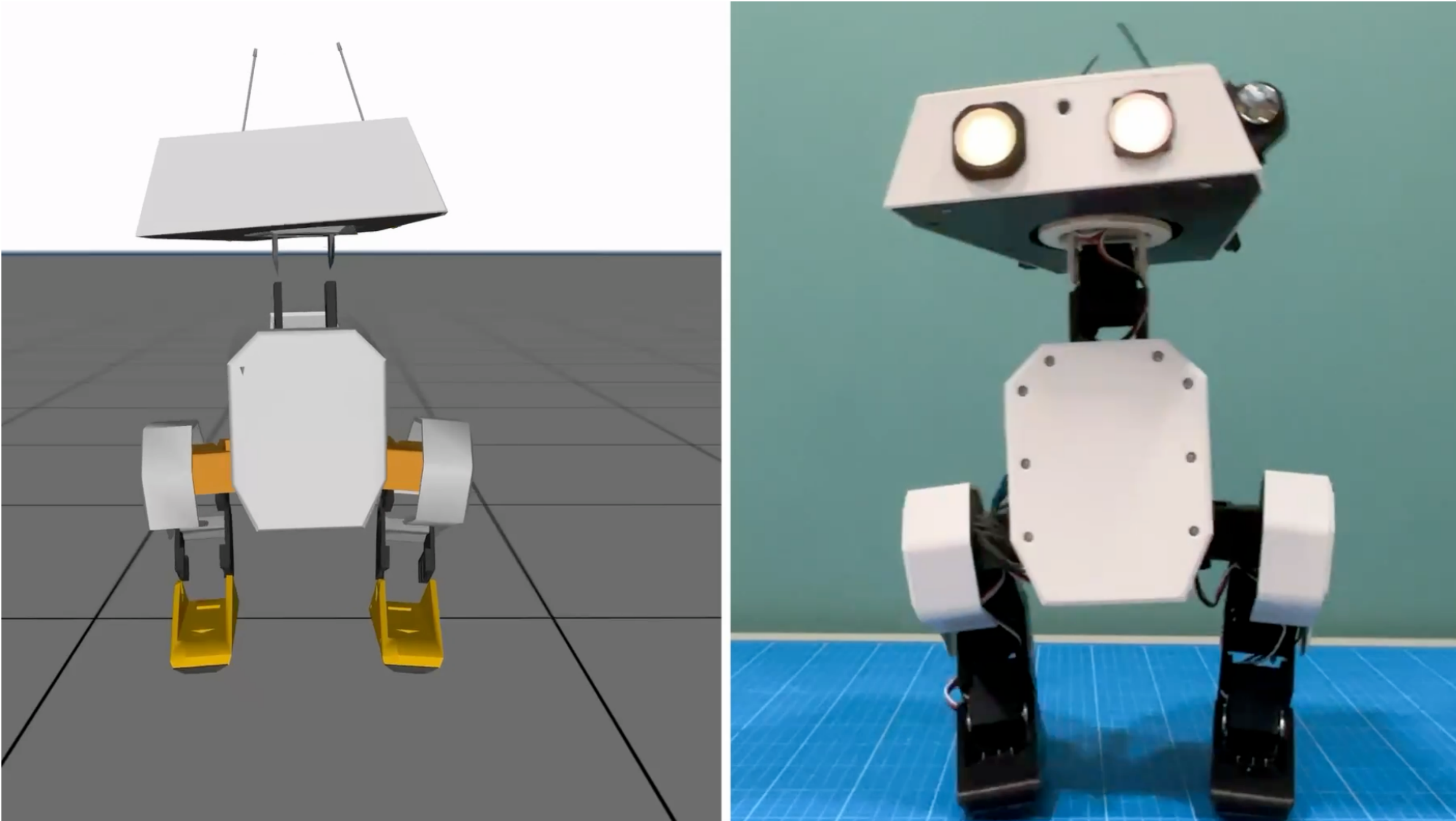}
    \caption{Simulated (left) vs. real-world (right) OpenDuck. Discrepancies in the simulation dynamics (\emph{reality gap}) pose risks to policy transfer.}
    \label{fig:synthetic-vs-real-duck}
\end{figure}

\begin{figure}
    \centering
    \includegraphics[width=0.9\linewidth]{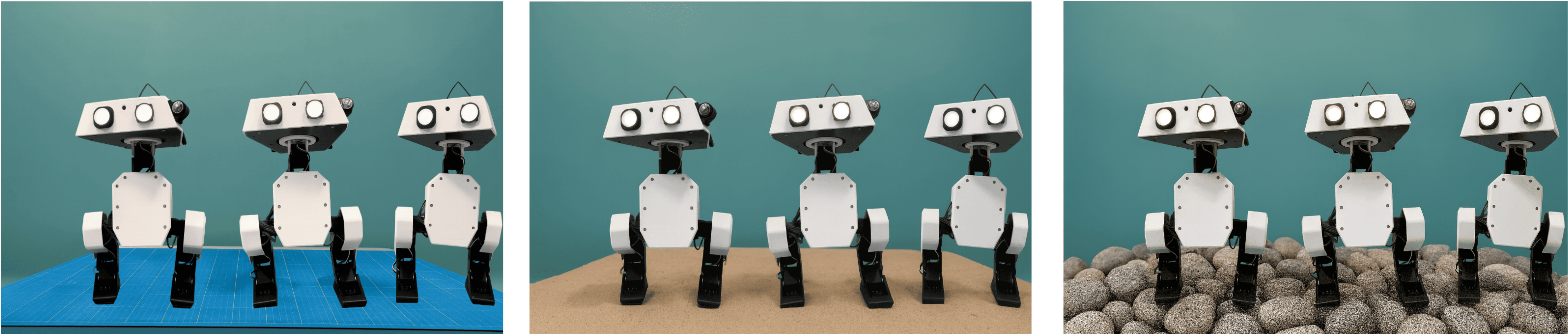}
    \caption{The same locomotion task can be carried out in different (simulated) domains (exemplified by the difference in terrains) at training time, resulting to increased robustness over diverse environment dynamics.}
    \label{fig:ducks-on-terrains}
\end{figure}

Training RL policies in simulation~\citep{tobinDomainRandomizationTransferring2017} addresses both issues, eliminating physical risk and dramatically increasing throughput.
Yet, simulators require significant modeling effort, and rely on assumptions (simplified physical modeling, instantaneous actuation, static environmental conditions, etc.) limiting the possibilities to transfer the policies learned in simulation, due the discrepancy between real and simulated environments (\emph{reality gap}, Figure~\ref{fig:synthetic-vs-real-duck}).
\emph{Domain randomization}~\citep{tobinDomainRandomizationTransferring2017} (DR) is a popular technique to overcome the reality gap, and consists in randomizing the parameters of the simulated environment during training, aiming at inducing robustness to specific disturbances.
In this, DR is typically employed to increase the diversity of scenarios over the course of training, improving on the performace sim-to-real transferred policies~\citep{akkayaSolvingRubiksCube2019,antonovaReinforcementLearningPivoting2017,jiDribbleBotDynamicLegged2023}.
In practice, DR is performed training in simulation on simulated dynamics \( \mathcal D \), further parametrized as \( \mathcal D \equiv \mathcal D_\xi \), with a \emph{dynamics} (random) vector \( \xi \) drawn an arbitrary distribution, \( \xi \sim \Xi \).
For instance, one could decide to randomize the friction coefficient of the surface in a locomotion task (Figure~\ref{fig:ducks-on-terrains}), or the center of mass of an object for a manipulation task.
Over the course of training---typically at each episode's reset---a new \( \xi \) is drawn, and used to specify the environment's dynamics for that episode.

While effective in transfering policies across the reality gap in real-world robotics~\citep{tobinDomainRandomizationTransferring2017,akkayaSolvingRubiksCube2019, jiDribbleBotDynamicLegged2023,tiboniDomainRandomizationEntropy2024}, DR often requires extensive manual engineering.
First, identifying which parameters to randomize---i.e., the \emph{support} \( \text{supp} (\Xi) \) of \( \Xi \)---is an inherently task specific process.
When locomoting over different terrains, choosing to randomize the friction coefficient is a reasonable choice, yet not completely resolutive as other factors (lightning conditions, external temperature, joints' fatigue, etc.) may prove just as important in practice, making selecting these parameters yet another source of brittlness.

Selecting the dynamics distribution \( \Xi \) is also non-trivial.
On the one hand, distributions with low entropy might risk to cause failure at transfer time, due to the limited robustness induced over the course of training.
On the other hand, excessive randomization may cause over-regularization and hinder performance~\citep{margolisRapidLocomotionReinforcement2022}.
Consequently, the research community investigated approaches to automatically select the randomization distribution \( \Xi \), using signals from the training process or tuning it to reproduce observed real-world trajectories.
\citet{akkayaSolvingRubiksCube2019} use a parametric uniform distribution \( \mathcal U(a, b) \) as \( \Xi \), widening the bounds \( a, b \) as training progresses and the agent's performance improves (AutoDR).
While effective, AutoDR requires significant tuning---the bounds are widened by a fixed, pre-specified amount \( \Delta \) along---and may disregard data when performance \emph{does not} improve after a distribution update~\citep{tiboniDomainRandomizationEntropy2024}. \citet{tiboniDomainRandomizationEntropy2024} propose a similar method to AutoDR (DORAEMON) to evolve \( \Xi \) based on the training signal, but with the key difference of explicitly maximizing the entropy of a parametric Beta distribution---inherently more flexible than uniform distributions---with learned updates instead of fixed \( \Delta \).
In this, DORAEMON proves particularly effective at dynamically increasing the entropy levels of the training distribution by employing an outer-loop max-entropy objective, tackled under performance constraints in the inner-loop RL problem.
Other approaches to automatically perform DR consist in specifically tuning \( \Xi \) to align as much as possible the simulation and real-world domains.
For instance,~\citet{chebotarClosingSimtorealLoop2019} interleave in-simulation policy training with repeated real-world policy rollouts used to adjust \( \Xi \) based on real-world data, while~\citet{tiboniDROPOSimtoRealTransfer2023} leverage a single, pre-collected set of real-world trajectories and tune \( \Xi \) under a simple likelihood objective.

While DR has shown promise, it does not address the main limitation that, even under the assumption that an ideal distribution \( \Xi \) was available, many robotics problems \highlight{cannot be simulated with high-enough fidelity under practical computational constraints}.
Simulating contact-rich manipulation of possibly deformable or soft materials---i.e., \emph{folding a piece of clothing}---can prove time-intensive, limiting the benefits of in-simulation training.

A perhaps more foundamental limitation of RL for robotics is the general unavailability of complicated tasks' \emph{dense} reward function, the design of which is essentially based on human expertise, ingenuity and trial-and-error.
In practice, \emph{sparse} reward functions can be used to conclude whether one specific goal has been attained---\emph{has this t-shirt been correctly folded?}---but unfortunately incur in more challenging learning.
As a result, despite notable successes, deploying RL directly on real-world robots at scale remains challenging.

To make the most of (1) the growing number of openly available datasets and (2) relatively inexpensive robots like the SO-100, RL could (1) be anchored in already-collected trajectories---limiting erratic and dangerous exploration---and (2) train in the real-world directly---bypassing the aforementioned issues with low-fidelity simulations.
In such a context, sample-efficient learning is also paramount, as training on the real-world is inherently time-bottlenecked.

Off-policy algorithms like Soft Actor-Critic (SAC)~\citep{haarnojaSoftActorCriticOffPolicy2018} tend to be more sample efficient then their on-policy counterpart~\citep{schulmanProximalPolicyOptimization2017}, due to the presence a \emph{replay buffer} used over the course of training.
Other than allowing to re-use past transitions \( \sars \), the replay buffer can also accomodate for the injection of previously-collected data in the training process~\citep{ballEfficientOnlineReinforcement2023}.
Using expert demonstrations to guide learning together with learned rewards, RL can be effectively carried out in the real-world~\citep{luoSERLSoftwareSuite2025}.
Interestingly, when complemented with in-training human interventions, real-world RL agents have been shown to learn policies with near-perfect success rates on challenging manipulation tasks in 1-2 hours~\citep{luoPreciseDexterousRobotic2024}.

\paragraph{Sample-efficient RL}
In an MDP, the optimal policy \( \pi^* \) can be derived from its associated \qfunction, \( Q^* \equiv Q_{\pi^*} \), and in particular the optimal action(s) \(\mu(\state)\) can be selected maximizing the optimal \qfunction \ over the action space,
\[
\mu(\state) = \max_{\action \in \mathcal A} Q^*(\state, \action).
\]
Interestingly, the \qopt-function satisfies a recursive relationship (\emph{Bellman equation}) based on a very natural intuition%
\footnote{Quote from~\citet{mnihPlayingAtariDeep2013}. The notation used has slightly been adapted for consistency with the rest of this tutorial.}:
\begin{quote}
    [...] If the optimal value \( Q^*(\stateplusone, a_{t+1}) \) of the [state] \(\stateplusone \) was known for all possible actions \(a_{t+1} \), then the optimal strategy is to select the action \( a_{t+1}\) maximizing the expected value of \( r_t + \gamma Q^*(s_{t+1}, a_{t+1}) \)
\[ 
Q^*(s_t, a_t) = \mathbb E_{s_{t+1} \sim \mathbb P(\bullet \vert s_t, a_t)} \left[ r_t + \gamma \max_{a_{t+1} \in \mathcal A} Q^*(s_{t+1}, a_{t+1}) \big\vert s_t, a_t  \right]
\]
\end{quote}

In turn, the optimal \qfunction \ %
is guaranteed to be self-consistent by definition.
\emph{Value-iteration} methods exploit this relationship (and/or its state-value counterpart, \( V^*(s_t) \) ) by iteratively updating an initial estimate of \qopt, \( Q_k \) using the Bellman equation as update rule (\emph{Q-learning}):
\[
    Q_{i+1}(s_t, a_t) \leftarrow \mathbb E_{s_{t+1} \sim \mathbb P(\bullet \vert s_t, a_t)} \left[ r_t + \gamma \max_{a_{t+1} \in \mathcal A} Q_i (s_{t+1}, a_{t+1}) \big\vert s_t, a_t  \right],  \quad i=0,1,2,\dots,K
\]
Then, one can derive the (ideally, near-optimal) policy by explicitly maximizing over the action space the final (ideally, near-optimal) estimate \( Q_K \approx Q^* \) at each timestep. 
Indeed, one can show that under certain assumptions on the MDP considered, \( Q_K \to Q^* \, \text{as } K \to \infty \).

Effective in its early applications to small-scale discrete problems, vanilla Q-learning was found complicated to scale to large \( \statespace \times \actionspace \) problems, in which storing \( Q : \statespace \times \actionspace \mapsto \mathbb R \) alone might result prohibitive. 
Also, vanilla Q-learning is not directly usable for \emph{continuous}, unstructured state-action space MPDs, such as those considered in robotics.
In their seminal work on \emph{Deep Q-Learning} (DQN),~\citet{mnihPlayingAtariDeep2013} propose learning Q-values using deep convolutional neural networks, thereby accomodating for large and even unstructured \emph{state} spaces.
DQN parametrizes the Q-function using a neural network with parameters \( \theta \), updating the parameters by sequentially minimizing the expected squared temporal-difference error (TD-error, \( \delta_i \)):
\begin{align}
\mathcal L(\theta_i) &= \mathbb E_{(s_t, a_t) \sim \chi(\bullet)} 
    \big[ 
        (\underbrace{y_i - Q_{\theta_i}(s_t, a_t)}_{\delta_i})^2 
    \big], \label{eq:dqn-loss} \\
    y_i &= \mathbb E_{s_{t+1} \sim \mathbb P(\bullet \vert s_t, a_t)} \big[ r_t + \gamma \max_{\action \in \mathcal A} Q_{\theta_{i-1}} (\stateplusone, a_{t+1}) \big], \label{eq:TD-target}
\end{align}
where \( \chi \) represents a behavior distribution over state-action pairs. 
Crucially, \( \chi \) can in principle be different from the policy being followed, effectively allowing to reuse prior data stored in a \emph{replay buffer} \( D \) in the form of \( \sars \) transitions, used to form the TD-target \( y_i \), TD-error \( \delta_i \) and loss function eq.~\ref{eq:dqn-loss} via Monte-Carlo (MC) estimates.

While effective in handling large, unstructured state spaces for discrete action-space problems, DQN's application to continous control problems proved challenging.
Indeed, in the case of high-capacity function approximators such as neural networks, solving \( \max_{a_t \in \mathcal A} Q_\theta(s_t, a_t) \) at each timestep is simply unfeasible due to the (1) continous nature of the action space (\( \actionspace \subset \mathbb R^n \) for some \( n \)) and (2) impossibility to express the policy with a cheap (ideally, even closed-form) formulation, so that \( \max Q_\theta \) could be solved analytically.
\citet{pmlr-v32-silver14} tackle these fundamental challenges by using a \emph{deterministic} function of the state \( s_t \) as policy, \( \mu_\phi(s_t) = a_t \), parametrized by \( \phi \). Thus, policies can be iteratively refined updating \( \phi \) along the direction:
\begin{equation}\label{eq:deterministic-pg}
    d_\phi = \mathbb E_{s_t \sim \mathbb P (\bullet)} \left[ \nabla_\phi Q(s_t, a_t)\vert_{a_t = \mu_\phi(s_t)} \right] = \mathbb E_{s_t \sim \mathbb P(\bullet)} \left[ \nabla_{a_t} Q(s_t, a_t) \vert_{a_t = \mu_\phi(s_t)} \cdot \nabla_\phi \mu(s_t) \right]
\end{equation}
Provably, eq.~\ref{eq:deterministic-pg} is the \emph{deterministic policy gradient} (DPG) of the policy \(\mu_\phi \)~\citep{pmlr-v32-silver14}, so that updates \( \phi_{k+1}\leftarrow \phi_k + \alpha d_\phi \) are guaranteed to increase the (deterministic) cumulative discounted reward, \( J(\mu_\phi) \).
~\citet{lillicrapContinuousControlDeep2019a} extended DPG to the case of (1) high-dimensional unstructured observations and (2) continuous action spaces, introducing Deep Deterministic Policy Gradient (DDPG), an important algorithm in RL and its applications to robotics.
DDPG adopts a modified TD-target compared to eq.~\ref{eq:TD-target}, by maintaining a policy network used to select actions, yielding
\begin{equation}\label{eq:TD-target-ddpg}
y_i = \mathbb E_{s_{t+1} \sim \mathbb P(\bullet \vert s_t, a_t)} \big[ r_t + \gamma Q_{\theta_{i-1}} (\stateplusone, \mu_\phi(\stateplusone)) \big] .
\end{equation}
Similarily to DQN, DDPG also employs the same replay buffer mechanism, reusing past transitions over training for increased sample efficiency and estimate the loss function via MC-estimates.

Soft Actor-Critic (SAC)~\citep{haarnojaSoftActorCriticOffPolicy2018} is a derivation of DDPG in the max-entropy (MaxEnt) RL framework, in which RL agents are tasked with \highlight{maximizing the discounted cumulative reward, while acting as randomly as possible}.
MaxEnt RL~\citep{haarnojaReinforcementLearningDeep2017b} has proven particularly robust thanks to the development of diverse behaviors, incentivized by its entropy-regularization formulation.
In that, MaxEnt revisits the RL objective \( J (\pi) \) to specifically account for the policy entropy \( \mathcal H(\pi (\bullet \vert s_t)) \),
\begin{align}
    J(\pi) &= \sum_{t=0}^T \mathbb{E}_{(s_t, a_t) \sim \chi} \left[ r_t + \alpha \mathcal H(\pi (\bullet \vert s_t)) \right].
    \label{eq:J-soft}
\end{align}
This modified objective results in the \emph{soft} TD-target:
\begin{equation}\label{eq:soft-td-target}
    y_i = \mathbb E_{s_{t+1} \sim \mathbb P( \bullet \vert s_t, a_t)} \left[ r_t + \gamma \left( Q_{\theta_{i-1}} (\stateplusone, a_{t+1}) - \alpha \log \pi_\phi(a_{t+1} \vert \stateplusone) \right) \right], \quad a_{t+1} \sim \pi_\phi(\bullet \vert s_t)
\end{equation}
Similarily to DDPG, SAC also maintains an explicit policy, trained under the same MaxEnt framework for the maximization of eq.~\ref{eq:J-soft}, updated using:
\begin{equation}\label{eq:sac-policy-update}
    \pi_{k+1} \leftarrow \arg\min_{\pi^\prime \in \Pi} \DKL \left(\pi^\prime (\bullet \vert \state) \bigg\Vert \frac{\exp(Q_{\pi_k}(s_t, \bullet))}{Z_{\pi_k}(s_t)} \right)
\end{equation}
The update rule provided in eq.~\ref{eq:sac-policy-update} optimizes the policy while projecting it on a set \( \Pi \) of tractable distributions (e.g., Gaussians,~\citet{haarnojaReinforcementLearningDeep2017b}).

\paragraph{Sample-efficient, data-driven RL}
Sampling \( \sars \) from the replay buffer \( D \) conveniently allows to approximate expectations for TD-target and TD-error through Monte-Carlo (MC) estimates.
The replay buffer \( D \) also proves extremely useful in maintaining a history of previous transitions and using it for training, improving on sample efficiency.
Furthermore, it also naturally provides an entry point to inject offline trajectories recorded by a human demonstrator into the training process.

Reinforcement Learning with Prior Data (RLPD)~\citep{ballEfficientOnlineReinforcement2023} is an Offline-to-Online RL algorithm leveraging prior data to effectively accelerate the training of a SAC agent.
Unlike previous works on Offline-to-Online RL, RLPD avoids any pre-training and instead only uses the available offline data \( D_\text{offline} \) to improve online-learning from scratch.
During each training step, transitions from both the offline and online replay buffers are sampled in equal proportions, and used in the underlying SAC routine.
Together with other implementation details (using LayerNorm layers to prevent value overestimation, and the use of ensembles techniques to form the TD-target), RLPD proves a particularly simple yet effective approach to use \( D_\text{offline} \) for Offline-to-Online RL.

\paragraph{Sample-efficient, data-driven, real-world RL}
Despite the possibility to leverage offline data for learning, the effectiveness of real-world RL training is still limited by the need to define a task-specific, hard-to-define reward function.
Further, even assuming to have access to a well-defined reward function, typical robotics pipelines rely on augmenting propioperceptive inputs with camera streams, and thus even well-defined rewards would need to be defined starting from unstructured observation---a challenging assumption in practice.
In their technical report,~\citet{luoSERLSoftwareSuite2025} empirically address the needs (1) to define a reward function and (2) to use it starting from unstructured, image observations.
In particular,~\citet[SERL]{luoSERLSoftwareSuite2025} introduces a suite of tools streamlining training of \emph{reward classifiers} \( c \), as well as jointly learn forward-backward controllers to speed up real-world RL.

Reward classifiers are particularly useful in treating complex, dynamic tasks---e.g., folding a t-shirt---for which a precise reward formulation is arbitrarily complex to obtain, or that do require significant shaping and are more easily learned directly from demonstrations of success (\(e^+\)) or failure (\(e^-\)) states, rather than from a precise formulation of \( r_t \), with a natural target for the reward classifier being \( r(s) = \log c(e^+ \ vert s ) \).
Furthermore,~\citet{luoSERLSoftwareSuite2025} demonstrate the benefits of learning separate (1) \emph{forward} and (2) \emph{backward} controllers---parametrized by separate policies---where (1) the former learns to execute a task to completion and (2) the latter learns to reset the environment to its initial state from terminal states, thereby aiding training in real-world episodic settings.

Lastly, in order to improve on the robustness of their approach to different goals while maintaing practical scalability,~\citet{luoSERLSoftwareSuite2025} introduced a modified state and action space, expressing proprioperceptive configurations \( q \)  and actions \( \dot q \) in the frame of the end-effector pose at \( t=0 \).
Randomizing the initial pose of the end-effector (\( s_0 \)),~\citet{luoSERLSoftwareSuite2025} achieved a similar result to that of manually randomizing the environment at every timestep, but with the benefit of maintaining the environment in the same condition across multiple training episodes, achieving higher scalability of their method thanks to the increased practicality of their approach.

\begin{figure}
    \centering
    \includegraphics[width=0.8\linewidth]{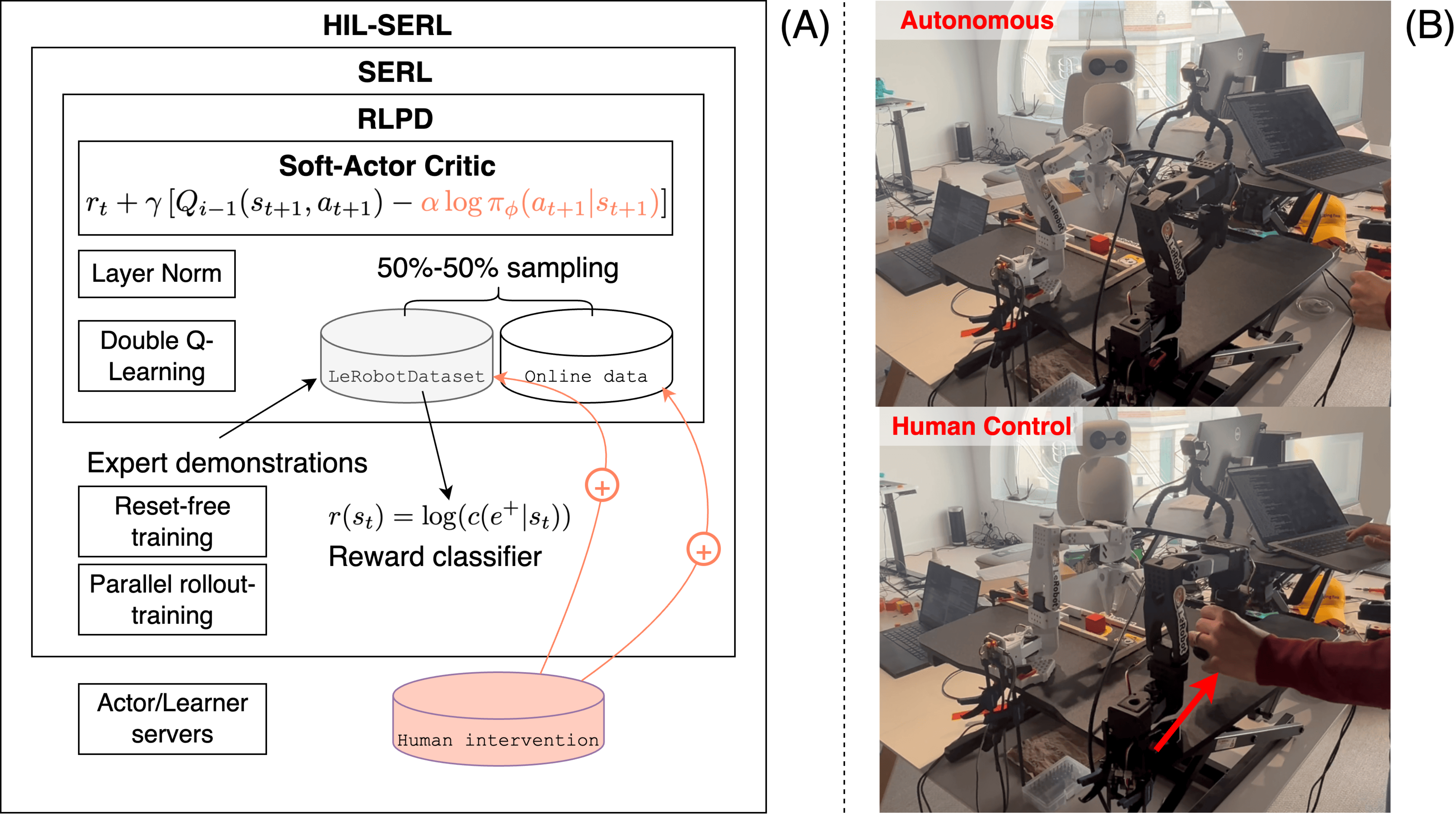}
    \caption{(A) HIL-SERL allows for real-world training of high performance RL agents by building on top advancements presented by of SAC, RLPD and SERL. (B) Example of human intervention during a HIL-SERL training process on a real-world SO-100.}
    \label{fig:hil-serl-blocks}
\end{figure}

Building on off-policy deep Q-learning with replay buffers, entropy regularization for better exploration, expert demonstrations to guide learning, and a series of tools and recommendations for real-world training using reward classifiers (Figure~\ref{fig:hil-serl-blocks}),~\citet{luoPreciseDexterousRobotic2024} introduce human interactions during training, learning near-optimal policies in challenging real-world manipulation tasks in 1-2 hours.

Human-in-the-Loop, Sample Efficient Robot reinforcement Learning (HIL-SERL)~\citep{luoPreciseDexterousRobotic2024} augments offline-to-online RL with targeted human corrections during training, and employs prior data to (1) train a reward classifier and (2) bootstrap RL training on expert trajectories.
While offline demonstrations provide the initial dataset seeding learning and constraining early exploration, interactive, online corrections allow a human supervisor to intervene on failure modes and supply targeted interventions, greatly aiding the learning process~\citep{luoPreciseDexterousRobotic2024}.
Crucially, human intervention data is stored in \emph{both} the offline and online replay buffers, differently from the autonomous transitions generated at training time and stored in the online buffer only.
In turn, given an intervention timestep \( k \in (0, T) \), length-\(K\) human intervention data \( \{ s^{\text{human}}_k, a^{\text{human}}_k, r^{\text{human}}_k, s^{\text{human}}_{k+1},\}_{k=1}^K \) is more likely to be sampled than the data generated online during training, providing stronger supervision to the agent while still allowing for autonomous learning.
Empirically, HIL-SERL attains near-perfect success rates (99\%+) on diverse manipulation tasks within 1-2 hours of training~\citep{luoPreciseDexterousRobotic2024}, underscoring how offline datasets with online RL can markedly improve stability and data efficiency, and ultimately even allow real-world RL-training.

\subsubsection{Code Example: Real-world RL}

\begin{figure}
    \centering
    \includegraphics[width=0.9\textwidth]{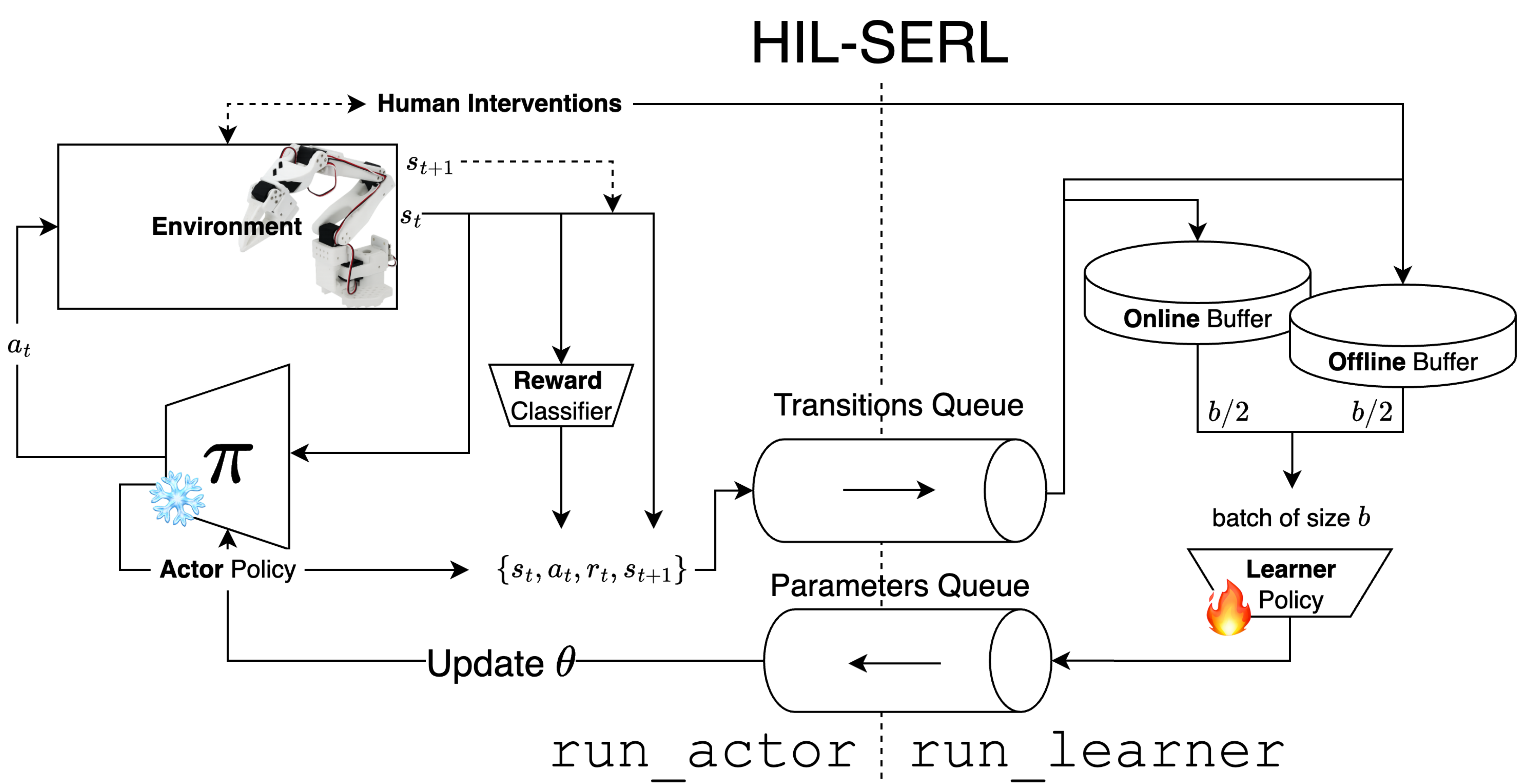}
    \caption{HIL-SERL is a SOTA RL algorithm for training control policies directly in the real-world. Its implementation in \lerobot~relies on a decoupled actor-learner architecture, communicating over processes (and possibly networks) with queues used to share (1) transitions \( \sars \) and (2) parameters \( \theta \).}
    \label{fig:ch3-hil-serl-architecture}
\end{figure}

This example shows how to use the HIL-SERL implementation supported by \lerobot.
This code example is organized into four parts: we first show how to train a reward classifier from a custom set of demonstrations, then define the \texttt{Actor} and \texttt{Learner} components, and finally, we bring them together in a complete script showing how to use HIL-SERL in practice.

At a higher level, the HIL-SERL architecture (Figure~\ref{fig:ch3-hil-serl-architecture}) relies on two main components:
\begin{itemize}
    \item An \texttt{Actor}, running a frozen policy network used to interact with the environment and obtain observations. Observations are used to both condition the frozen actor in selecting the action to enact, and to form \( \sars \) transitions that are shared with the \texttt{Learner}. Rewards are inferred using a custom, learned reward classifier trained on a dataset of offline demonstrations.
    \item A \texttt{Learner}, used to optimize the policy's parameters \( \theta \) for maximum expected return. The learner samples batches of offline data from online and offline buffers in equal proportion~\citep{ballEfficientOnlineReinforcement2023}, and shares updated parameters with the \texttt{Actor}.
\end{itemize}

The HIL-SERL architecture presented in this example can be exclusively run locally, but the implementation in \lerobot~also allows the \texttt{Actor} and \texttt{Learner} to run on two separate machines connected by the network.

\begin{pbox}[label={ex:train_reward_classifier}]{Training a Reward Classifier \\ \url{https://github.com/fracapuano/robot-learning-tutorial/blob/main/snippets/ch3/01_reward_classifier.py}}
\begin{lstlisting}[language=python]
import torch

from lerobot.datasets.lerobot_dataset import LeRobotDataset
from lerobot.policies.factory import make_policy, make_pre_post_processors
from lerobot.policies.sac.reward_model.configuration_classifier import RewardClassifierConfig

# Device to use for training
device = "mps"  # or "cuda", or "cpu"

# Load the dataset used for training
repo_id = "lerobot/example_hil_serl_dataset"
dataset = LeRobotDataset(repo_id)

# Configure the policy to extract features from the image frames
camera_keys = dataset.meta.camera_keys

config = RewardClassifierConfig(
    num_cameras=len(camera_keys),
    device=device,
    # backbone model to extract features from the image frames
    model_name="microsoft/resnet-18",
)

# Make policy, preprocessor, and optimizer
policy = make_policy(config, ds_meta=dataset.meta)
optimizer = config.get_optimizer_preset().build(policy.parameters())
preprocessor, _ = make_pre_post_processors(policy_cfg=config, dataset_stats=dataset.meta.stats)


# your HF username and model repo id for the reward classifier
classifier_id = "lerobot/reward_classifier_hil_serl_example"

# Instantiate a dataloader
dataloader = torch.utils.data.DataLoader(dataset, batch_size=16, shuffle=True)

# Training loop
num_epochs = 5
for epoch in range(num_epochs):
    total_loss = 0
    total_accuracy = 0
    for batch in dataloader:
        # Preprocess the batch and move it to the correct device.
        batch = preprocessor(batch)

        # Forward pass
        loss, output_dict = policy.forward(batch)

        # Backward pass and optimization
        optimizer.zero_grad()
        loss.backward()
        optimizer.step()

        total_loss += loss.item()
        total_accuracy += output_dict["accuracy"]

    avg_loss = total_loss / len(dataloader)
    avg_accuracy = total_accuracy / len(dataloader)
    print(
        f"Epoch {epoch + 1}/{num_epochs}, Loss: {avg_loss:.4f}, Accuracy: {avg_accuracy:.2f}%"
    )

print("Training finished!")

# You can now save the trained policy.
policy.push_to_hub(classifier_id)
\end{lstlisting}
\end{pbox}

\begin{pbox}[label={ex:hil_serl_defining_actor}]{Defining the \texttt{Actor} \\ \url{https://github.com/fracapuano/robot-learning-tutorial/blob/main/snippets/ch3/02_actor.py}}
\begin{lstlisting}[language=python]
import multiprocessing as mp
from queue import Empty

import torch
from pathlib import Path

from lerobot.envs.configs import HILSerlRobotEnvConfig
from lerobot.policies.sac.modeling_sac import SACPolicy
from lerobot.policies.sac.reward_model.modeling_classifier import Classifier
from lerobot.rl.gym_manipulator import make_robot_env
from lerobot.teleoperators.utils import TeleopEvents

MAX_EPISODES = 5
MAX_STEPS_PER_EPISODE = 20

def make_policy_obs(obs, device: torch.device = "cpu"):
    return {
        "observation.state": torch.from_numpy(obs["agent_pos"]).float().unsqueeze(0).to(device),
        **{
            f"observation.image.{k}": 
                torch.from_numpy(obs["pixels"][k]).float().unsqueeze(0).to(device)
            for k in obs["pixels"]
        },
    }

def run_actor(
    transitions_queue: mp.Queue,
    parameters_queue: mp.Queue,
    shutdown_event: mp.Event,
    policy_actor: SACPolicy,
    reward_classifier: Classifier,
    env_cfg: HILSerlRobotEnvConfig,
    device: torch.device = "mps",
    output_directory: Path | None = None
):
    """The actor process - interacts with environment and collects data.
    The policy is frozen and only the parameters are updated, popping the most recent ones 
    from a queue."""
    policy_actor.eval()
    policy_actor.to(device)

    reward_classifier.eval()
    reward_classifier.to(device)

    # Create robot environment inside the actor process
    env, teleop_device = make_robot_env(env_cfg)

    try:
        for episode in range(MAX_EPISODES):
            if shutdown_event.is_set():
                break

            obs, _info = env.reset()
            episode_reward = 0.0
            step = 0
            episode_transitions = []

            print(f"[ACTOR] Starting episode {episode + 1}")

            while step < MAX_STEPS_PER_EPISODE and not shutdown_event.is_set():
                try:
                    new_params = parameters_queue.get_nowait()
                    policy_actor.load_state_dict(new_params)
                    print("[ACTOR] Updated policy parameters from learner")
                except Empty:  # No new updated parameters available from learner, waiting
                    pass

                # Get action from policy
                policy_obs = make_policy_obs(obs, device=device)
                # predicts a single action, not a chunk of actions!
                action_tensor = policy_actor.select_action(policy_obs)
                action = action_tensor.squeeze(0).cpu().numpy()

                # Step environment
                next_obs, _env_reward, terminated, truncated, _info = env.step(action)
                done = terminated or truncated

                # Predict reward
                policy_next_obs = make_policy_obs(next_obs, device=device)
                reward = reward_classifier.predict_reward(policy_next_obs)

                if reward >= 1.0:  # success detected! halt episode
                    if not done:
                        terminated = True
                        done = True

                # In HIL-SERL, human interventions come from the teleop device
                is_intervention = False
                if hasattr(teleop_device, "get_teleop_events"):
                    # Real intervention detection from teleop device
                    teleop_events = teleop_device.get_teleop_events()
                    is_intervention = teleop_events.get(TeleopEvents.IS_INTERVENTION, False)

                # Store transition with intervention metadata
                transition = {
                    "state": policy_obs,
                    "action": action,
                    "reward": float(reward) if hasattr(reward, "item") else reward,
                    "next_state": policy_next_obs,
                    "done": done,
                    "truncated": truncated,
                    "complementary_info": {
                        "is_intervention": is_intervention,
                    },
                }

                episode_transitions.append(transition)

                episode_reward += reward
                step += 1

                obs = next_obs

                if done:
                    break

            # Send episode transitions to learner
            transitions_queue.put_nowait(episode_transitions)

    except KeyboardInterrupt:
        print("[ACTOR] Interrupted by user")
    finally:
        # Clean up
        if hasattr(env, "robot") and env.robot.is_connected:
            env.robot.disconnect()
        if teleop_device and hasattr(teleop_device, "disconnect"):
            teleop_device.disconnect()
        if output_directory is not None:
            policy_actor.save_pretrained(output_directory)
            print(f"[ACTOR] Latest actor policy saved at: {output_directory}")
        
        print("[ACTOR] Actor process finished")
\end{lstlisting}
\end{pbox}

\begin{pbox}[label={ex:hil_serl_defining_learner}]{Defining the \texttt{Learner} \\ \url{https://github.com/fracapuano/robot-learning-tutorial/blob/main/snippets/ch3/03_learner.py}}
\begin{lstlisting}[language=python]
import multiprocessing as mp
from queue import Empty, Full

import torch
import torch.optim as optim

from lerobot.policies.sac.modeling_sac import SACPolicy
from lerobot.rl.buffer import ReplayBuffer

LOG_EVERY = 10
SEND_EVERY = 10

def run_learner(
    transitions_queue: mp.Queue,
    parameters_queue: mp.Queue,
    shutdown_event: mp.Event,
    policy_learner: SACPolicy,
    online_buffer: ReplayBuffer,
    offline_buffer: ReplayBuffer,
    lr: float = 3e-4,
    batch_size: int = 32,
    device: torch.device = "mps",
):
    """The learner process - trains SAC policy on transitions streamed from the actor, 
    updating parameters for the actor to adopt."""
    policy_learner.train()
    policy_learner.to(device)

    # Create Adam optimizer from scratch - simple and clean
    optimizer = optim.Adam(policy_learner.parameters(), lr=lr)

    print(f"[LEARNER] Online buffer capacity: {online_buffer.capacity}")
    print(f"[LEARNER] Offline buffer capacity: {offline_buffer.capacity}")

    training_step = 0

    while not shutdown_event.is_set():
        # retrieve incoming transitions from the actor process
        try:
            transitions = transitions_queue.get(timeout=0.1)
            for transition in transitions:
                # HIL-SERL: Add ALL transitions to online buffer
                online_buffer.add(**transition)

                # HIL-SERL: Add ONLY human intervention transitions to offline buffer
                is_intervention = \
                    transition.get("complementary_info", {}).get("is_intervention", False)
                if is_intervention:
                    offline_buffer.add(**transition)
                    print(
                        f"[LEARNER] Human intervention detected!"
                        f"Added to offline buffer (now {len(offline_buffer)} transitions)"
                    )

        except Empty:
            pass  # No transitions available, continue

        # Train if we have enough data
        if len(online_buffer) >= policy_learner.config.online_step_before_learning:
            # Sample from online buffer (autonomous + human data)
            online_batch = online_buffer.sample(batch_size // 2)

            # Sample from offline buffer (human demonstrations only)
            offline_batch = offline_buffer.sample(batch_size // 2)

            # Combine batches - this is the key HIL-SERL mechanism!
            batch = {}
            for key in online_batch.keys():
                if key in offline_batch:
                    batch[key] = torch.cat([online_batch[key], offline_batch[key]], dim=0)
                else:
                    batch[key] = online_batch[key]

            loss, _ = policy_learner.forward(batch)

            optimizer.zero_grad()
            loss.backward()
            optimizer.step()
            training_step += 1

            if training_step % LOG_EVERY == 0:
                print(
                    f"[LEARNER] Training step {training_step}, Loss: {loss.item():.4f}, "
                    f"Buffers: Online={len(online_buffer)}, Offline={len(offline_buffer)}"
                )

            # Send updated parameters to actor every 10 training steps
            if training_step % SEND_EVERY == 0:
                try:
                    state_dict = {k: v.cpu() for k, v in policy_learner.state_dict().items()}
                    parameters_queue.put_nowait(state_dict)
                    print("[LEARNER] Sent updated parameters to actor")
                except Full:
                    # Missing write due to queue not being consumed (should happen rarely)
                    pass

    print("[LEARNER] Learner process finished")
\end{lstlisting}
\end{pbox}

\begin{pbox}[label={ex:hil_serl_full}]{Using HIL-SERL \\ \url{https://github.com/fracapuano/robot-learning-tutorial/blob/main/snippets/ch3/04_hil_serl.py}}
\begin{lstlisting}[language=python]
import multiprocessing as mp
import signal
from typing import Callable
from pathlib import Path

from lerobot.datasets.lerobot_dataset import LeRobotDataset
from lerobot.datasets.utils import hw_to_dataset_features
from lerobot.envs.configs import HILSerlProcessorConfig, HILSerlRobotEnvConfig
from lerobot.policies.sac.configuration_sac import SACConfig
from lerobot.policies.sac.modeling_sac import SACPolicy
from lerobot.policies.sac.reward_model.modeling_classifier import Classifier
from lerobot.rl.buffer import ReplayBuffer
from lerobot.rl.gym_manipulator import make_robot_env
from lerobot.robots.so100_follower import SO100FollowerConfig
from lerobot.teleoperators.so100_leader import SO100LeaderConfig


run_learner: Callable = ...  # use/modify the functions defined earlier
run_actor: Callable = ...

"""Main function - coordinates actor and learner processes."""

device = "mps"  # or "cuda" or "cpu"
output_directory = Path("outputs/robot_learning_tutorial/hil_serl")
output_directory.mkdir(parents=True, exist_ok=True)

# find ports using lerobot-find-port
follower_port = ...
leader_port = ...

# the robot ids are used the load the right calibration files
follower_id = ...
leader_id = ...

# A pretrained model (to be used in-distribution!)
reward_classifier_id = "lerobot/reward_classifier_hil_serl_example"
reward_classifier = Classifier.from_pretrained(reward_classifier_id)

reward_classifier.to(device)
reward_classifier.eval()

MAX_EPISODES = 5
MAX_STEPS_PER_EPISODE = 20

# Robot and environment configuration
robot_cfg = SO100FollowerConfig(port=follower_port, id=follower_id)
teleop_cfg = SO100LeaderConfig(port=leader_port, id=leader_id)
processor_cfg = HILSerlProcessorConfig(control_mode="leader")

env_cfg = HILSerlRobotEnvConfig(robot=robot_cfg, teleop=teleop_cfg, processor=processor_cfg)

# Create robot environment
env, teleop_device = make_robot_env(env_cfg)

obs_features = hw_to_dataset_features(env.robot.observation_features, "observation")
action_features = hw_to_dataset_features(env.robot.action_features, "action")

# Create SAC policy for action selection
policy_cfg = SACConfig(
    device=device,
    input_features=obs_features,
    output_features=action_features,
)

policy_actor = SACPolicy(policy_cfg)
policy_learner = SACPolicy(policy_cfg)

demonstrations_repo_id = "lerobot/example_hil_serl_dataset"
offline_dataset = LeRobotDataset(repo_id=demonstrations_repo_id)

# Online buffer: initialized from scratch
online_replay_buffer = ReplayBuffer(device=device, state_keys=list(obs_features.keys()))
# Offline buffer: Created from dataset (pre-populated it with demonstrations)
offline_replay_buffer = ReplayBuffer.from_lerobot_dataset(
    lerobot_dataset=offline_dataset, device=device, state_keys=list(obs_features.keys())
)

# Create communication channels between learner and actor processes
transitions_queue = mp.Queue(maxsize=10)
parameters_queue = mp.Queue(maxsize=2)
shutdown_event = mp.Event()


# Signal handler for graceful shutdown
def signal_handler(sig):
    print(f"\nSignal {sig} received, shutting down...")
    shutdown_event.set()


signal.signal(signal.SIGINT, signal_handler)
signal.signal(signal.SIGTERM, signal_handler)

# Create processes
learner_process = mp.Process(
    target=run_learner,
    args=(
        transitions_queue,
        parameters_queue,
        shutdown_event,
        policy_learner,
        online_replay_buffer,
        offline_replay_buffer,
    ),
    kwargs={"device": device},  # can run on accelerated hardware for training
)

actor_process = mp.Process(
    target=run_actor,
    args=(
        transitions_queue,
        parameters_queue,
        shutdown_event,
        policy_actor,
        reward_classifier,
        env_cfg,
        output_directory,
    ),
    kwargs={"device": "cpu"},  # actor is frozen, can run on CPU or accelerate for inference
)

learner_process.start()
actor_process.start()

try:
    # Wait for actor to finish (it controls the episode loop)
    actor_process.join()
    shutdown_event.set()
    learner_process.join(timeout=10)

except KeyboardInterrupt:
    print("Main process interrupted")
    shutdown_event.set()
    actor_process.join(timeout=5)
    learner_process.join(timeout=10)

finally:
    if learner_process.is_alive():
        learner_process.terminate()
    if actor_process.is_alive():
        actor_process.terminate()
\end{lstlisting}
\end{pbox}

\subsubsection{Limitations of RL in Real-World Robotics: Simulators and Reward Design}

Despite the advancements in real-world RL training, training RL agents for real-world tasks still suffers from the following limitations:
\begin{itemize}
\item In those instances where real-world training experience is prohibitively expensive to gather (e.g., Tokamak control~\citep{degraveMagneticControlTokamak2022}, Autonomous Stratospehere Navigation~\citep{bellemareAutonomousNavigationStratospheric2020})in-simulation training is often the only viable option. 
However, high-fidelity simulators for real-world problems can be difficult to build and maintain, especially for contact-rich manipulation and tasks involving deformable or soft materials.

\item Reward design is a fundamental source of brittleness in real-world RL pipelines. While shaping dense rewards is often necessary to guide exploration in long-horizon tasks, the process is error-prone and heavily reliant on human expertise and intuition. Poorly tuned terms can lead to specification gaming or convergence to local optima, making reward shaping a critical challenge for applying RL in practice. Sparse rewards that only signal successful trajectories can avoid these pitfalls but typically result in much slower learning due to reduced supervision.
\end{itemize}

Advances in learning to act from potentially large corpora of human demonstrations via Behavioral Cloning (BC) address both of these concerns.
Although suffering from an inherent suboptimality---imitation learning can at most match the performance level of the demonstrator---learning to reproduce expert demonstrations via BC has proven increasingly competitive and practical, bypassing the need for simulated environments and hard-to-define reward functions.

\newpage
\section{Robot (Imitation) Learning}
\label{sec:learning-imitation}

\epigraph{\emph{The best material model for a cat is another, or preferably the same cat}}{Norbert Wiener}

\begin{tldr}
Behavioral Cloning provides a natural platform to learn from real-world interactions without the need to design any reward function, and generative models prove more effective than point-wise policies at dealing with multimodal demonstration datasets.
\end{tldr}



\begin{figure}
    \centering
    \includegraphics[width=0.8\textwidth]{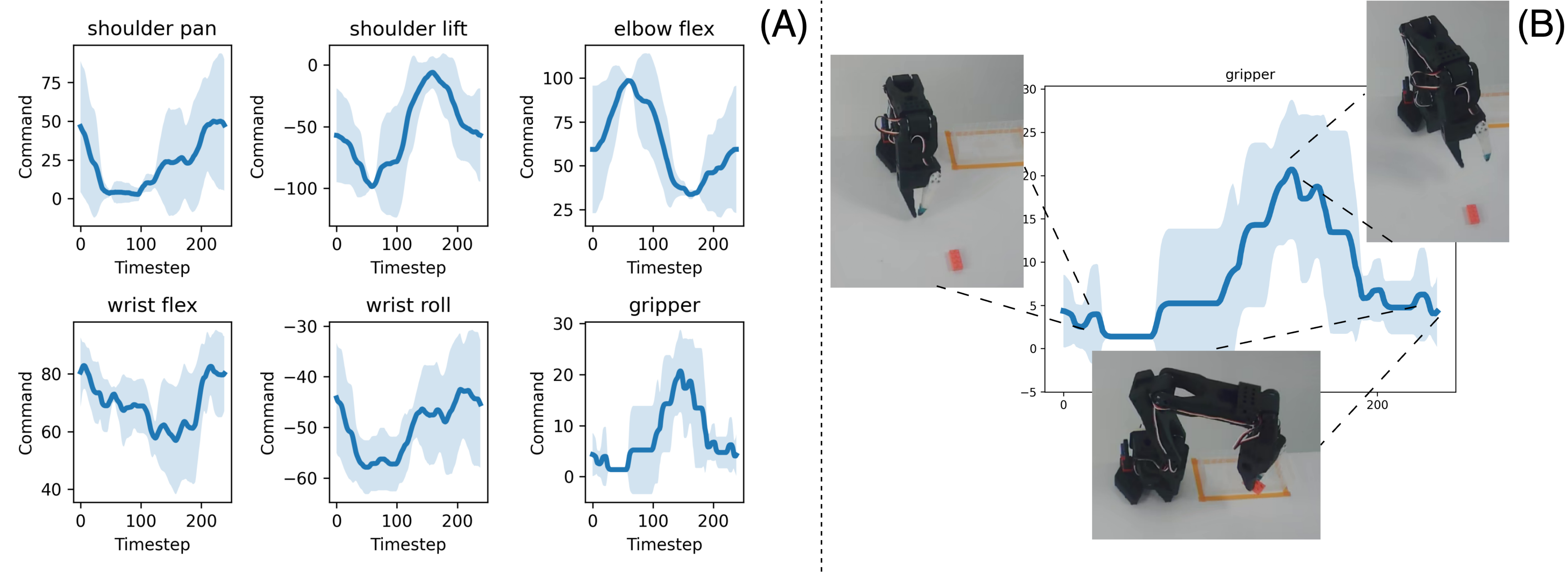}
    \caption{(A) Average (with standard deviation) evolution of the actuation levels over the first 5 recorded episodes in \url{lerobot/svla_so101_pickplace}. Proprioperceptive states provide invaluable to determine the robot's state during an episode. (B) Camera frames are also recorded alongside measurements on the robot's state, capturing information about the robot's interaction with its environment.}
    \label{fig:ch4-bc-trajectories}
\end{figure}

Learning from human demonstrations provides a pragmatic alternative to the RL pipeline discussed in Section~\ref{sec:learning-rl}.
Indeed, especially in real-world robotics, online exploration is typically \highlight{costly and potentially unsafe}, and designing (dense) reward signals is a \highlight{brittle and task-specific} process.
Further, even success detection itself often requires bespoke instrumentation, while episodic training demands reliable resets---all factors complicating training RL algorithms on hardware at scale.
Behavioral Cloning (BC) sidesteps these constraints by \highlight{casting control an imitation learning problem}, leveraging previously collected expert demonstrations to anchor the learned autonomous behavior.
Most notably, by \emph{learning-to-imitate}, autonomous systems naturally adhere to the objectives, preferences, and success criteria implicitly encoded in the data, which reduces early-stage exploratory failures and obviates hand-crafted reward shaping altogether.

\begin{figure}
    \centering
    \includegraphics[width=0.9\textwidth]{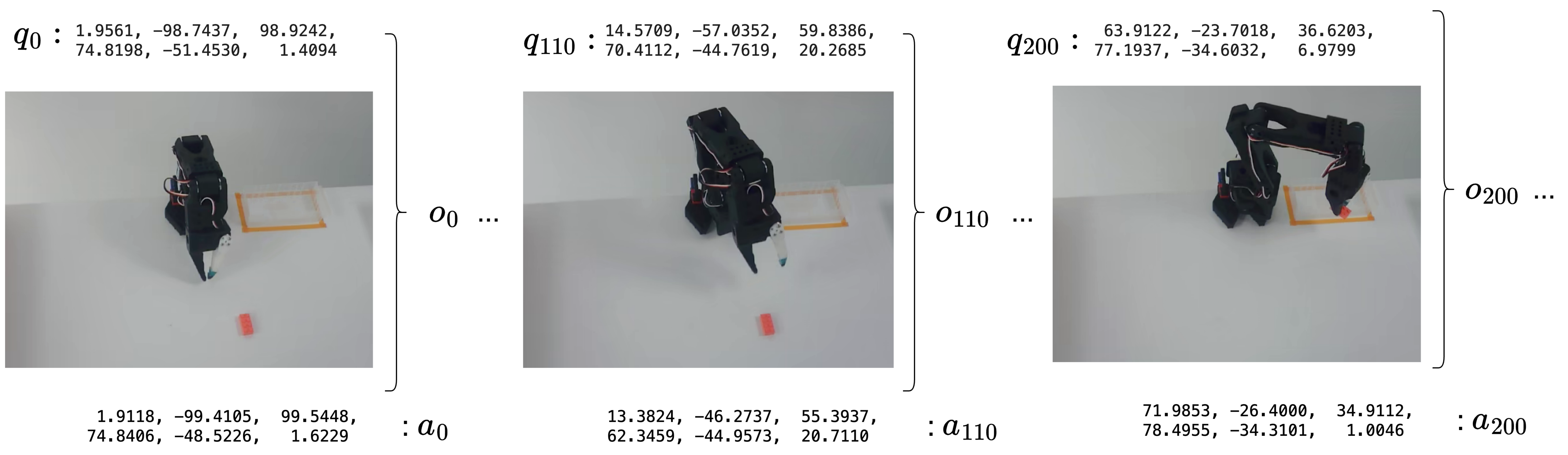}
    \caption{Sample observations and action pairs over the course of a given trajectory recorded in \url{lerobot/svla_so101_pickplace}. Observations, comprising of both proprioperceptive and visual information, are recorded alongside the configuration of a second, leader robot controlled by a human expert, providing complete information for regressing actions given observations.}
    \label{fig:ch4-observation-action-mapping}
\end{figure}

Formally, let \( \mathcal D = \{ \tau^{(i)} \}_{i=1}^N \) be a set of expert trajectories, with \( \tau^{(i)} = \{(o_t^{(i)}, a_t^{(i)})\}_{t=0}^{T_i} \) representing the \(i\)-th length-\(T_i\) trajectory in \( \mathcal D \), \(o_t \in \obsspace \) denoting observations (e.g., images and proprioception altogether), and \(a_t \in \actionspace \) the expert actions.
Typically, observations \( o \in \obsspace \) consist of both image and proprioperceptive information, while actions \( a \in \actionspace \) represent control specifications for the robot to execute, e.g. a joint configuration.
Note that differently from Section~\ref{sec:learning-rl}, in the imitation learning context \( \mathcal D \) denotes an offline dataset collecting \( N \) length-\( T_i \) reward-free (expert) human trajectories \( \tau^{(i)} \), and \emph{not} the environment dynamics.
Similarily, in this section \( \tau^{(i)} \) represent a length-\(T_i\) trajectory of observation-action pairs, which crucially \emph{omits entirely any reward} information.
Figure~\ref{fig:ch4-bc-trajectories} graphically shows trajectories in terms of the average evolution of the actuation on the 6 joints of a teleoperated SO-100 manipulator.
Notice how proprioperceptive states are captured jointly with camera frames over the course of the recorded episodes, providing a unified high-frame rate collection of both image and joint teleoperation data.
Figure~\ref{fig:ch4-observation-action-mapping} shows \( (o_t, a_t) \)-pairs for the same dataset, with the actions performed by the human expert illustrated alongside the corresponding observation.
In principle, (expert) trajectories \( \tau^{(i)} \) can have different lengths since demonstrations might exhibit multi-modal strategies to attain the same goal, resulting in multiple, different behaviors.

Behavioral Cloning (BC)~\citep{pomerleauALVINNAutonomousLand1988} aims at producing synthetic behaviors by learning the mapping from observations to actions, and in its most natural formulation can be effectively tackled as a \emph{supevised} learning problem, consisting of learning the (deterministic) mapping \(f: \obsspace \mapsto \actionspace, \ a_t = f(o_t) \) by solving
\begin{equation}\label{eq:loss-minimization-SL}
    \min_{f} \mathbb{E}_{(o_t, a_t) \sim p(\bullet)} \mathcal L(a_t, f(o_t)),
\end{equation}
given an arbitrary risk function \( \mathcal L:  \mathcal A \times \mathcal A \mapsto \mathbb{R}, \ \mathcal L (a, a^\prime) \).

Typically, the expert's joint observation-action distribution \( p: \obsspace \times \actionspace \mapsto [0,1] \) is assumed to be unknown, in keeping with a classic Supervised Learning (SL) framework\footnote{Throughout, we will adopt the terminology and notation for SL used in~\citet{shalev-shwartzUnderstandingMachineLearning2014}}.
However, differently from standard SL assumptions, the samples collected in \( \mathcal D \)---realizations of the underlying \( p \)---are \emph{not} i.i.d., as expert demonstrations are collected \emph{sequentially} in the form of trajectories.
In practice, this aspect can be partially mitigated by considering pairs in a non-sequential order---\emph{shuffling} the samples in \(\mathcal D \)---so that the expected risk under \( p \) can be approximated using MC estimates, although these estimates may in general be less accurate.
Another strategy to mitigate the impact of regressing over non-i.i.d. samples relies on the possibility of interleaving BC and data collection~\citep[DAgger]{rossReductionImitationLearning2011}, aggregating multiple datasets iteratively.
However, because we only consider the case where a single offline dataset \( \mathcal D \) of trajectories is available and no more data can be collected, DAgger falls out of our scope.

Despite the inherent challenges of learning from non-i.i.d. data, the BC formulation presents several operational advantages in robotics.
First, training happens offline and naturally accomodates for expert, demonstration data, hereby severily limiting exploration risks by preventing the robot from performing dangerous actions altogether, by anchoring action in imitation.
Second, reward design is entirely unnecessary in BC, as demonstrations already reflect human intent.
The absence of rewards also prevents the risk of misalignment and specification gaming (\emph{reward hacking}), otherwise inherent in purely reward-based RL~\citep{heessEmergenceLocomotionBehaviours2017}.
Third, because expert trajectories encode terminal conditions, success detection and resets are implicit in the dataset.
Finally, empirical evidence suggests the performance of BC scales naturally with growing corpora of demonstrations collected across tasks, embodiments, and environments.
Nonetheless, BC can, in principle, only reproduce behaviors that are at best as good as those of the demonstrator, and therefore offers no remedy for the suboptimal decisions that humans may enact.
This limitation is particularly problematic in sequential decision-making tasks where expert demonstrations are scarce--—either because data collection is costly or because human performance is inherently suboptimal. 
Yet, many robotics applications still benefit from relatively inexpensive pipelines for collecting high-quality human-generated trajectories, justifying the use of BC in such settings.

\begin{figure}
    \centering
    \includegraphics[width=0.8\textwidth]{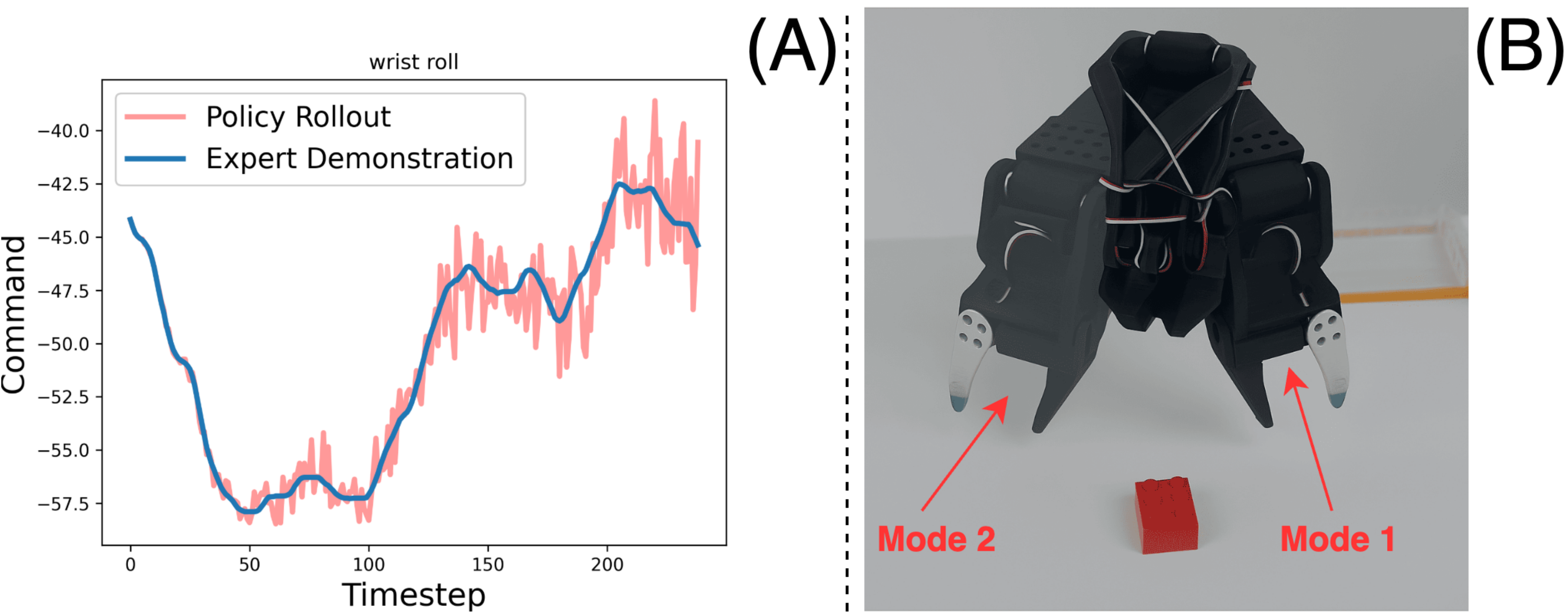}
    \caption{Point-wise policies suffer from limitations due to (A) covariate shifts and (B) poor approximation of multimodal demonstrations. (A) Small errors may drive the policy out of distribution, incuring in a vicious circle ultimately resulting in failure. (B) Both modes of reaching for a target object in the scene---either left or right-first---are equally as good and thus equally as likely to be present in a dataset of human demonstrations, ultimately resulting in multimodal demonstrations.}
    \label{fig:ch4-issues-with-bc}
\end{figure}

While conceptually elegant, \emph{point-estimate policies} \( f : \obsspace \mapsto \actionspace \) learned by solving eq.~\ref{eq:loss-minimization-SL} have been observed to suffer from (1) compounding errors~\citep{rossReductionImitationLearning2011} and (2) poor fit to multimodal distributions~\citep{florenceImplicitBehavioralCloning2022, keGraspingChopsticksCombating2020}.
Figure~\ref{fig:ch4-issues-with-bc} illustrates these two key issues related to learning \emph{explicit policies}~\citep{florenceImplicitBehavioralCloning2022}.
Besides sequentiality in \( \mathcal D \), compounding errors due to \emph{covariate shift} may also prove catastrophic, as even small \( \epsilon \)-prediction errors \( 0 < \Vert \mu(o_t) - a_t \Vert \leq \epsilon \) can quickly drive the policy into out-of-distribution states, incuring in less confident generations and thus compounding errors (Figure~\ref{fig:ch4-issues-with-bc}, left).
Moreover, point-estimate policies typically fail to learn \emph{multimodal} targets, which are very common in human demonstrations solving real-world robotics problems, as multiple trajectories can be equally as good towards the accomplishment of a goal (e.g., symmetric grasps, Figure~\ref{fig:ch4-issues-with-bc}, right).
In particular, unimodal regressors tend to average across modes, yielding indecisive or even unsafe commands~\citep{florenceImplicitBehavioralCloning2022}.
To address poor multimodal fitting,~\citet{florenceImplicitBehavioralCloning2022} propose learning the \emph{generative model} \( p(o, a) \) underlying the samples in \( \mathcal D \), rather than explicitly learning a prediction function \( f: a = f(o) \).

\subsection{A (Concise) Introduction to Generative Models}
Generative Models (GMs) aim to learn the stochastic process underlying the very generation of the data collected, and typically do so by fitting a probability distribution that approximates the unknown \emph{data distribution}, \( p \).
In keeping with the GM literature, \( p(x) \leftarrow \mathbb P(x), x \sim p \).
In the case of BC, the unknown data distribution \( p \) may represent the expert's joint distribution over \( (o, a) \)-pairs.
Thus, given a finite set of \( N \) pairs \(\mathcal D = \{ (o,a)_i \}_{i=0}^N\) available as an imitation learning target (and thus assumed to be i.i.d.), GMs seek to learn a \emph{parametric} distribution \( p_\theta(o,a) \) such that (1) new samples \( (o,a) \sim p_\theta(\bullet) \) resemble those stored in \( \mathcal D \), and (2) high likelihood is assigned to the \emph{observed} regions of the \emph{unobservable} \( p \).
Likelihood-based learning provides a principled training objective to achieve both goals, and it is thus extensively used in GMs~\citep{prince2023understanding}.

\subsubsection{Variational Auto-Encoders}

\begin{figure}
    \centering
    \includegraphics[width=0.8\textwidth]{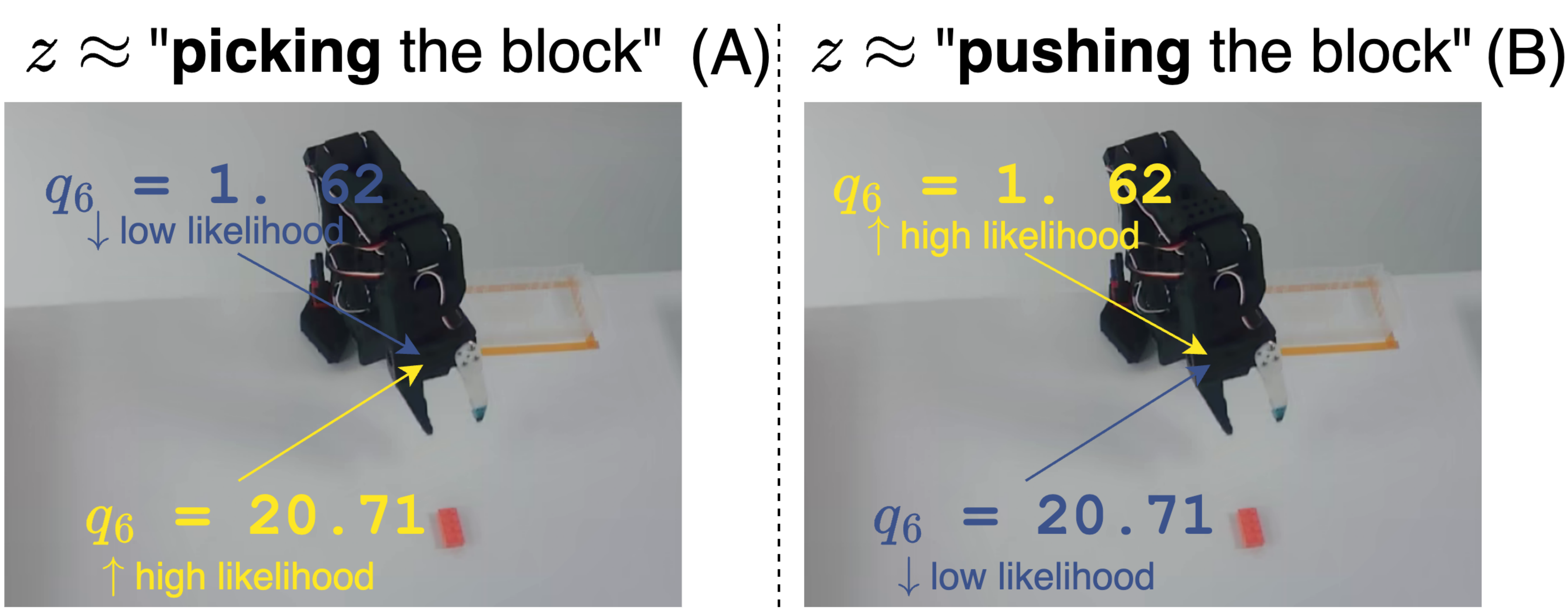}
    \caption{Intuitively, latent variable in a single latent model may contain information regarding the task being performed, which directly results in the likelihood of the same observation-action pair being different for two different tasks. When (A) picking a block the likelihood of a wide gripper's opening should be higher than narrower one, while it should be the opposite when (B) pushing the block.}
    \label{fig:ch4-task-effect-on-pairs}
\end{figure}

\begin{figure}
    \centering
    \includegraphics[width=0.9\textwidth]{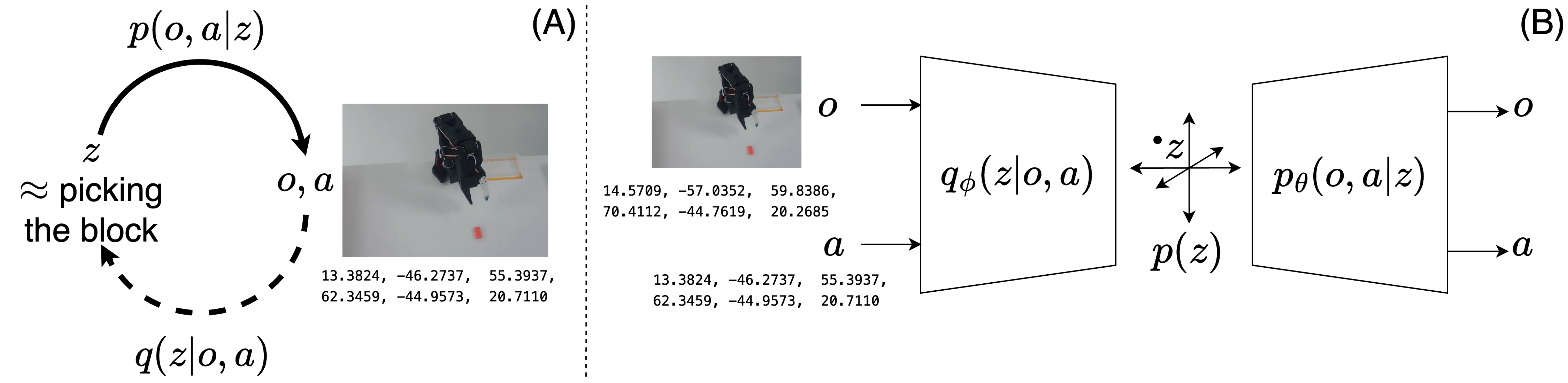}
    \caption{(A) The latent variable model in a robotics application regulates influence between observed (\(o,a) \) variables and an unobservable latent variable. (B) VAEs approximate exact latent variable models by means of variational inference. }
    \label{fig:ch4-latent-variable-model}
\end{figure}

A common inductive bias used in GM posits samples \( (o,a) \) are influenced from an unobservable latent variable \( z \in Z \), resulting in:
\begin{equation}\label{eq:BC-latent-variable}
    p (o,a) = \int_{\supp{Z}} p(o,a \vert z) p(z)
\end{equation}
Intuitively, in the case of observation-action pairs \( (o, a) \) for a robotics application, \( z \) could be interpreted as some high level representation of the underlying task being performed by the human demonstrator.
In such case, treating \( p(o,a) \) as a marginalization over \( \supp{Z} \) of the complete joint distribution \( p(o,a,z) \) natively captures the effect different tasks have on the likelihood of observation-action pairs.
Figure~\ref{fig:ch4-task-effect-on-pairs} graphically illustrates this concept in the case of a (A) picking and (B) pushing task, for which, nearing the target object, the likelihood of actions resulting in opening the gripper---the higher \( q_6 \), the wider the gripper's opening---should intuitively be (A) high or (B) low, depending on the task performed.
While the latent space \( Z \) typically has a much richer structure than the set of all actual tasks performed, eq.~\ref{eq:BC-latent-variable} still provides a solid framework to learn joint distribution conditioned on unobservable yet relevant factors.
Figure~\ref{fig:ch4-latent-variable-model} represents this latent-variable framework in the context of a robotics application: the true, \( z \)-conditioned generative process assigns \emph{likelihood} \( p((o,a) \vert z) \) to the single \( (o,a) \)-pair.
Using Bayes' theorem, one can reconstruct the \emph{posterior} distribution on \( \supp{Z} \), \( q_\theta(z \vert o,a) \) from the likelihood \( p_\theta(o,a \vert z) \), \emph{prior} \( p_\theta(z) \) and \emph{evidence} \( p_\theta(o,a) \).
VAEs approximate the latent variable model presented in eq.~\ref{eq:BC-latent-variable} using an \emph{approximate posterior} \(q_\phi(z \vert o,a) \) while regressing parameters for a parametric likelihood, \( p_\theta(o,a \vert z) \) (Figure~\ref{fig:ch4-latent-variable-model}).

Given a dataset \( \mathcal D \) consisting of \( N \) i.i.d. observation-action pairs, the log-likelihood of all datapoints under \( \theta \) (in Bayesian terms, the \emph{evidence} \( p_\theta(\mathcal D)\)) can be written as:
\begin{align}
    \log p_\theta(\mathcal D) &= \log \sum_{i=0}^N p_\theta ((o,a)_i) \label{eq:evidence-definition-1}\\
                              &= \log \sum_{i=0}^N \int_{\supp{Z}} p_\theta((o,a)_i \vert z) p(z) \label{eq:evidence-definition-2}\\
                              &= \log \sum_{i=0}^N \int_{\supp{Z}} \frac{q_\theta(z \vert (o,a)_i)}{q_\theta(z \vert (o,a)_i)} \cdot p_\theta((o,a)_i \vert z) p(z) \label{eq:evidence-definition-3}\\
                              &= \log \sum_{i=0}^N \mathbb E_{z \sim q_\theta(\bullet \vert (o,a)_i)} \left[ \frac{p(z)}{q_\theta(z \vert (o,a)_i)} \cdot p_\theta((o,a)_i \vert z) \right], \label{eq:evidence-definition}
\end{align}
where we used eq.~\ref{eq:BC-latent-variable} in eq.~\ref{eq:evidence-definition-1}, multiplied by \(1 = \frac{q_\theta(z \vert (o,a)_i)}{q_\theta(z \vert (o,a)_i)} \) in eq.~\ref{eq:evidence-definition-2}, and used the definition of expected value in eq.~\ref{eq:evidence-definition}.

In the special case where one assumes distributions to be tractable, \( p_\theta (\mathcal D) \) is typically tractable too, and \(\max_\theta \log p_\theta(\mathcal D) \) provides a natural target for (point-wise) infering the unknown parameters \( \theta \) of the generative model.
Unfortunately, eq.~\ref{eq:evidence-definition} is rarely tractable when the distribution \( p \) is modeled with approximators such as neural networks, especially for high-dimensional, unstructured data.

In their seminal work on Variational Auto-Encoders (VAEs),~\citet{kingma2013auto} present two major contributions to learn complex latent-variable GMs from unstructured data, proposing (1) a tractable, variational lower-bound to eq.~\ref{eq:evidence-definition} as an optimization target to jointly learn likelihood and posterior and (2) using high-capacity function approximators to model the likelihood \(p_\theta(o,a\vert z)\) and (approximate) posterior distribution \( q_\phi(z \vert o,a) \approx q_\theta(z \vert o,a) \).

In particular, the lower bound on eq.~\ref{eq:evidence-definition} (Evidence LOwer Bound, \emph{ELBO}) can be derived from eq.~\ref{eq:evidence-definition} applying Jensen's inequality---\(\log \mathbb{E}[\bullet] \geq \mathbb{E} [\log (\bullet)] \)---yielding:
\begin{align}
    \log p_\theta(\mathcal D) &\geq \sum_{i=0}^{N} \left(
            \mathbb{E}_{z \sim q_\theta(\bullet \vert (o,a)_i)} \big[ \log p_\theta((o,a)_i \vert z) \big]
            + \mathbb{E}_{z \sim q_\theta(\bullet \vert (o,a)_i)} \left[ \log \left( \frac{p(z)}{q_\theta(z \vert (o,a)_i)} \right) \right]
        \right) \\
        &= \sum_{i=0}^{N} \left(
            \mathbb{E}_{z \sim q_\theta(\bullet \vert (o,a)_i)} \big[ \log p_\theta((o,a)_i \vert z) \big]
        - \DKL \big[ q_\theta(z \vert (o,a)_i) \Vert p(z) \big]
        \right) \label{eq:ELBO-intractable}
\end{align}

The true, generally intractable, posterior \( q_\theta (z \vert o,a) \) prevents computing both the expectation and KL divergence terms in eq.~\ref{eq:ELBO-intractable}, and therefore~\citet{kingma2013auto} propose deriving the ELBO using an \emph{approximate} posterior \( q_\phi(z \vert o,a) \), resulting in the final, tractable, ELBO objective,
\begin{align}
\text{ELBO}_{\mathcal D}(\theta, \phi) = \sum_{i=0}^{N} \left(
            \mathbb{E}_{z \sim q_\phi(\bullet \vert (o,a)_i)} \big[ \log p_\theta((o,a)_i \vert z) \big]
        - \DKL \big[ q_\phi(z \vert (o,a)_i) \Vert p(z) \big]
        \right)
        \label{eq:ELBO}
\end{align}
From Jensen's inequality, maximizing ELBO results in maximizing the log-likelihood of the data too, thus providing a natural, tractable optimization target.
Indeed, expectations can be estimated using MC estimates from the learned distributions in eq.~\ref{eq:ELBO}, while the KL-divergence term can typically be computed in closed-form (1) modeling  \(q_\phi \) as a Gaussian \(q_\phi(z \vert o,a) = \mathcal N\big(\mu_\phi(o,a), \Sigma_\phi(o,a) \big) \) with learned mean vector \( \mu_\phi(o,a) \) and learned variance-covariance matrix \( \Sigma_\phi(o,a) \) and (2) imposing a standard Gaussian prior on the latent space, \( p(z) = \mathcal N(\mathbf{0}, \mathbf{I}) \).

An intuitive explanation of the learning dynamics of VAEs can be given considering the equivalent case of \emph{minimizing the negative ELBO}, which admits the particularly interpretable factorization (considering, without loss of generality, only one \( (o,a) \sim \mathcal D \)):
\begin{align}
\min_{\theta, \phi} - \text{ELBO}_{\mathcal (o,a) \sim \mathcal D}(\theta, \phi) &= \min_{\theta, \phi}\mathbf{L^{\text{rec}}}(\theta) + \mathbf{L^{\text{reg}}}(\phi), \label{eq:VAE-min-neg-ELBO}\\
\mathbf{L^{\text{rec}}}(\theta) &= \mathbb{E}_{z \sim q_\phi(\bullet \vert o,a}) \big[ \log p_\theta(o,a \vert z) \big] \label{eq:VAE-Lrec} \\
\mathbf{L^{\text{reg}}}(\phi) &= \DKL \big[ q_\phi(z \vert o,a) \Vert p(z) \big]. \label{eq:VAE-Lreg}
\end{align}

For any given \((o,a) \) pair, the expected value term in eq.~\ref{eq:VAE-Lrec} is typically computed via MC estimates, resulting in
\[ 
-\mathbb{E}_{z \sim q_\phi(\bullet \vert o,a)} \big[ \log p_\theta(o,a \vert z) \big] = \mathbf{L^{\text{rec}}} \approx - \frac{1}{n} \sum_{i=0}^n \log p_\theta(o,a \vert z_i).
\]
Assuming \( p_\theta(o,a \vert z) \) to be parametrized with an isotropic Gaussian distribution with mean \(\mu_\theta (z) \in \mathbb R^d \) and variance \( \sigma^2 \), the log-likelihood thus simplifies to:
\[
\log p(o,a \vert z_i) = -\frac{1}{2\sigma^{2}} \big \Vert (o,a)-\mu_\theta(z_i) \big\Vert_2^2 -\frac{d}{2}\log(2\pi \sigma^{2}) \implies \mathbf{L^\text{rec}} \approx \frac {1}{n} \sum_{i=0}^n \big\Vert (o,a) - \mu_\theta(z_i) \big \Vert^2_2
\]
In practice, it is common to approximate the learned likelihood \( p_\theta(o,a \vert z) \) with a parametric distribution (e.g., Gaussian) whose parameters are given by a learned coefficient vector derived from \( \mu_\theta(z), \ z \sim p(\bullet) \). 
Under this formulation, learning a VAE amounts to (1) \emph{reconstructing} the examples in \( \mathcal{D} \) by minimizing (1) the reconstruction loss \( \mathbf{L^{\text{rec}}}\)---a standard \emph{supervised learning} objective for regression---while (2) \emph{regularizing} the latent representation by minimizing \( \mathbf{L^{\text{reg}}} \). 
The latter enforces information compression, since with the common prior choice \( p(z) = \mathcal{N}(\mathbf{0}, \mathbf{I})\) in eq.~\ref{eq:VAE-Lreg}, the regularizer constrains the posterior and thereby limits the expressivity of \( q_\phi(z \vert o,a) \).

\subsubsection{Diffusion Models}
\begin{figure}
    \centering
    \includegraphics[width=0.5\textwidth]{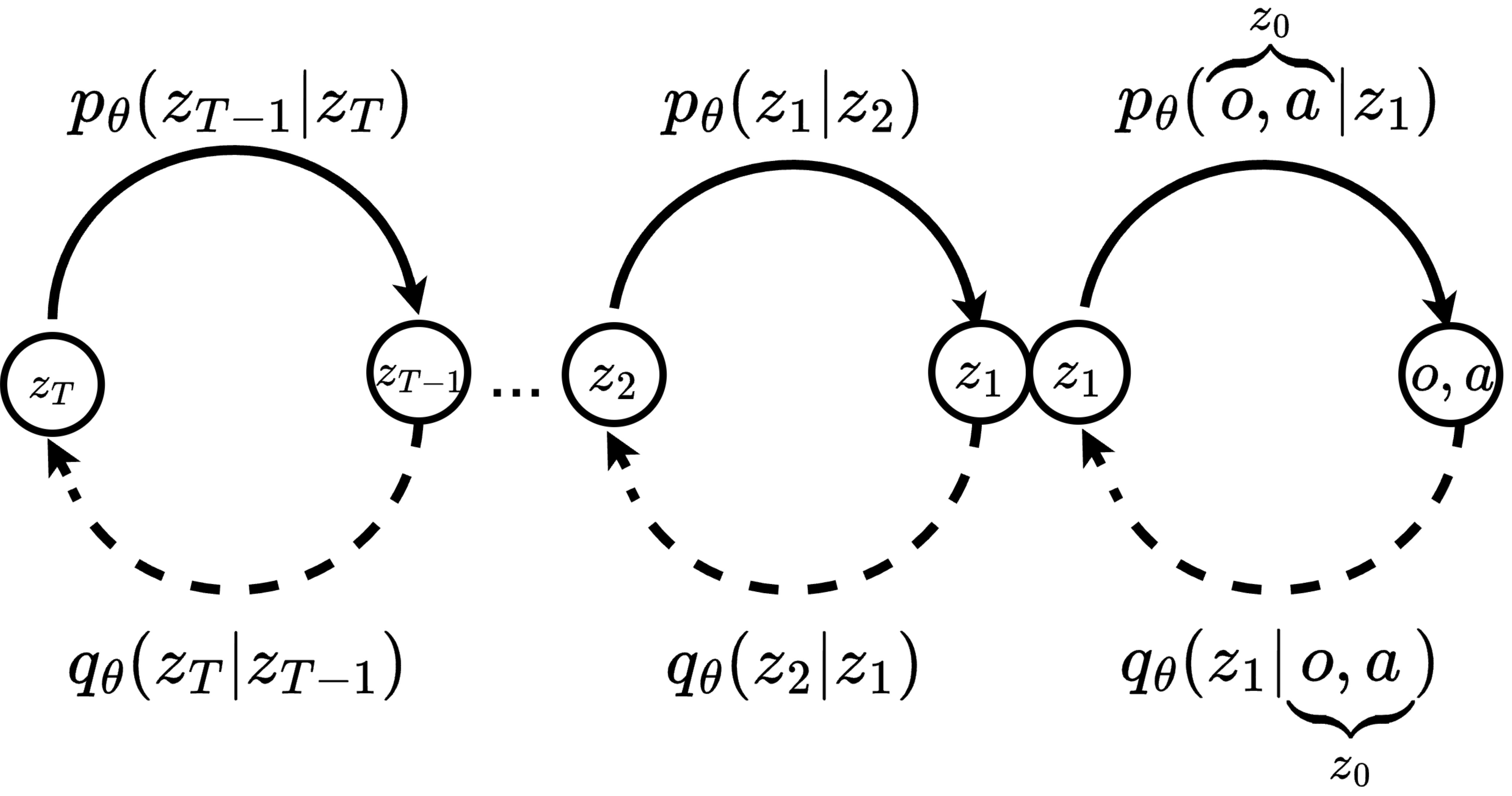}
    \caption{HMLV models posit the data generation process is influenced by a stack of Markov-dependent latent variables, with samples from the posterior distribution being progressively higher up in the hierarchy.}
    \label{fig:ch4-many-latents}
\end{figure}

VAEs approximate probability distributions via a \emph{single} latent variable model, assuming the underlying unknown distribution can be factored according to eq.~\ref{eq:BC-latent-variable}, and solve the variational-inference problem of jointly learning the likelihood \( p_\theta \) and (approximate) posterior \( q_\phi \) for such model.
In that, the unknown data distribution \( p(o,a) \) is effectively approximated via \( \int_Z p(z) p_\theta(o,a \vert z) \), and the underlying generative process reproduced by (1) sampling a latent variable and (2) learning to decode it into a high-likelihood sample under the (unknown) \( p(o,a) \).
Diffusion Models (DMs)~\citep{hoDenoisingDiffusionProbabilistic2020} are another class of GMs which treat the similar problem of approximating an underlying unknown data distribution---\emph{variational inference}---by \emph{partially} extending VAEs to the case where \emph{multiple} latent variables influence each other and the generative process underlying \(o,a\) itself.
In particular, DMs posit the generative process can be decomposed to a series of piece-wise (Markovian) interactions between (latent) variables (Figure~\ref{fig:ch4-many-latents}), resulting in
\begin{align}
    p(\underbrace{o,a}_{= z_0}) &= \int_{\supp{Z_0}} \int_{\supp{Z_1}} \hdots \int_{\supp{Z_T}} p(z_0, z_1, \dots z_T) \label{eq:BC-multi-latent-model-1} \\ 
    p(z_0, z_1, \dots z_T) &= p(z_T) \prod_{t=1}^{T} p(z_{t-1} \vert z_t), \label{eq:BC-multi-latent-model-2}
\end{align}
where we explicitly showed the marginalization over the multiple latents in eq.~\ref{eq:BC-multi-latent-model-1}, and used the law of conditional probability and Markov property in eq.~\ref{eq:BC-multi-latent-model-2}.
Also, for ease of notation, we will refer to observation-action pairs \( o,a \) as \( z_0 \).

Similar to VAEs, it is generally not possible to assign an \emph{exact} interpretation to the latent variables. 
Nevertheless, a reasonable application-driven intuition is that Hierarchical Markov Latent Variable (HMLV) models, by capturing hierarchical and decoupled interactions among latent variables, can reflect the different resolutions at which conditioning factors intervene. 
For example, in a robotics setting, one might naturally distinguish between high-level trajectory planning (higher up in the hierarchy, \(t \to T\)) and fine-grained motion adjustments (closer to empirical observations, \(t \to 0\)).
In that, HMLV models thus provide a framework to perform variational inference via multiple, sequential sampling steps from different higher level distributions instead of approximating the generative process with a single-latent variable model.
DMs are a particular instantiation of HMLV models for which the posterior is fixed to \( q( z_t \vert z_{t-1}) = \mathcal N(z_t \sqrt{1-\beta_t}, \beta_t \mathbf{I}) \), for a given \( \beta_t \in \mathbb R^+ \). 
In practice, \( \beta_t \) is used to iteratively reduce the signal-to-noise ratio along the latents' hierarchy, similarily to how a diffusion process influences the information of a physical system.

Just like VAEs, DMs attemp to learn to reproduce an underlying data distribution \( p (o,a) \) given a collection of i.i.d. samples approximating the model posited to have generated the data in the first place (eq.~\ref{eq:BC-multi-latent-model-1}).
Similarily to VAEs, DMs approximate the process of sampling from the unknown \( p(o,a) \) by (1) sampling from an easy-to-sample distribution (e.g., Gaussian) and (2) learning to reconstruct high-likelihood samples under the unknown distribution.
However, in stark contrast with VAEs, the easy-to-sample distribution contains \emph{no mutual information} regarding the data distribution \( p(o,a) \).
Crucially, as no information from the sample \( (o,a) \) (denoted as \( z_0 \equiv (o,a) \) for simplicity of notation) is assumed to be propagated throughout the chain of latents, the posterior \( q(z_t \vert z_{t-1})\) assumes a relatively amicable structure in DMs, reducing complexity.
The \emph{true} likelihood \( p(z_{t-1} \vert z_t) \) is instead typically approximated using the parametrization \(  p_\theta (z_{t-1} \vert z_t) \).
In that, the information contained in the unknwon data distribution is \emph{reconstructed} via a process in which samples from a fixed distribution are iteratively turned into (ideally) high-likelihood samples under \( p(o,a) \)---a process referred to as \emph{denoising}.

Under such model, we can express the log-likelihood of an arbitrary sample \( z_0 \) as:
\begin{align}
    \log p_\theta (z_0) &= \log \int_{\supp{Z_1} \times \supp{Z_2} \times \dots \times \supp{Z_T}} p_\theta(\underbrace{z_0, z_1, z_2, \dots z_T}_{z_{0:T}}) \\
    &= \log \int_{\supp{Z_{1:T}}} \frac{p_\theta(z_{0:T}) \cdot q(z_{1:T} \vert z_0)}{q(z_{1:T} \vert z_0)} \label{eq:diffusion-1} \\
    &= \log \mathbb{E}_{z_{1:T} \sim q(\bullet \vert z_0)} \bigg[ \frac{p_\theta(z_{0:T})}{q(z_{1:T} \vert z_0)} \bigg] \\
    &\geq \mathbb{E}_{z_{1:T} \sim q(\bullet \vert z_0)} \bigg[ \log \frac{p_\theta(z_{0:T})}{q(z_{1:T} \vert z_0)} \bigg] \label{eq:diffusion-jensen} \\
    &= \mathbb{E}_{z_{1:T} \sim q(\bullet \vert z_0)} \bigg[ \log \frac{p(z_T) \prod_{t=1}^{T} p_\theta (z_{t-1} \vert z_t)}{\prod_{t=1}^T q(z_t \vert z_{t-1})} \bigg] \label{eq:diffusion-2} \\
    &= \mathbb{E}_{z_{1:T} \sim q(\bullet \vert z_0)} \bigg[ \log \frac{p(z_T) \cdot p_\theta (z_0 \vert z_1) \prod_{t=2}^{T} p_\theta (z_{t-1} \vert z_t)}{q(z_T \vert z_{T-1}) \prod_{t=1}^{T-1} q(z_t \vert z_{t-1})} \bigg] \label{eq:diffusion-3} \\
    &= \mathbb{E}_{z_{1:T} \sim q(\bullet \vert z_0)} \bigg[ \log \frac{p(z_T) \cdot p_\theta (z_0 \vert z_1) \prod_{t=1}^{T-1} p_\theta (z_{t} \vert z_{t+1})}{q(z_T \vert z_{T-1}) \prod_{t=1}^{T-1} q(z_t \vert z_{t-1})} \bigg] \label{eq:diffusion-4} \\
    &= 
        \mathbb{E}_{z_{1:T} \sim q(\bullet \vert z_0)} \bigg[ \log \frac{p(z_T) \cdot p_\theta (z_0 \vert z_1)}{q(z_t \vert z_{t-1})} \bigg] + 
        \mathbb{E}_{z_{1:T} \sim q(\bullet \vert z_0)} \bigg[ \log \prod_{t=1}^{T-1} \frac{p_\theta (z_{t} \vert z_{t+1})}{q(z_t \vert z_{t-1})}\bigg]
    \label{eq:diffusion-5} \\
    &=
        \mathbb{E}_{z_{1:T} \sim q(\bullet \vert z_0)} \big[ \log  p_\theta (z_0 \vert z_1) \big] + 
        \mathbb{E}_{z_{1:T} \sim q(\bullet \vert z_0)} \bigg[ \log \frac{p (z_T)}{q(z_T \vert z_{T-1})} \bigg] +
        \sum_{t=1}^{T-1} \mathbb{E}_{z_{1:T} \sim q(\bullet \vert z_0)} \bigg[ \log \frac{p_\theta (z_{t} \vert z_{t+1})}{q(z_t \vert z_{t-1})}\bigg]
    \label{eq:diffusion-6} \\
    &= 
        \mathbb{E}_{z_1 \sim q(\bullet \vert z_0)} \big[ \log  p_\theta (z_0 \vert z_1) \big] + 
        \mathbb{E}_{z_{T-1:T} \sim q(\bullet \vert z_0)} \bigg[ \log \frac{p (z_T)}{q(z_T \vert z_{T-1})} \bigg] +
        \sum_{t=1}^{T-1} \mathbb{E}_{z_{t-1:t+1} \sim q(\bullet \vert z_0)} \bigg[ \log \frac{p_\theta (z_{t} \vert z_{t+1})}{q(z_t \vert z_{t-1})}\bigg]
    \label{eq:diffusion-expectation-indices} \\
    &= \mathbb{E}_{z_1 \sim q(\bullet \vert z_0)} \log p_\theta (z_0 \vert z_1) - \mathbb{E}_{z_{T-1} \sim q(\bullet \vert z_0)} \big[ \DKL (q(z_T \vert z_{T-1}) \Vert p(z_T) ) \big] \label{eq:diffusion-likelihood} \\
    &- \sum_{t=1}^{T-1} \mathbb{E}_{(z_{t-1}, z_{t+1}) \sim q(\bullet \vert z_0)} \big[ \DKL (q(z_t \vert z_{t-1}) \Vert p_\theta(z_t \vert z_{t+1}) ) \big], \notag
\end{align}
where we: used eq.~\ref{eq:BC-multi-latent-model-1} and multiplied by \( 1 = \tfrac{q(z_{1:T} \vert z_0)}{q(z_{1:T} \vert z_0)} \) in eq.~\ref{eq:diffusion-1}; used Jensen's inequality in eq.~\ref{eq:diffusion-jensen}; used the law of conditional probability for both numerator and denominator in eq.~\ref{eq:diffusion-2}; stepped forward and backward the products in the numerator and denominator products in eq.~\ref{eq:diffusion-3}, respectively; reindexed the product terms in eq.~\ref{eq:diffusion-4}; removed out-of-expectation variables in eq.~\ref{eq:diffusion-expectation-indices}; used the defintion of KL-divergence in eq.~\ref{eq:diffusion-likelihood}.
In turn, eq.~\ref{eq:diffusion-likelihood} provides an optimization target to \emph{learn} \( p_\theta \) solving \( \max_\theta \log p_\theta (\mathcal D) \).

In their seminal work on using DMs for variational inference,~\citet{hoDenoisingDiffusionProbabilistic2020} introduce major contributions regarding solving \( \min_\theta -\log p_\theta(z_0) \).
In particular,~\citet{hoDenoisingDiffusionProbabilistic2020} exclusively adopt a \emph{fixed, isotropic Gaussian posterior} in the form of \( q(z_t \vert z_{t-1}) = \mathcal{N}(\sqrt{1-\beta_t}z_{t-1}, \beta_t \mathbf I) \).
The choice of adopting Gaussians has profound implications on the generative process modeled. 
Indeed, under the (mild) assumption that the variance is sufficiently small \( \beta_t \leq \eta, \eta \in \mathbb R^+ \),~\citet{sohnLearningStructuredOutput2015} proved that the likelihood \( p(z_{t-1} \vert z_t) \) is Gaussian as well, which allows for the particularly convenient parametrization of the approximate likelihood \( p_\theta (z_{t-1} \vert z_t) = \mathcal N(\mu_\theta(z_t, t), \Sigma_\theta(z_t,t)), \ t \in [1,T] \), as well as for closed-form tractability of the KL-divergence terms in eq.~\ref{eq:diffusion-likelihood}.
Further, the posterior's structure also enables the analytical description of the distribution of the \( t\)-th latent variable, \( q(z_t \vert z_0) = \mathcal N (\sqrt{\bar{\alpha}_t}z_0, (1-\bar{\alpha}_t) \mathbf{I}) \), with \( \alpha_t = 1-\beta_t, \ \bar \alpha_t = \prod_{k=1}^t \alpha_k \), conveniently preventing iterative posterior sampling simplifying computing eq.~\ref{eq:diffusion-likelihood}.
It follows:
\begin{align}
    \nabla_\theta \log p_\theta (z_0) = \mathbb E_{z_1 \sim q(\bullet \vert z_0)} \nabla_\theta \log p_\theta (z_0 \vert z_1) - \sum_{t=1}^{T-1} \mathbb E_{z_{t-1}, z_{t+1} \sim q(\bullet \vert z_0)} \nabla_\theta \DKL (q(z_t \vert z_{t-1}) \Vert p_\theta(z_t \vert z_{t+1}), \label{eq:diffusion-likelihood-gradient}
\end{align}
where the former term is equivalent to the reconstruction term in eq.~\ref{eq:VAE-min-neg-ELBO} and the latter term can be obtained in closed form.

\begin{figure}
    \centering
    \includegraphics[width=0.9\textwidth]{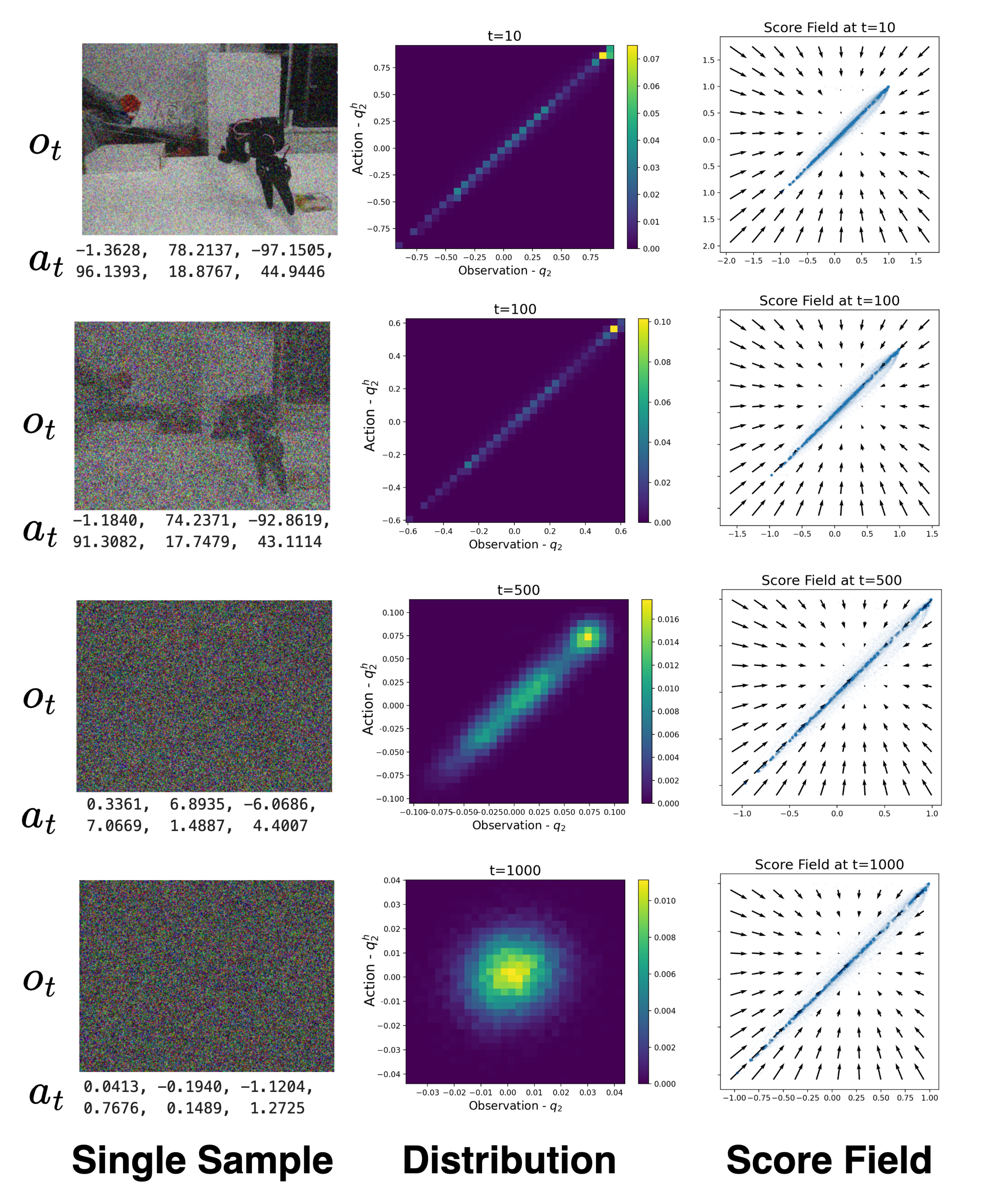}
    \caption{DMs iteratively corrupt samples (left) from an unknown distribution into a quasi-standard Gaussian (center), learning the displacement field (right) that permits to reconstruct samples from the unknown target distribution by iteratively denoising samples of a tractable, easy-to-sample distribution.}
    \label{fig:diffusion-robot-actions}
\end{figure}

\begin{figure}
    \centering
    \includegraphics[width=0.3\textwidth]{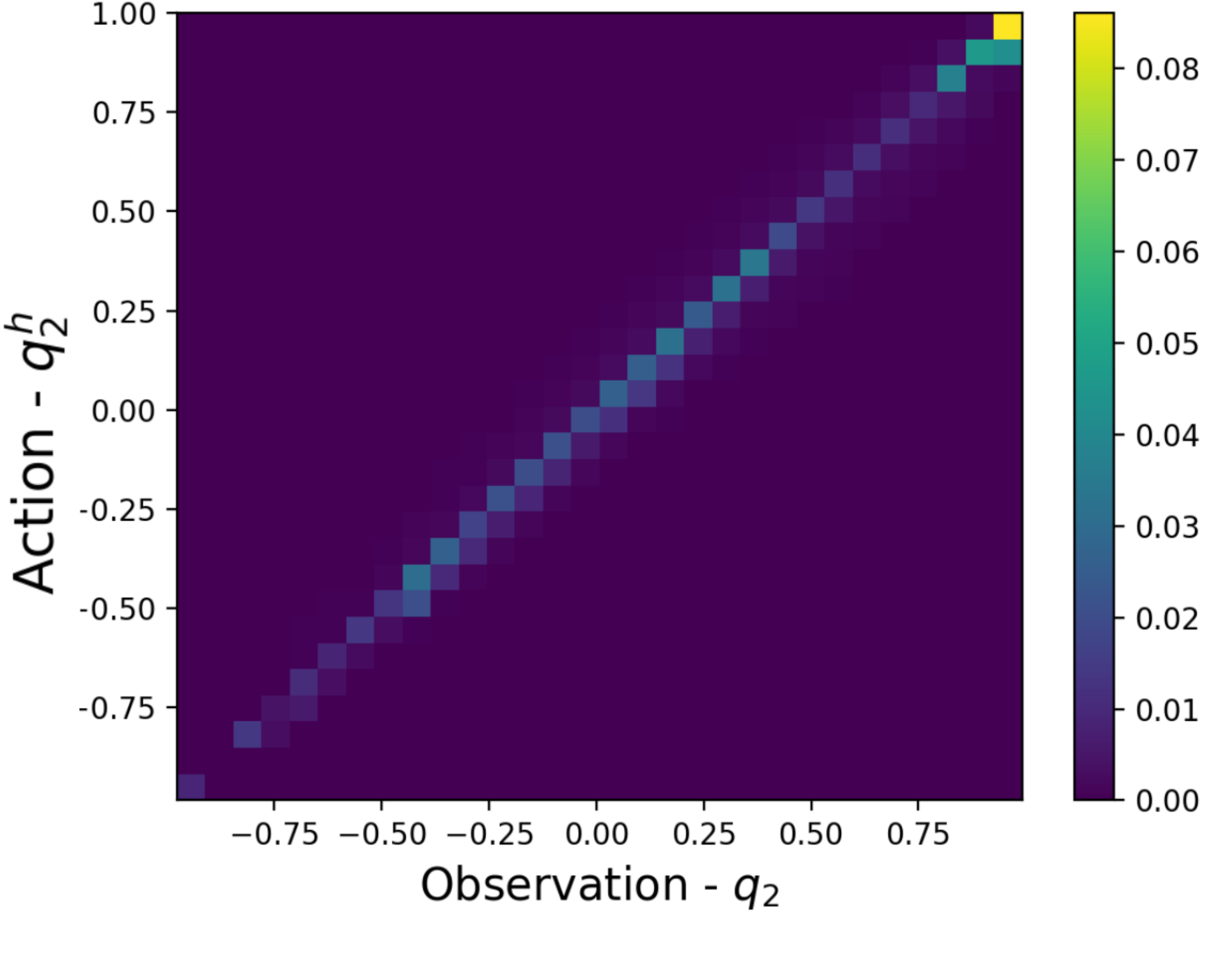}
    \caption{A joint action-observation distribution, in the simplified case where the observation is the elbow-flex actuation in a SO-100, and the action is the recorded position for the same joint from the teleoperator arm. The motion recorded being teleoperated, the points distribute along a the diagonal.}
    \label{fig:ch4-action-vs-observation-distribution}
\end{figure}

Besides mathematical tractability of eq.~\ref{eq:diffusion-likelihood-gradient}, adopting Gaussian posteriors allows for a particularly intuitive interpretation of the training dynamics of DMs~\citep{permenterInterpretingImprovingDiffusion2024}. 
As the hierarchical latent variables are repeatedly corrupted by applying increasingly more Gaussian noise, they progressively lose information about the original (unknown) sample \( z_0 \), converging toward a standard Gaussian which eventually contains no information at all (Figure~\ref{fig:diffusion-robot-actions}).
Figure~\ref{fig:diffusion-robot-actions} illustrates this process on a simplified, bidimensional observation-action distribution, where we considered \( o=q_2 \) and \( a=q^h_2 \), with \( q_2 \) denoting the robot's \emph{elbow flex} actuation and \( q^h_2 \) the corresponding human teleoperator's elbow flex.
Because the recorded behavior is teleoperated, measurements mostly distribute along the line \( a = o + \eta, \eta \sim N(0,1) \), with \( \eta \)-variability accouting for minor control inconsistencies (Figure~\ref{fig:ch4-action-vs-observation-distribution}).
Notice how corrupted samples distribute differently from the most reasonable structure \( a \simeq o \), further underscoring how diffusion corrupts both the individual samples and the global distribution (Figure~\ref{fig:diffusion-robot-actions}, left and center).
In this, using Gaussian posteriors---i.e., adding Gaussian noise---effectively simulates a \emph{Brownian motion} for the elements in the distribution's support (in Figure~\ref{fig:diffusion-robot-actions}, \( \obsspace \times \actionspace \)), whereby information \emph{diffuses away} from the samples.
Comparing the diffused samples to the original data points, one can derive an estimate of the total displacement induced by the diffusion process, and, under the assumption that the likelihood of the totally diffused samples is low under the original unknown data distribution, one can effectively approximate the unkwown distribution by \emph{learning to reverse} such displacement.
This key intuition allows to write a simplified training objective\footnote{See~\citet["Three equivalent interpretations"]{luoUnderstandingDiffusionModels2022} for a complete derivation}:
\begin{align}\label{eq:diffusion-simplified-loss}
    \mathcal L(\theta) = \mathbb{E}_{t, z_0, \epsilon} \big[
        \Vert \epsilon - \epsilon_\theta(\sqrt{\bar \alpha_t} z_0 + \epsilon \sqrt{1 - \bar \alpha_t}, t) \Vert^2 \big], \quad t \sim \mathcal{U}(\{1,\dots,T\}), \quad
        z_0 \sim \mathcal{D}, \quad
        \epsilon \sim \mathcal{N}(\mathbf{0},\mathbf{I}).
\end{align}

In this simplified (minimization) objective, the optimization process differs from eq.~\ref{eq:diffusion-likelihood} in that, rather than maximizing \( p_\theta \) directly, the parameters \( \theta \) of the pairwise likelihood \( p_\theta(z_{t-1} \vert z_t) \) are adjusted to \emph{predict the total displacement} \( \epsilon \) for a randomly long (\( t \sim \mathcal{U}(\{1,\dots,T\}) \)) diffusion process starting from a sample of the target distribution.

By learning the total displacement from a generally, uninformative corrupted sample obtained diffusing information and a sample from an unknown distribution~\citet{hoDenoisingDiffusionProbabilistic2020} show that one can approximate the underlying distribution reversing the displacement, \emph{denoising} samples.
Interestingly, under the hypothesis that real-world data belongs to a single, higher-dimensional manifold (Manifold Hypothesis),~\citet{permenterInterpretingImprovingDiffusion2024} show that diffusion learns the gradient of a distance function from any off-point manifold (such as perturbed, uniformative samples), and the data manifold itself.
Following this gradient---i.e., denoising a sample from an uninformative distribution---corresponds to projecting back into the manifold, yielding a procedure to sample from unknown distributions by means of Euclidean projection.
Indeed, under the assumption that \(p_\theta (z_{t-1} \vert z_t) \) is Gaussian, sampling \(z_{t-1} \sim p_\theta(\bullet \vert z_{t}) \) corresponds to computing:
\begin{align}
    z_{t-1} = \frac{1}{\sqrt{\alpha_t}} \left( z_t - \frac{\beta_t}{\sqrt{1 - \bar\alpha_t}} \epsilon_\theta(z_t, t) \right) + \sigma_t \epsilon, \quad \epsilon \sim \mathcal N(\mathbf{0}, \mathbf{I}), \label{eq:diffusion-denoising-definition}
\end{align}
thus showing that the lower-level latent variables in a DM can be obtained by iteratively removing noise from the one-step higher order variable, using the noise regressor \( \epsilon_\theta(z_t, t)\) learned minimizing eq.~\ref{eq:diffusion-simplified-loss}.

\subsubsection{Flow Matching}
\label{sec:ch4-flow-matching}
 
The posterior parametrization adopted by DMs proved traditionally effective, yet it raised concerns circa its \emph{efficiency} at inference time, where a possibly large number (hundreds) of compute-expensive denoising steps are needed in order to recover a sample from the target distribution.
Flow Matching (FM)~\citep{lipmanFlowMatchingGenerative2023} extends DMs to the general case of arbitrary likelihood and posteriors, and in this defines a superseding class of GMs providing a unified framework for learning \emph{continuous transformations} between distributions, encompassing and generalizing DMs.
Instead of a \emph{stochastic, discrete, multi-step} denoising process, FM aims to learn a \emph{deterministic, continuous, differentiable flow} \( \psi: [0,1] \times Z \mapsto Z \), formalized starting from a (possibly time-dependent) vector field \( v: [0,1] \times Z \mapsto Z \) \emph{transporting over time} samples from a simple prior distribution \( p_0 \)---e.g., a standard Gaussian---to a more complex, typically unknown data distribution \( p_1 \).
In this, FM accomodates for arbitrary intermediate distributions, breaking free from the particular case where posterior and likelihood are exclusively Gaussians.
Note also how FM models time \( t \in [0,1] \) to be varying continuously while moving away \emph{from} an easy-to-sample distribution \( p_0 \) \emph{towards} the unknown data-distribution, \( p_1 \).
This results in a continuous (and deterministic) trajectory at inference, which is in practice more efficient compared to following stochastic paths like in DMs.
Formally, FM can be fully characterized by an ordinary differential equation (ODE) relating instantaneous variations of flows with the underlying vector field, and hence providing complete trajectories over the distributions' support when integrating over time,
\begin{align}
    \frac{d}{dt} \psi(z, t) &= v(t, \psi(t, z)), \\
    \psi(0, z) &= z .
\end{align}
In practice, flow models learn to approximate these dynamics by estimating a vector field \( v \) that matches the true, unknown \( u \), so that the induced flows \( \psi \) can approximate the ideal trajectories \( \psi^* \).

FM proved very effective in a variety of applications, ranging from image~\citep{esserScalingRectifiedFlow2024} and video generation~\citep{polyakMovieGenCast2025} to robotics control~\citep{black$p_0$VisionLanguageActionFlow2024}.
Most notably, in their introductory work on FM for GM,~\citet{lipmanFlowMatchingGenerative2023} show how DMs can be seen as a specific instance of FM where the \emph{conditional} target vector field \( v \) learned by the noise regressor \( \eps_\theta \) corresponds to:
\begin{equation}\label{eq:fm-diffusion-vector-field}
    u(t, z\vert z_0) = \frac{\frac{d}{dt}\alpha(1-t)}{1 - (\alpha(1-t))^2}(\alpha(1-t)z - z_0), \quad \alpha(t) = e^{-\frac12 \int_0^t \beta(s) ds}, \quad \forall z_0 \in \mathcal D.
\end{equation}
Conditional vector fields are defined not only over their argument \( z \) and time \( t\), but do also vary with respect to an auxiliary variable \( z_0 \), thereby extending the standard notion of a vector field to incorporate additional conditioning.
Note that the traditional discrete-time noise-scheduler \( \{\beta_t\}_{t=0}^T \) is now generalized to a continuous map \( \beta : [0,1] \mapsto \mathbb R^+ \).
Crucially,~\citet{lipmanFlowMatchingGenerative2023} prove that by exclusively optimizing the vector field for individual data points \( z_0 \in \mathcal D \), one also retrieves the optimal flow to morph the entire support of the initial distribution \( p_0 \) into \( p_1 \ \text{s.t.} \mathcal D \sim p_1 \).

\begin{figure}
    \centering
    \includegraphics[width=0.9\textwidth]{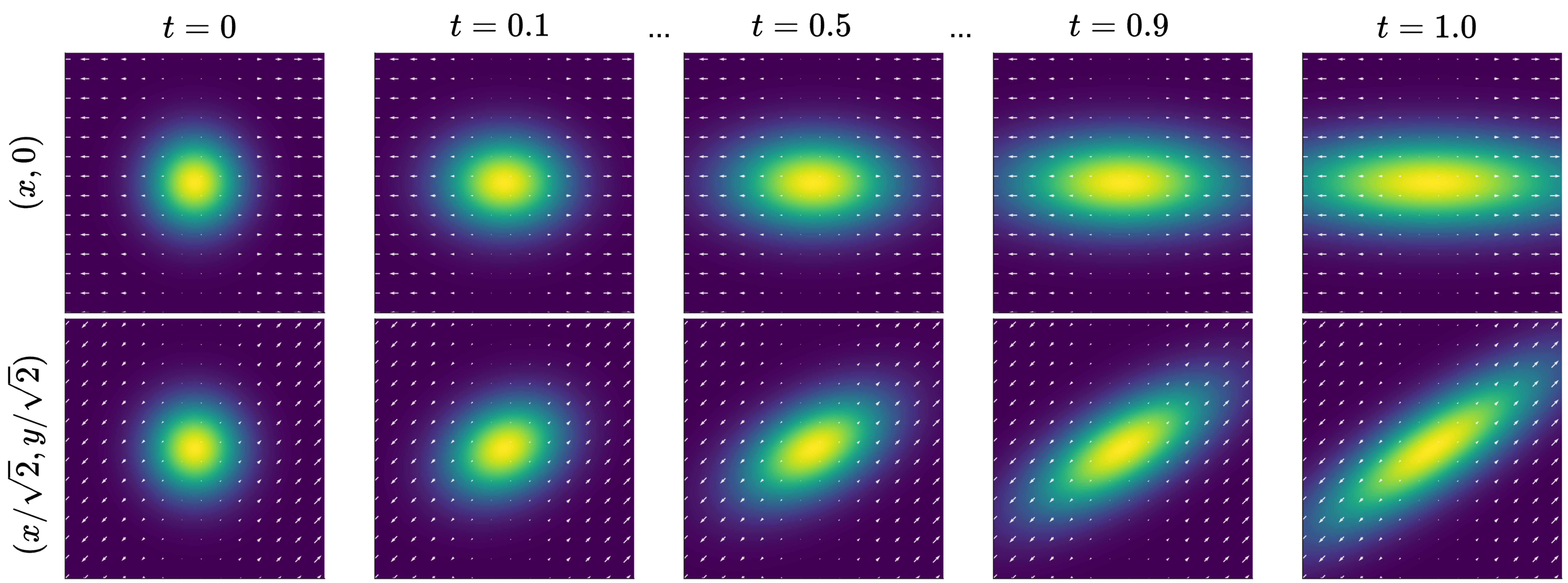}
    \caption{Probability distributions can be modified differently by applying different vector fields, inducing different flows of mass across the same support (top versus bottom, using two different time-invariant 2D-fields \( u_1(x,y) = (x,0) \) and \( u_2(x,y) = (x/\sqrt{2}, y/\sqrt{2}) \)). Notice time flows \emph{continuously} in \( [0,1] \). FM models learn to approximate a target vector field, thereby producing arbitrary (goal) transformations of an easy-to-sample initial distribution.}
    \label{fig:ch4-normalizing-flows}
\end{figure}

\begin{figure}
    \centering
    \includegraphics[width=\textwidth]{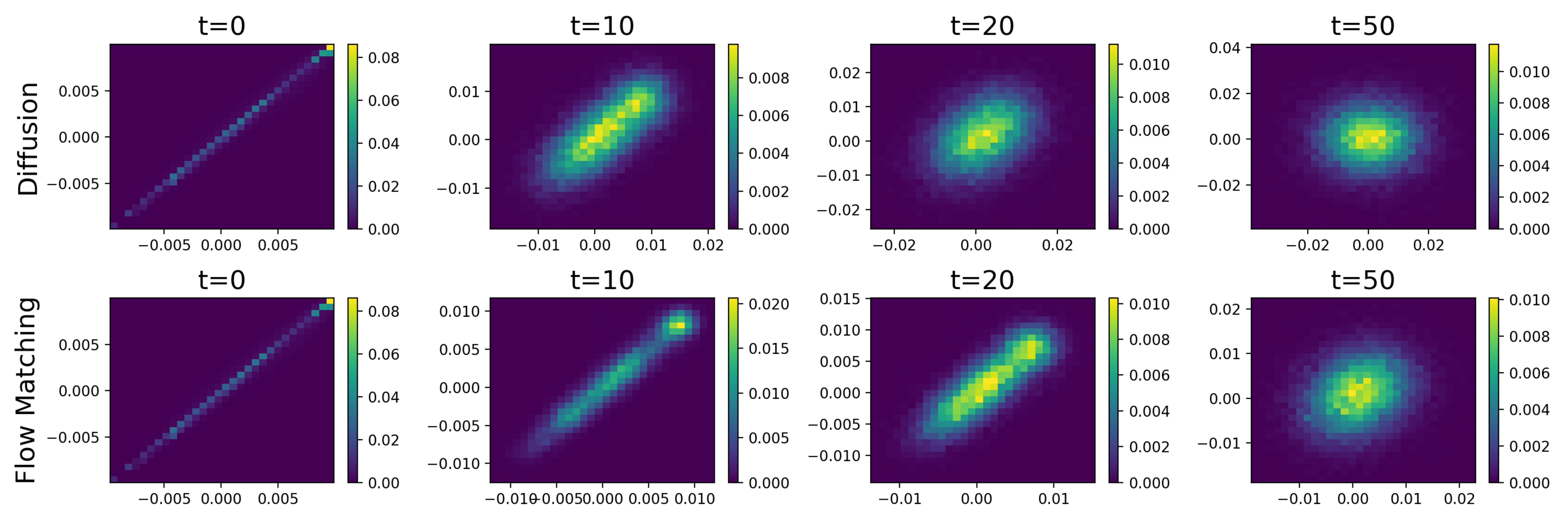}
    \caption{Compared to diffusion, flow matching distorts distribution along a less randomic pattern, resulting in a clearer interpolation between source and target distribution. The visualization shows an example comparison between these two methods on joint distribution of robot observations and actions over \( T=50 \) steps.}
    \label{fig:ch4-diffusion-paths-versus-fm}
\end{figure}

While the noising schedule of DMs results in a stochastic resembling a random (Brownian) walk, FM allows for more general---potentially, deterministic---likelihood and posterior parametrization.
In the FM literature the likelihood and posterior probabilty densities defined along a HMLV model are typically referred to as a \emph{probability path}, where the distributions for successive adjacent transitions in the HMLV model are related by the (normalized) flow between them (Figure~\ref{fig:ch4-normalizing-flows}).
The inherent flexibility of FM is one of their key advantages over DMs, as it opens up the possibility of \emph{learning} more efficient paths.
For instance, one can design probability paths inspired by Optimal Transport (OT), a mathematical framework concerned with characterizing the most efficient morphings between probability distributions.
Probability paths obtained through OT paths tend to be \emph{straighter} than diffusion paths (Figure~\ref{fig:ch4-diffusion-paths-versus-fm}), which can lead to faster and more stable training, as well as empirically result in higher-quality generations with fewer denoising steps at inference time.
In particular, by avoiding unnecessary backtracking associated with the inherent stochastic nature of both the noising and denoising process in DMs, test-time compute is typically significantly reduced in FM, while retaining comparable results~\citep{lipmanFlowMatchingGenerative2023}.

In practice, FM can be applied to generative modeling by learning a vector field regressor \( v_\theta(z, t) \) to approximate a given target vector field \( u(t, z) \).
In the particular case of DMs, \( u(t, z) \) is defined as in eq.~\ref{eq:fm-diffusion-vector-field}, while in priciple the target vector field can be learned to induce an arbitrary mass displacement, or fixed according to OT.
Given a sample from the data distribution \( z_1 \sim p_1 \) and a sample from an easy-to-sample prior \( z_0 \sim p_0 \), Conditional FM (CFM) defines a simple path between them using \emph{linear interpolation} between samples \( z_t = (1-t)z_0 + t z_1 \), which in turn results in the target vector field \( u(t, z_t) = z_1 - z_0 \).
FM models can then be trained with a simple regression objective defined as:
\begin{align}\label{eq:flow-matching-objective}
    \mathcal L(\theta) = \mathbb{E}_{t, z_0, z_1} \big[
        \Vert v_\theta((1-t)z_0 + t z_1, t) - (z_1 - z_0) \Vert^2 \big], \quad t \sim \mathcal{U}([0,1]),
\end{align}
where \( z_0 \sim p_0(\bullet) \) and \( z_1 \sim p_1(\bullet) \). 
Note how in eq.~\ref{eq:flow-matching-objective}---differently from eq.~\ref{eq:diffusion-simplified-loss}---time is assumed to be varying continuously \( t \sim \mathcal U([0,1]) \) rather than discretely \( t \sim \mathcal U(\{0, \Delta t, 2 \Delta t, \dots, 1 \})\), a key property of flow-based models.
Therefore, the objective in eq.~\ref{eq:flow-matching-objective} directly regresses the learned vector field onto the simple, straight path connecting a point from the prior and a point from the data, providing a simulation-free training procedure that is both stable and efficient.
At inference time, samples are generated by starting with \( z_0 \sim p_0 \) and iteratively refined according to \( \frac{dz}{dt} = v_\theta(z_t, t) \) for \(t \in [0,1] \)---an operation that can be numerically carried out with standard ODE solvers, and that in practice is often carried out numerically via forward-Euler integrating over tens of denoising steps.

\subsection{Action Chunking with Transformers}
While GMs prove useful in learning complex, high-dimensional multi-modal distributions, they do not natively address the compouding errors problem characteristic of modeling online, sequential predictions.
In Action Chunking with Transformers (ACT),~\citet{zhaoLearningFineGrainedBimanual2023} present an application of VAEs to the problem of learning purely from offline trajectories, and introduce a simple, yet effective method to mitigate error compounding, learning high-fidelity autonomous behaviors via BC.
Drawing inspiration from how humans plan to enact \emph{sequences} of actions \( a_{t:t+k} \) instead of single actions \( a_t \),~\citet{zhaoLearningFineGrainedBimanual2023} propose learning a GM on a dataset of input demonstrations by modeling \emph{chunks} of multiple actions directly.
Besides contributions to learning high-performance autonomous behaviors,~\citet{zhaoLearningFineGrainedBimanual2023} also introduce hardware contributions in the form of a low-cost bimanual robot setup (ALOHA) capable of performing fine-grained manipulation tasks, such as opening a lid, slotting a battery in its allotment or even prepare tape for application.
Notably, ALOHA bimanual setup costs just as much as a mono-arm Franka arm and can be assembled from easy-to-source parts, underscoring its higher accessibility.

\citet{zhaoLearningFineGrainedBimanual2023} do also present significant algorithmic contributions related to synthetizing performant autonomous behaviors for the ALOHA setup, adopting transformers as the architectural backbone to learn a \emph{Conditional} VAE~\citep{sohnLearningStructuredOutput2015} from demonstrations. 
Conditional VAEs are a variation of the standard VAE  introducing an arbitrary conditioning on sampling from the latent prior, modeling \emph{one-to-many} relationships between latent and data samples.
Further, in stark contrast with previous work~\citep{florenceImplicitBehavioralCloning2022,jannerPlanningDiffusionFlexible2022},~\citet{zhaoLearningFineGrainedBimanual2023} do not learn a full joint \( p_\theta(o,a) \) on observation and actions, and rather focus on the conditional \( p_\theta(a \vert o) \).
While the \emph{policy} distribution \( p_\theta(a \vert o) \) can in principle be entirely described from the joint \( p_\theta(o,a) \), conditional distributions are often intractable when using function approximators, as \( p_\theta(a \vert o) = \tfrac{p_\theta(o,a)}{\int_\actionspace p_\theta(o,a)} \), and the integral in the denominator is typically intractable.
Thus, instead of modeling the full joint using a vanilla VAE,~\citet{zhaoLearningFineGrainedBimanual2023} propose learning a \emph{conditional} VAE~\citep{sohnLearningStructuredOutput2015} modeling the policy distribution directly, hence approximating \( p (a \vert o) \).

In practice, when learning from demonstrations adopting CVAEs results in a slight modification to the VAE objective in eq.~\ref{eq:ELBO}, which is adapted to:
\begin{align}\label{eq:c-ELBO}
    \text{ELBO}_{\mathcal D}(\theta, \phi, \omega) = \sum_{i=0}^{N} \left(
            \mathbb{E}_{z \sim q_\phi(\cdot \vert o_i, a_i)} \big[ \log p_\theta(a_i \vert z, o_i) \big]
        - \DKL \big[ q_\phi(z \vert o_i, a_i) \Vert p_\omega(z \vert o_i) \big]
        \right)
\end{align}
Notice how in eq.~\ref{eq:c-ELBO} we are now also learning a new set of parameters \( \omega \) for the prior distribution in the latent space.
Effectively, this enables conditioning latent-space sampling (and thus reconstruction) during training (and potentially inference too), providing useful when learning inherently conditional distributions like policies.
Further, ACT is trained as a \( \beta\)-CVAE~\citep{higgins2017beta}, weighing the KL regularization term in eq.~\ref{eq:c-ELBO} with an hyperparameter \( \beta \in \mathbb R^+ \) regulating the information condensed in the latent space, where \emph{higher} \( \beta \) results in a \emph{less} expressive latent space.

In their work,~\citet{zhaoLearningFineGrainedBimanual2023} ablated using a GM to learn from human demonstrations compared to a simpler, supervised objective, \( \mathcal L_1(a,a^\prime) = \Vert a - a^\prime \Vert_1 \).
Interestingly, they found the performance of these two approaches to be comparable when learning from \emph{scripted} demonstrations. 
That is, when learning from data collected rolling out a predetermined set of commands \( [q^c_0, q^c_1, \dots] \), GM did \emph{not} prove competitive compared to standard supervised learning.
However, when learning from human demonstrations---i.e., from data collected executing commands coming from a human controller \( [q^h_0, q^h_1, \dots] \)---~\citet{zhaoLearningFineGrainedBimanual2023} found performance (defined as the success rate on a downstream task) to be severily (-33.3\%) hindered from adopting a standard supervised learning objective compared to a richer, potentially more complex to learn variational objective.
The result of such ablation reflects from the multimodal nature of human demonstrations data, and is consistent with the findings presented by~\citet{florenceImplicitBehavioralCloning2022}.
The authors also ablate the action chunking paradigm, reporting significant performance gains deriving from using action chunking (1\% vs. 44\% success rate).
To reduce acting open-loop,~\citet{zhaoLearningFineGrainedBimanual2023} also design an inference process consisting in performing inference at every timestep \( t \) and then aggregate multiple chunks using an exponential moving average (EMA) on the overlapping chunks.

\begin{figure}
    \centering
    \includegraphics[width=0.9\textwidth]{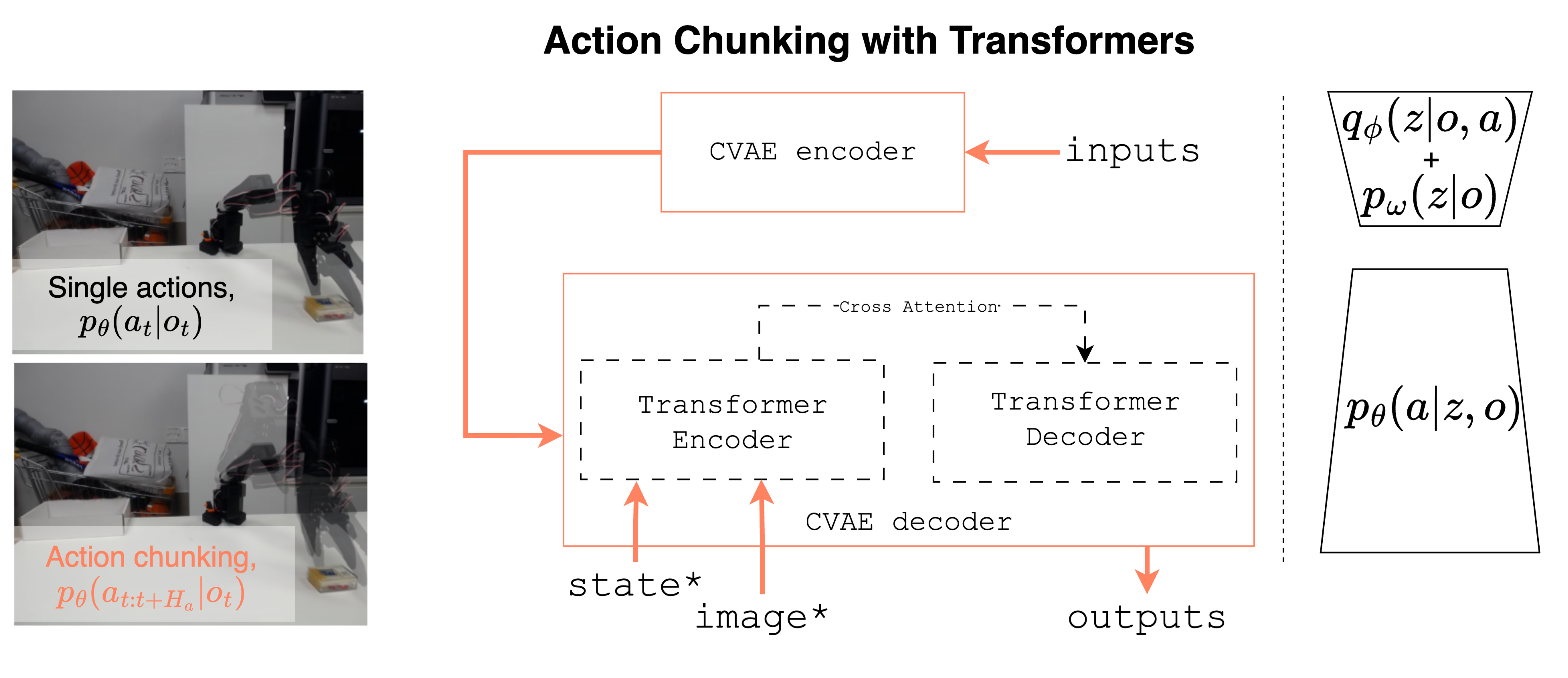}
    \caption{Action Chunking with Transformer (ACT), as in~\citet{zhaoLearningFineGrainedBimanual2023}. ACT introduces an action chunking paradigm to cope with high-dimensional multi-modal demonstration data, and a transformer-based CVAE architecture.}
    \label{fig:ch4-act}
\end{figure}

In ACT (Figure~\ref{fig:ch4-act}), inference for a given observation \( o \in \mathcal O \) could be performed by (1) defining a prior \( p_\omega(z \vert o) \) for the latent variable \( z \) and (2) decoding an action chunk from a sampled latent \( z \sim p_\omega(\bullet \vert o) \), similarily to how sampling from standard VAEs takes place, with the exception that vanilla VAEs typically pose \( p(z\vert o) \equiv p(z) \sim \mathcal N(\mathbf{0}, \mathbf{I}) \) and thus skip (1).

\begin{figure}
    \centering
    \includegraphics[width=0.75\textwidth]{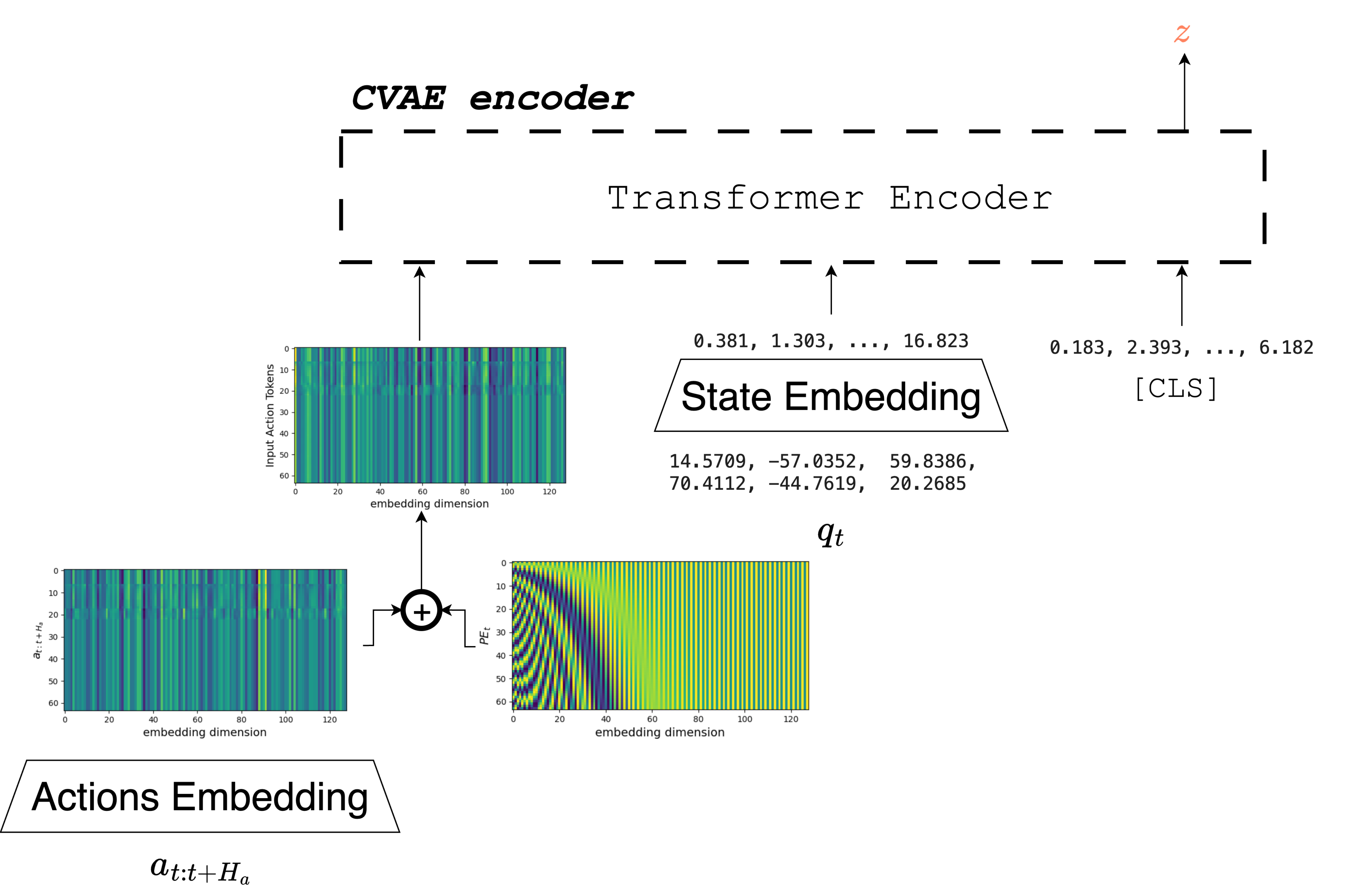}
    \caption{The CVAE encoder used in ACT. Input action chunks are first embedded and aggregated with positional embeddings, before being processed alongside embedded proprioperceptive information, and a learned \texttt{[CLS]} token used to aggregate input level information, and predict the style variable \( z \). The encoder is exclusively used to \emph{train} the decoder, and it is entirely disregarded at inference time.}
    \label{fig:ch4-act-encoder}
\end{figure}

However, the authors claim that using a deterministic procedure to sample \( z \) benefits policy evaluation, and thus avoid using the conditional prior at all at inference time, effectively using the CVAE framework exclusively to train a more expressive decoder.
At test time,~\citet{zhaoLearningFineGrainedBimanual2023} propose simply using \( z = \mathbf{0} \), as the conditional prior on \( z \) used in training is set to be a standard Gaussian.
Further, conditioning on the observation \( o \) is achieved through explicitly feeding proprioperceptive and visual observations to the decoder, \( p_\theta(a \vert z, o) \) at test time.
If at inference \( z \) is sampled from a standard Gaussian, during training \( z \) is sampled from an approximate posterior distribution \(q_\phi(z \vert o, a)\), which, however, disregards image observations and exclusively uses proprioperceptive states to form \( o \) for efficiency reasons.

\begin{figure}
    \centering
    \includegraphics[width=0.75\textwidth]{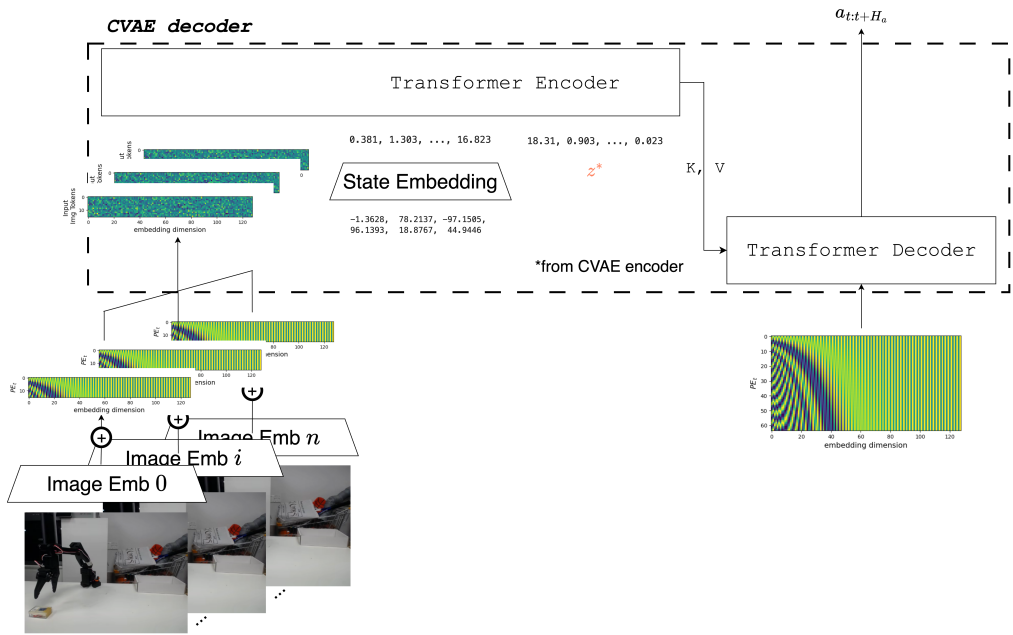}
    \caption{The CVAE decoder used in ACT, comprising of a full encoder-decoder Transformer architecture. Camera observations from all \( n \) camera views are first embedded using pre-trained visual encoders, and then aggregated with the corresponding positional embeddings. Then, the proprioperceptive information and style variable \( z \) retrieved from the CVAE encoder, are fed to the encoder-decoder Transformer for inference. The encoder shares the matrices \( K,V \) with the decoder, and is trained to decode fixed position embeddings into action chunks.}
    \label{fig:ch4-act-decoder}
\end{figure}

\subsubsection{Code Example: Training and Using ACT in Practice}

\begin{pbox}[label={ex:act_training}]{Training ACT \\ \url{https://github.com/fracapuano/robot-learning-tutorial/snippets/ch4/01_training_act.py}}
\begin{lstlisting}[language=python]
from pathlib import Path

import torch

from lerobot.configs.types import FeatureType
from lerobot.datasets.lerobot_dataset import LeRobotDataset, LeRobotDatasetMetadata
from lerobot.datasets.utils import dataset_to_policy_features
from lerobot.policies.act.configuration_act import ACTConfig
from lerobot.policies.act.modeling_act import ACTPolicy
from lerobot.policies.factory import make_pre_post_processors


def make_delta_timestamps(delta_indices: list[int] | None, fps: int) -> list[float]:
    if delta_indices is None:
        return [0]

    return [i / fps for i in delta_indices]


output_directory = Path("outputs/robot_learning_tutorial/act")
output_directory.mkdir(parents=True, exist_ok=True)

# Select your device
device = torch.device("mps")  # or "cuda" or "cpu"

dataset_id = "lerobot/svla_so101_pickplace"

# This specifies the inputs the model will be expecting and the outputs it will produce
dataset_metadata = LeRobotDatasetMetadata(dataset_id)
features = dataset_to_policy_features(dataset_metadata.features)

output_features = {key: ft for key, ft in features.items() if ft.type is FeatureType.ACTION}
input_features = {key: ft for key, ft in features.items() if key not in output_features}

cfg = ACTConfig(input_features=input_features, output_features=output_features)
policy = ACTPolicy(cfg)
preprocessor, postprocessor = make_pre_post_processors(
    cfg, dataset_stats=dataset_metadata.stats
)

policy.train()
policy.to(device)

# To perform action chunking, ACT expects a given number of actions as targets
delta_timestamps = {
    "action": make_delta_timestamps(cfg.action_delta_indices, dataset_metadata.fps),
}

# add image features if they are present
delta_timestamps |= {
    k: make_delta_timestamps(cfg.observation_delta_indices, dataset_metadata.fps)
    for k in cfg.image_features
}

# Instantiate the dataset
dataset = LeRobotDataset(dataset_id, delta_timestamps=delta_timestamps)

# Create the optimizer and dataloader for offline training
optimizer = cfg.get_optimizer_preset().build(policy.parameters())
batch_size = 32
dataloader = torch.utils.data.DataLoader(
    dataset,
    batch_size=batch_size,
    shuffle=True,
    pin_memory=device.type != "cpu",
    drop_last=True,
)

# Number of training steps and logging frequency
training_steps = 1
log_freq = 1

# Run training loop
step = 0
done = False
while not done:
    for batch in dataloader:
        batch = preprocessor(batch)
        loss, _ = policy.forward(batch)
        loss.backward()
        optimizer.step()
        optimizer.zero_grad()

        if step % log_freq == 0:
            print(f"step: {step} loss: {loss.item():.3f}")
        step += 1
        if step >= training_steps:
            done = True
            break

# Save the policy checkpoint, alongside the pre/post processors
policy.save_pretrained(output_directory)
preprocessor.save_pretrained(output_directory)
postprocessor.save_pretrained(output_directory)

# Save all assets to the Hub
policy.push_to_hub("fracapuano/robot_learning_tutorial_act_example_model")
preprocessor.push_to_hub("fracapuano/robot_learning_tutorial_act_example_pipeline")
postprocessor.push_to_hub("fracapuano/robot_learning_tutorial_act_example_pipeline")
\end{lstlisting}
\end{pbox}

\begin{pbox}[label={ex:act_using}]{Using ACT \\ \url{https://github.com/fracapuano/robot-learning-tutorial/snippets/ch4/02_using_act.py}}
\begin{lstlisting}[language=python]
import torch

from lerobot.cameras.opencv.configuration_opencv import OpenCVCameraConfig
from lerobot.datasets.lerobot_dataset import LeRobotDatasetMetadata
from lerobot.policies.act.modeling_act import ACTPolicy
from lerobot.policies.factory import make_pre_post_processors
from lerobot.policies.utils import build_inference_frame, make_robot_action
from lerobot.robots.so100_follower.config_so100_follower import SO100FollowerConfig
from lerobot.robots.so100_follower.so100_follower import SO100Follower

device = torch.device("mps")  # or "cuda" or "cpu"
model_id = "fracapuano/robot_learning_tutorial_act_example_model"
model = ACTPolicy.from_pretrained(model_id)

dataset_id = "lerobot/svla_so101_pickplace"
# This only downloads the metadata for the dataset, ~10s of MB even for large-scale datasets
dataset_metadata = LeRobotDatasetMetadata(dataset_id)
preprocess, postprocess = make_pre_post_processors(
    model.config, dataset_stats=dataset_metadata.stats
)

# # find ports using lerobot-find-port
follower_port = ...  # something like "/dev/tty.usbmodem58760431631"

# # the robot ids are used the load the right calibration files
follower_id = ...  # something like "follower_so100"

MAX_EPISODES = 5
MAX_STEPS_PER_EPISODE = 20

# Robot and environment configuration
# Camera keys must match the name and resolutions of the ones used for training!
# You can check the camera keys expected by a model in the info.json card on the Hub
camera_config = {
    "side": OpenCVCameraConfig(index_or_path=1, width=640, height=480, fps=30),
    "up": OpenCVCameraConfig(index_or_path=1, width=640, height=480, fps=30),
}

robot_cfg = SO100FollowerConfig(port=follower_port, id=follower_id, cameras=camera_config)
robot = SO100Follower(robot_cfg)
robot.connect()

for _ in range(MAX_EPISODES):
    for _ in range(MAX_STEPS_PER_EPISODE):
        obs = robot.get_observation()
        obs_frame = build_inference_frame(obs, dataset_metadata.features, device)

        obs = preprocess(obs_frame)

        action = model.select_action(obs)
        action = postprocess(action)

        action = make_robot_action(action, dataset_metadata.features)

        robot.send_action(action)

    print("Episode finished! Starting new episode...")
\end{lstlisting}
\end{pbox}

\subsection{Diffusion Policy}
DMs have proven very effective in approximating complex highly dimensional distributions, such as distributions over images~\citep{hoDenoisingDiffusionProbabilistic2020} or videos~\citep{polyakMovieGenCast2025}, thanks to their inherent capability to deal with multimodal data, and their training stability.
In Diffusion Policy (DP),~\citet{chiDiffusionPolicyVisuomotor2024} present an application of DMs the field of robot learning, leveraging diffusion to model expert demonstrations in a variety of simulated and real-world tasks.
Similarily to ACT~\citep{zhaoLearningFineGrainedBimanual2023},~\citet{chiDiffusionPolicyVisuomotor2024} (1) adopt a modified \emph{observation-conditioned target distribution} instead of the full joint \( p(o,a) \), and (2) predict multiple actions into the future instead of a single action.
Besides the intractability of the observations' marginal \( p_\theta(o) \) given \(p_\theta(o,a) \), DP's choice to model the data distribution through \( p_\theta(a \vert o) \) also stems from the computational burden of diffusion at test time: generating actions together with observations would require a large number of denoising steps—an unnecessarily slow and ultimately unhelpful process, given that robotics focuses on producing controls rather than reconstructing observations.

In practice, conditioning on observation data is achieved conditioning the noise regressor \( \epsilon_\theta \) introduced in eq.~\ref{eq:diffusion-simplified-loss} on a stack of \( H_o \) observations, resulting in the \emph{conditional}, simplified diffusion objective:
\begin{align}
    \mathcal L(\theta) &= \mathbb{E}_{t, a_{t:t+H_a}, \epsilon} \big[
        \Vert \epsilon - \epsilon_\theta(\sqrt{\bar \alpha_t} a_{t:t+H_a} + \epsilon \sqrt{1 - \bar \alpha_t}, t, o_{t-H_o:t}) \Vert^2 \big], \label{eq:diffusion-policy-objective} \\
        & t \sim \mathcal{U}(\{1,\dots,T\}), \quad
        a_{t:t+H_a}, o_{t-H_o:t} \sim \mathcal{D}, \quad
        \epsilon \sim \mathcal{N}(\mathbf{0},\mathbf{I}). \notag 
\end{align}
Note how in eq.~\ref{eq:diffusion-policy-objective} the noise regressor is conditioned on both the latent variable rank \( t \) \emph{and} on a stack of previous observations \(o_{t-H_o:t} \).
\citet{chiDiffusionPolicyVisuomotor2024} claim the combination of (1) conditioning on a horizon of previous observations and (2) predicting multiple actions into the future allows DP to \emph{commit to specific modes} in the data at inference time, which proves essential for good performance and avoiding undecisiveness.

\begin{figure}
    \centering
    \includegraphics[width=0.9\textwidth]{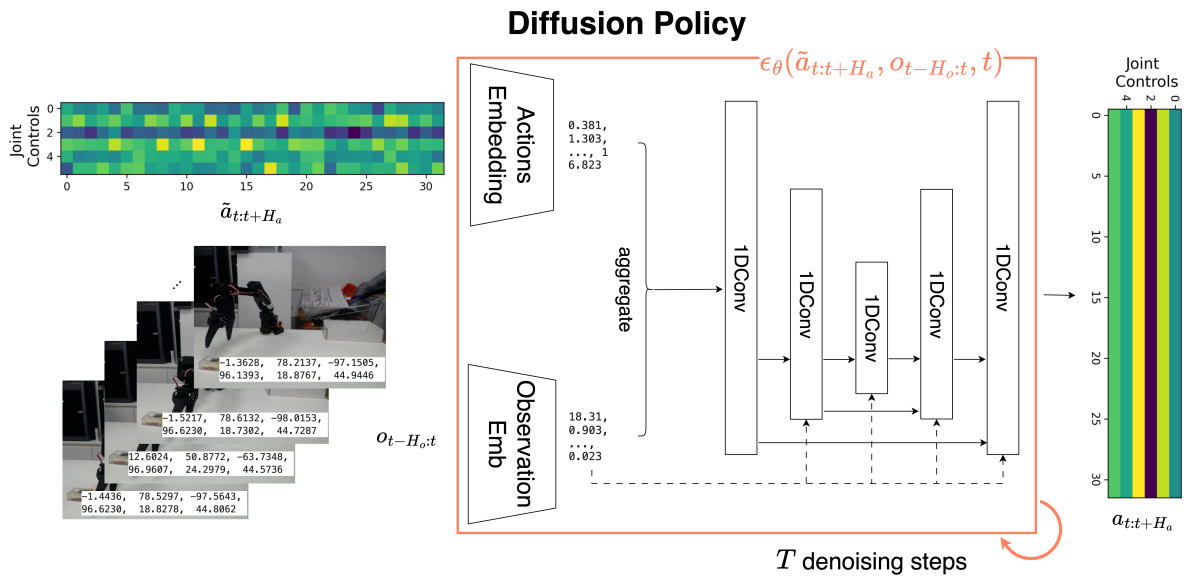}
    \caption{The Diffusion Policy archicture, as in~\citet{chiDiffusionPolicyVisuomotor2024}. A stack of \( H_o \) previous observations is used as external conditioning to denoise a group of \( H_a \) actions. Conditioning is performed at every layer of a U-Net block. Diffusion Policy allows to obtain fully-formed action chunks with as little as \(T=10\) denoising steps.}
    \label{fig:diffusion-policy-architecture}
\end{figure}

Figure~\ref{fig:diffusion-policy-architecture} shows the convolution-based version of the architecture proposed by~\citet{chiDiffusionPolicyVisuomotor2024}, illustrating inference on a single sample drawn from \( \mathcal D \), for simplicity.
The starting, arbitrarily noisy chunk of \( H_a \) actions \(\tilde a_{t:t+H_a} \) is first mapped to a (learned) high-dimensional space.
Similarily, both image observations and poses are also embedded before being aggregated to the action embeddings.
Then, a U-Net~\citep{ronnebergerUNetConvolutionalNetworks2015} is trained to regress the noise added into \( \tilde a_{t:t+H_a} \), conditioned on observation information at every layer, thus seeking to optimize eq.~\ref{eq:diffusion-policy-objective}.
At inference time, the noise predictor is used to predict the quantity of noise at every \( t \in [T, \dots, 0 ] \) and iteratively subtract it from \(\tilde a_{t:t+H_a} \), reversing the diffusion process simulated in training conditioned on \(o_{t-H_o:t} \) to predict \(a_{t:t+H_a} \).

DP can be trained with as little as 50-150 demos (ca. 15-60 minutes of teleoperation data), and exhibit strong performance on a variety of simulated and real-world tasks, including dexterous and deformable manipulation tasks such as sauce pouring and yoga-mat unrolling.
Notably, the authors ablated the relevance of using RGB camera streams as input to their policy, and observed how high frame-rate visual observations can be used to attain performance (measured as success rate) comparable to that of state-based policies, which are typically trained in simulation with priviledged information not directly available in real-world deployments.
As high-frame rate RGB inputs naturally accomodate for dynamic, fast changing environments,~\citet{chiDiffusionPolicyVisuomotor2024}'s conclusion offers significant evidence for learning streamlined control policies directly from pixels.
In their work,~\citet{chiDiffusionPolicyVisuomotor2024} also ablate the performance of DP against the size of the dataset collected, showing that DP reliably outperforms the considered baseline for all benchmark sizes considered.
Further, in order  accelerate inference,~\citet{chiDiffusionPolicyVisuomotor2024} employ Denoising Diffusion Implicit Models~\citep{songDenoisingDiffusionImplicit2022}, a variant of Denoising Diffusion Probabilistic Models~\citep{hoDenoisingDiffusionProbabilistic2020} (DDPM) adopting a strictly deterministic denoising paradigm (differently from DDPM's natively stochastic one) inducing the same final distribution's as DDPM's, and yet resulting in 10x less denoising steps at inference time~\citep{chiDiffusionPolicyVisuomotor2024}.
Across a range of simulated and real-world tasks,~\citet{chiDiffusionPolicyVisuomotor2024} find DPs particularly performant when modeling \( \epsilon_\theta \) with a transformer-based network, although the authors note the increased sensitivity of transformer networks to hyperparameters.
Thus,~\citet{chiDiffusionPolicyVisuomotor2024} explicitly recommend starting out with a simpler, convolution-based architecture for diffusion (Figure~\ref{fig:diffusion-policy-architecture}), which is however reported to be biased towards learning low-frequency components~\citep{tancikFourierFeaturesLet2020}, and thus may prove more challenging to train with non-smooth action sequences.

\subsubsection{Code Example: Training and Using Diffusion Policies in Practice}

\begin{pbox}[label={ex:diffusion_training}]{Training Diffusion Policy \\ \url{https://github.com/fracapuano/robot-learning-tutorial/blob/main/snippets/ch4/03_training_diffusion.py}}
\begin{lstlisting}[language=python]
from pathlib import Path

import torch

from lerobot.configs.types import FeatureType
from lerobot.datasets.lerobot_dataset import LeRobotDataset, LeRobotDatasetMetadata
from lerobot.datasets.utils import dataset_to_policy_features
from lerobot.policies.diffusion.configuration_diffusion import DiffusionConfig
from lerobot.policies.diffusion.modeling_diffusion import DiffusionPolicy
from lerobot.policies.factory import make_pre_post_processors


def make_delta_timestamps(delta_indices: list[int] | None, fps: int) -> list[float]:
    if delta_indices is None:
        return [0]

    return [i / fps for i in delta_indices]


output_directory = Path("outputs/robot_learning_tutorial/diffusion")
output_directory.mkdir(parents=True, exist_ok=True)

# Select your device
device = torch.device("mps")  # or "cuda" or "cpu"

dataset_id = "lerobot/svla_so101_pickplace"

# This specifies the inputs the model will be expecting and the outputs it will produce
dataset_metadata = LeRobotDatasetMetadata(dataset_id)
features = dataset_to_policy_features(dataset_metadata.features)

output_features = {key: ft for key, ft in features.items() if ft.type is FeatureType.ACTION}
input_features = {key: ft for key, ft in features.items() if key not in output_features}

cfg = DiffusionConfig(input_features=input_features, output_features=output_features)
policy = DiffusionPolicy(cfg)
preprocessor, postprocessor = make_pre_post_processors(
    cfg, dataset_stats=dataset_metadata.stats
)

policy.train()
policy.to(device)

# To perform action chunking, ACT expects a given number of actions as targets
delta_timestamps = {
    "observation.state": make_delta_timestamps(
        cfg.observation_delta_indices, dataset_metadata.fps
    ),
    "action": make_delta_timestamps(cfg.action_delta_indices, dataset_metadata.fps),
}

# add image features if they are present
delta_timestamps |= {
    k: make_delta_timestamps(cfg.observation_delta_indices, dataset_metadata.fps) 
    for k in cfg.image_features
}

# Instantiate the dataset
dataset = LeRobotDataset(dataset_id, delta_timestamps=delta_timestamps)

# Create the optimizer and dataloader for offline training
optimizer = cfg.get_optimizer_preset().build(policy.parameters())
batch_size = 32
dataloader = torch.utils.data.DataLoader(
    dataset,
    batch_size=batch_size,
    shuffle=True,
    pin_memory=device.type != "cpu",
    drop_last=True,
)

# Number of training steps and logging frequency
training_steps = 1
log_freq = 1

# Run training loop
step = 0
done = False
while not done:
    for batch in dataloader:
        batch = preprocessor(batch)
        loss, _ = policy.forward(batch)
        loss.backward()
        optimizer.step()
        optimizer.zero_grad()

        if step % log_freq == 0:
            print(f"step: {step} loss: {loss.item():.3f}")
        step += 1
        if step >= training_steps:
            done = True
            break

# Save the policy checkpoint, alongside the pre/post processors
policy.save_pretrained(output_directory)
preprocessor.save_pretrained(output_directory)
postprocessor.save_pretrained(output_directory)

# Save all assets to the Hub
policy.push_to_hub("fracapuano/robot_learning_tutorial_diffusion_example_model")
preprocessor.push_to_hub("fracapuano/robot_learning_tutorial_diffusion_example_model")
postprocessor.push_to_hub("fracapuano/robot_learning_tutorial_diffusion_example_model")
\end{lstlisting}
\end{pbox}

\begin{pbox}[label={ex:diffusion_using}]{Using Diffusion Policy \\ \url{https://github.com/fracapuano/robot-learning-tutorial/blob/main/snippets/ch4/04_using_diffusion.py}}
\begin{lstlisting}[language=python]
import torch

from lerobot.cameras.opencv.configuration_opencv import OpenCVCameraConfig
from lerobot.datasets.lerobot_dataset import LeRobotDatasetMetadata
from lerobot.policies.diffusion.modeling_diffusion import DiffusionPolicy
from lerobot.policies.factory import make_pre_post_processors
from lerobot.policies.utils import build_inference_frame, make_robot_action
from lerobot.robots.so100_follower.config_so100_follower import SO100FollowerConfig
from lerobot.robots.so100_follower.so100_follower import SO100Follower

device = torch.device("mps")  # or "cuda" or "cpu"
model_id = "fracapuano/robot_learning_tutorial_diffusion_example_model"

model = DiffusionPolicy.from_pretrained(model_id)

dataset_id = "lerobot/svla_so101_pickplace"
# This only downloads the metadata for the dataset, ~10s of MB even for large-scale datasets
dataset_metadata = LeRobotDatasetMetadata(dataset_id)
preprocess, postprocess = make_pre_post_processors(
    model.config, model_id, dataset_stats=dataset_metadata.stats
)

MAX_EPISODES = 5
MAX_STEPS_PER_EPISODE = 20


# # find ports using lerobot-find-port
follower_port = ...  # something like "/dev/tty.usbmodem58760431631"

# # the robot ids are used the load the right calibration files
follower_id = ...  # something like "follower_so100"

# Robot and environment configuration
# Camera keys must match the name and resolutions of the ones used for training!
# You can check the camera keys expected by a model in the info.json card on the Hub
camera_config = {
    "side": OpenCVCameraConfig(index_or_path=1, width=640, height=480, fps=30),
    "up": OpenCVCameraConfig(index_or_path=1, width=640, height=480, fps=30),
}

robot_cfg = SO100FollowerConfig(port=follower_port, id=follower_id, cameras=camera_config)
robot = SO100Follower(robot_cfg)
robot.connect()


for _ in range(MAX_EPISODES):
    for _ in range(MAX_STEPS_PER_EPISODE):
        obs = robot.get_observation()
        obs_frame = build_inference_frame(obs, dataset_metadata.features, device)

        obs = preprocess(obs_frame)

        action = model.select_action(obs)
        action = postprocess(action)
        action = make_robot_action(action, dataset_metadata.features)
        robot.send_action(action)

    print("Episode finished! Starting new episode...")
\end{lstlisting}
\end{pbox}

\subsection{Optimized Inference}
\label{sec:ch4-async-inference}
Modern visuomotor policies output \emph{action chunks}--sequences \(\pi(o_t) = \bigl(a_t,a_{t+1},\dots,a_{t+H_a}\bigr) = \actionchunk_t \) with \(\actionchunk_t \) a sequence of \(H_a \gg 1 \) low-level commands scheduled for execution in an action queue, all originating from a single environment observation, \(o_t\).
Predicting series of actions instead of single commands proved essential in learning complex, multi-modal behavior~\citep{zhaoLearningFineGrainedBimanual2023,chiDiffusionPolicyVisuomotor2024}, and it also holds the premise to be useful to optimize how inference is carried out in practice.

A robot may indeed execute an entire action chunk \(\actionchunk_t \) \emph{before} a new observation \( o_{t+H_a} \) is passed to the policy \( \pi \) to predict the next chunk, which would result in open-loop control between observations captured every \( H_a \) timesteps.
\citet{zhaoLearningFineGrainedBimanual2023} adopt a different strategy, whereby the robot controller interleaves chunk prediction \( \actionchunk_t \gets \pi(o_t) \) and chunk consumption \( a_t \gets \textsc{PopFront(\( \actionchunk_t \))} \), and computes a new chunk of actions at every timestep \( t \), to then aggregate the predicted chunks on overlapping sections.
While adaptive---every observation at every timestep \( o_t\) is processed---such an approach relies on running inference continuously, which can be prohibitive in resource-constrained scenarios, such as edge deployments.
A less resource-intensive approach is to entirely exhaust the chunk \( \actionchunk \) before predicting a new chunk of actions, a strategy we refer to as \emph{synchronous} (sync) inference. 
Sync inference allocates computation every \( H_a \) timesteps, resulting in a reduced computational burden (on average) at control time. 
In contrast, sync inference also inherently hinders the responsiveness of robot systems, introducing blind lags due to the robot being \emph{idle} while computing \( \actionchunk \).

\begin{figure}
    \centering
    \begin{minipage}[t]{\textwidth}
        \centering
        \includegraphics[width=0.9\textwidth]{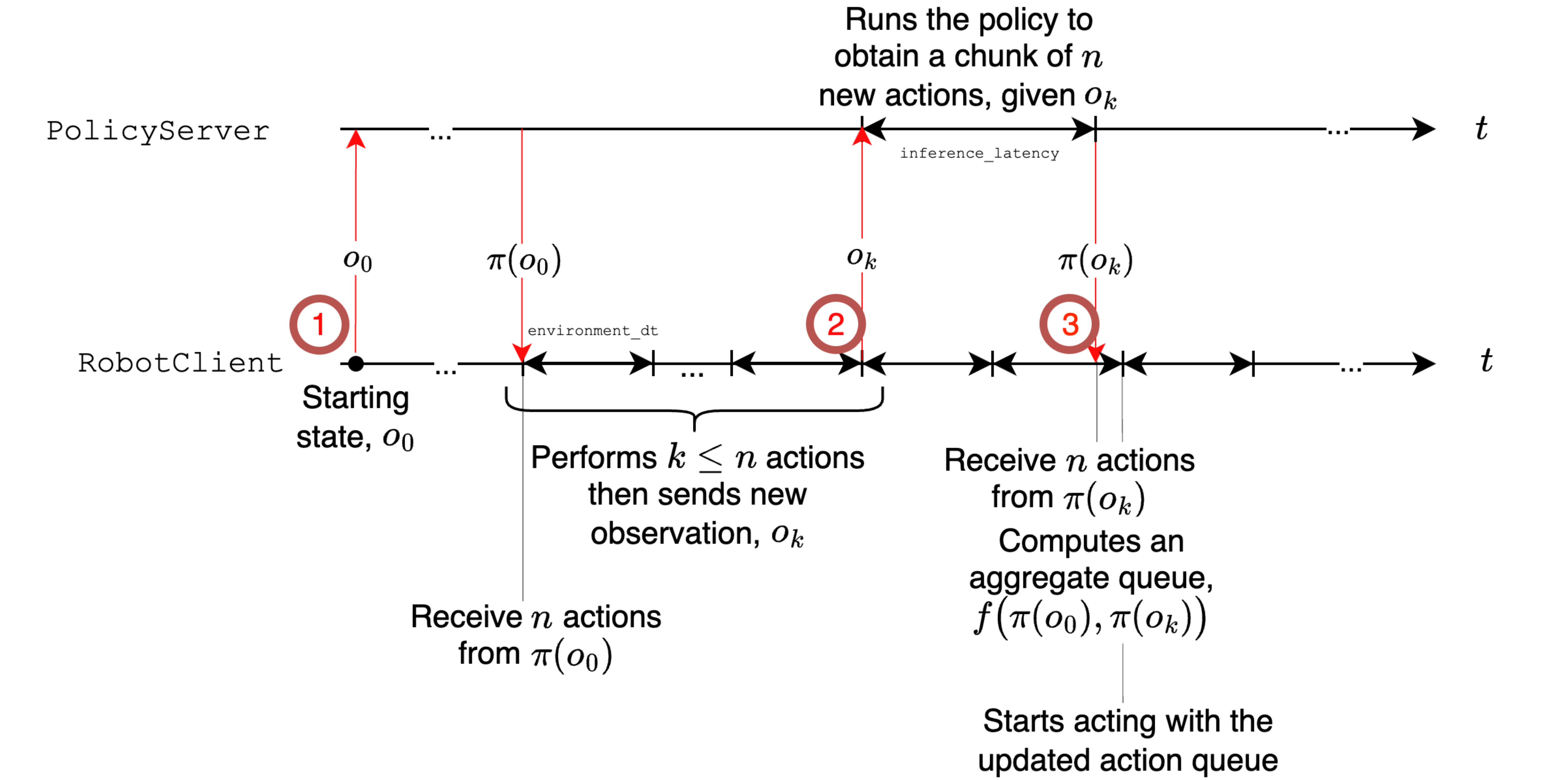}
        \caption{\textbf{Asynchronous inference}. Illustration of the asynchronous inference stack. Note that the policy can be run on a remote server, possibly with GPUs.}
        \label{fig:ch4-async-inference}
    \end{minipage}
    \vspace{-0.6cm}
\end{figure}

One can use the fact that policies output multiple actions at the same time to directly (1) the lack of adaptiveness and (2) the presence of lags at runtime by decoupling action chunk \emph{prediction} \( \actionchunk \) from action \emph{execution} \( a_t \gets \textsc{PopFront}(\actionchunk_t) \).
This decoupled stack, which we refer to as \emph{asynchronous} (async) inference (\ref{alg:async-inference}), also enables optimized inference by allowing action-chunk inference to run on a separate machine, typically equipped with better computational resources than the ones onboard a robot.
In async inference, a \( \textsc{RobotClient} \) sends an observation \( o_t \) to a \( \textsc{PolicyServer} \), receiving an action chunk \( \actionchunk_t \) once inference is complete (Figure~\ref{fig:ch4-async-inference}).
In this, we avoid execution lags by triggering chunk prediction while the control loop is still consuming a previously available chunk, aggregating the previous and incoming chunks whenever the latter is available to the \( \textsc{RobotClient} \).
In turn, async-inference tightens the loop between action prediction and action execution efficienty, by increasing the frequency at which observations are processed for chunk prediction while not running inference at every timestep.
Crucially, decoupling action prediction from action execution also allows to allocate more computational resources on a remote policy server sending actions to the robot client over the network.

\begin{algorithm}
  \caption{Asynchronous inference control-loop}
  \label{alg:async-inference}
  \begin{algorithmic}[1]
    \State \textbf{Input:} horizon \( T \), chunk size \( H_a \), threshold \( g\in[0,1] \)
    \State \textbf{Init:} capture \( o_0 \); send \( o_0 \) to \textsc{PolicyServer};
           receive \( \actionchunk_0 \gets \pi(o_0) \)
    \For{\( t \) \textbf{to} \( H_a \)}
        \State \( a_t \gets \textsc{PopFront}(\actionchunk_t) \)
        \State \textsc{Execute}(\( a_t \)) \Comment{execute action at step \( t \)}
        \If{\( \tfrac{|\actionchunk_t|}{H_a} < g \)} \Comment{queue below threshold}
            \State capture new observation, \( o_{t+1} \)
            \If{\textsc{NeedsProcessing} \( (o_{t+1}) \) } \Comment{similarity filter, or triggers direct processing}
                \State \texttt{async\_handle} \( \gets \textsc{AsyncInfer}(o_{t+1})\) 
                \Comment{Trigger new chunk prediction (non blocking)}
                \State \( \tilde{\actionchunk}_{t+1} \gets \pi(o_{t+1}) \) \Comment{New queue is predicted with the policy}
                \State \( \actionchunk_{t+1} \gets f(\actionchunk_t,\tilde{\actionchunk}_{t+1}) \) \Comment{aggregate overlaps (if any)}
                
            \EndIf
        \EndIf
        \If {\textsc{NotCompleted}(\texttt{async\_handle})}
            \State \( \actionchunk_{t+1} \gets \actionchunk_t \) \Comment{No update on queue (inference is not over just yet)}
        \EndIf
    \EndFor
  \end{algorithmic}
\end{algorithm}

In practice, \emph{async} inference (1) tightens the control loop by capturing observations more often, eliminating idle gaps at runtime (2) and directly allows to run inference on more powerful computational resources than the ones typically available onboard autonomous robotic platforms.
Algorithmically, one can attain (1) on the \textsc{RobotClient}-side by consuming actions from a readily available queue until a given condition on the number of remaining actions in the queue (\(\vert \actionchunk_t \vert / H_a < g \)) is met. When this condition is triggered, a new observation of the environment is captured and sent to the (possibly remote) \textsc{PolicyServer}. 
To avoid redundant server calls and erratic behavior at runtime observations are compared in joint-space, and near-duplicates are dropped.
Two observations are considered near-duplicates if their distance in joint-space falls under a predetermined threshold, \( d_{\text{lim}} \in \mathbb R_+\).
Importantly, should the queue available to the robot client eventually empty out, the most recent observation is processed regardless of similarity.

Interestingly, the behavior of async inference can be studied analytically. First, let \( \ell \) be a random variable modeling the time needed to receive an action chunk \( \actionchunk \) after sending an observation \( o \), i.e. the sum of (1) the time to send across the observation \( o \) between the \textsc{RobotClient} and \textsc{PolicyServer}, \( t_{C \to S}\) (2) the inference latency on the \textsc{PolicyServer}, \( \ell_S \) and (3) the time to send \( \actionchunk \) between the \textsc{PolicyServer} and \textsc{RobotClient}, \( t_{S \to C} \). Under the (reasonable) assumption of independence, \( \mathbb E [\ell] = \mathbb E[t_{C \to S}] + \mathbb E[\ell_S] + \mathbb E[t_{S \to C}] \), which can be further simplified to \( \mathbb E[\ell] \simeq \mathbb E[\ell_S]  \), assuming communication time is (1) equal in both directions and (2) negligible with respect to the inference latency. Second, let \(\Delta t\) be the environment's control cycle. With a real-world frame-rate of 30 frames-per-second (fps), \(\Delta t=33\text{ms}\). Consequently, exhausted queues at runtime---i.e. being idle awaiting for a new chunk---are avoided for \( g \geq \frac{\mathbb E[\ell_S] / \Delta t}{H_a} \). In this, the action queue threshold \( g \) below which to capture and send a new observation for processing plays a major role relatively to the availability of actions to the \( \textsc{RobotClient} \).

\begin{figure}
    \centering
    \begin{minipage}[t]{0.99\textwidth}
        \centering
        \includegraphics[width=\textwidth]{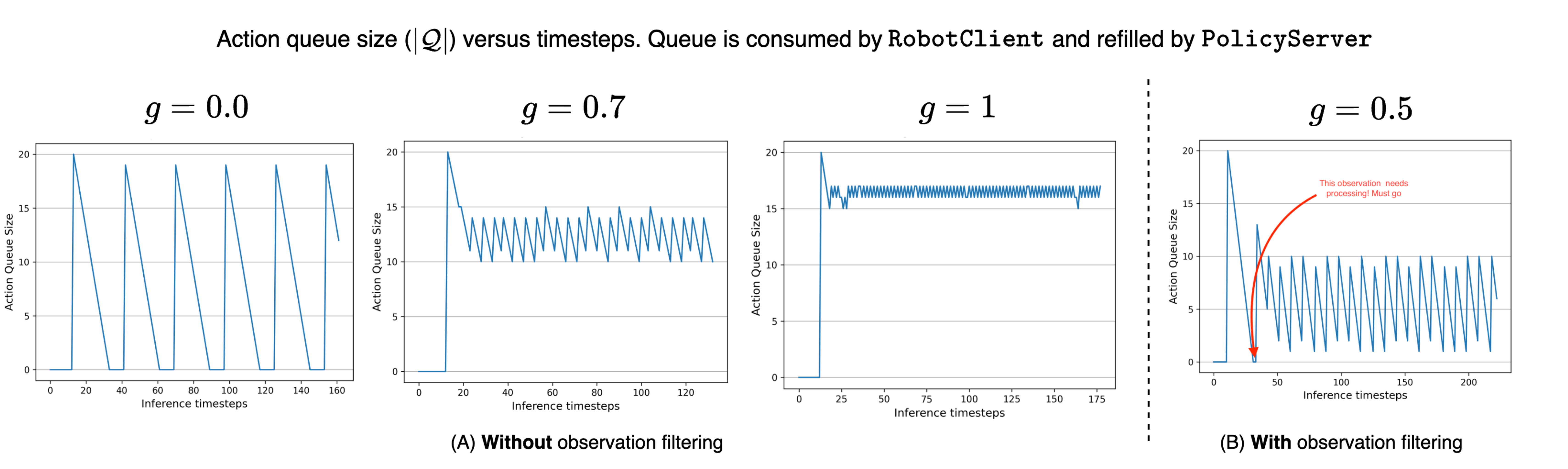}
        \caption{Action queue size evolution at runtime for various levels of \( g\) when (A) not filtering out observation based on joint-space similarity and (B) filtering out near-duplicates observation, measuring their similarity in joint-space.}
        \label{fig:ch4-queues}
    \end{minipage}
\end{figure}

Figure~\ref{fig:ch4-queues} illustrates how the size of the action chunk \(\lvert \actionchunk_t \rvert\) evolves over time for three representative values of \(g\), detailing the following key scenarios:
\begin{itemize}
    \item \textbf{Sequential limit \((g=0)\).} The client drains the entire chunk before forwarding a new observation to the server. During the round-trip latency needed to compute the next chunk, the queue is empty, leaving the robot \emph{incapable of acting}.  This reproduces the behavior of a fully sequential deployment and results in an average of \( \mathbb E[\ell_S] \) idle seconds.
    \item \textbf{Asynchronous inference \((g \in (0,1))\).} Allowing the client to consume a \(1-g\) fraction of its available queue \( \actionchunk_{t-1}\) \emph{before} triggering inference for a new action queue \( \actionchunk_{t} \), computation is amortized while keeping the queue from emptying. The overlap between successive chunks provides a buffer against modeling errors without the full cost of the \(g=1\) regime. The updated queue \( \actionchunk_t\) is obtained aggregating queues on the overlapping timesteps between \( \actionchunk_{t-1}\) and the incoming \(\tilde{\actionchunk}_{t}\).
    \item \textbf{Sync-inference limit \((g=1)\).}  As an extreme case, and in keeping with~\citet{zhaoLearningFineGrainedBimanual2023}, an observation is sent at \emph{every} timestep. The queue is therefore almost always filled, with only a minor saw-tooth due to \(\Delta t/\mathbb E[\ell_s] < 1\). While maximally reactive, this setting incurs one forward pass per control tick and can prove prohibitively expensive on limited hardware. Importantly, because the client is consuming actions while the server computes the next chunk, the available queue never gets entirely filled.
\end{itemize}

Figure~\ref{fig:ch4-queues} emphasizes the trade-off governed by \(g\): small values of \( g \) result in idle periods, whereas \(g\approx 1\) assumes a highly accurate model and pays a significant compute price. 
In practice, choosing \(g\in(0,1)\) allows to strike a balance between reactivity against resource budgets.
If not for the aforementioned similarity filter, the \( \textsc{RobotClient} \) would send observations for processing every \( (1 - g) H_a \cdot \Delta t\) seconds, receiving a new chunk of actions every \( (1 - g) H_a \cdot \Delta t + \mathbb E[\ell_S] \), on average. 
The presence of the filter for observation similarity dilates this processing time, and serves the scope of avoiding the robot stalling due to the queue being constantly integrated with an incoming, nearly identical, action chunk. 
In particular, Figure~\ref{fig:ch4-queues} results in a queue which is filled with incoming actions \emph{unless} near-duplicate observations are filtered out from the processing pipeline. 
For clarity, the red arrow in~\ref{fig:ch4-queues} highlights a timestep where the observation similarity mechanism is bypassed, forcing a (nearly identical) observation to be processed as the queue results empty.

\subsubsection{Code Example: Using Async Inference}

\begin{pbox}[label={ex:spinning-up-server}]{Spinning up a Remote Server \\ \url{https://github.com/fracapuano/robot-learning-tutorial/blob/main/snippets/ch4/05_policy_server.py}}
\begin{lstlisting}[language=python]
from lerobot.async_inference.configs import PolicyServerConfig
from lerobot.async_inference.policy_server import serve

host = ...  # something like "127.0.0.1" if you're exposing to localhost
port = ...  # something like 8080

config = PolicyServerConfig(
    host=host,
    port=port,
)
serve(config)
\end{lstlisting}
\end{pbox}

\begin{pbox}[label={ex:latching-a-robot-client}]{Attaching a Robot Client \\ \url{https://github.com/fracapuano/robot-learning-tutorial/blob/main/snippets/ch4/06_robot_client.py}}
\begin{lstlisting}[language=python]
import threading
from lerobot.robots.so100_follower import SO100FollowerConfig
from lerobot.cameras.opencv.configuration_opencv import OpenCVCameraConfig
from lerobot.async_inference.configs import RobotClientConfig
from lerobot.async_inference.robot_client import RobotClient
from lerobot.async_inference.helpers import visualize_action_queue_size

# these cameras must match the ones expected by the policy (use lerobot-find-cameras)
# check the config.json on the Hub for the policy you are using to see the expected camera specs
camera_cfg = {
    "top": OpenCVCameraConfig(index_or_path=0, width=640, height=480, fps=30),
    "side": OpenCVCameraConfig(index_or_path=1, width=640, height=480, fps=30)
}

# # find ports using lerobot-find-port
follower_port = ...  # something like "/dev/tty.usbmodem58760431631"

# # the robot ids are used the load the right calibration files
follower_id = ...  # something like "follower_so100"

robot_cfg = SO100FollowerConfig(
  port=follower_port,
  id=follower_id,
  cameras=camera_cfg
)

server_address = ...  # something like 127.0.0.1:8080 if using localhost

# 3. Create client configuration
client_cfg = RobotClientConfig(
    robot=robot_cfg,
    server_address=server_address,
    policy_device="mps",
    policy_type="smolvla",
    pretrained_name_or_path="fracapuano/smolvla_async",
    chunk_size_threshold=0.5,  # g
    actions_per_chunk=50,  # make sure this is less than the max actions of the policy
)

# 4. Create and start client
client = RobotClient(client_cfg)

# 5. Provide a textual description of the task
task = ...  

if client.start():
    # Start action receiver thread
    action_receiver_thread = threading.Thread(target=client.receive_actions, daemon=True)
    action_receiver_thread.start()

    try:
        # Run the control loop
        client.control_loop(task)
    except KeyboardInterrupt:
        client.stop()
        action_receiver_thread.join()
        # (Optionally) plot the action queue size
        visualize_action_queue_size(client.action_queue_size)
\end{lstlisting}
\end{pbox}

\newpage
\section{Generalist Robot Policies}
\label{sec:learning-foundation}

\epigraph{\textit{Specialization is for insects}}{Robert A. Heinlein}

\begin{tldr}
Openly available, large-scale datasets and the development of stable-to-train, expressive and efficient architectures fostered research on the development of generalist robot policies that can operate across embodiment and tasks.
\end{tldr}

The advent of large models trained on internet-scale datasets has drastically influenced fields like Computer Vision (CV) and Natural Language Processing (NLP), shifting the previously task-specific paradigm towards combining (1) an initial, task-agnostic large-scale pre-training stage and a (2) task-specific, adjustment phase.
This \emph{pre-train-and-adaptat} paradigm has now largely replaced more classic approaches consisting of task-specific data collection, curation and model training in many subdomains within CV and NLP, and it is motivated by the main drawback of limited scalability for \emph{task-specific approaches}, which have been traditionally more labor intensive.
Factors including (1) the advancements in generalist models learned with self-supervision for perception~\citep{oquabDINOv2LearningRobust2024} or semantic understanding~\citep{devlinBERTPretrainingDeep2019} and (2) the popularization of collective efforts to aggregate large-scale openly available datasets~\citep{oneillOpenXEmbodimentRobotic2025,khazatskyDROIDLargeScaleInTheWild2025} are increasingly pushing the field of robot learning towards the pre-train-and-adapt paradigm.
This shift taps into the long-standing challenge of developing generalist robot policies, and holds the premise to surpass traditionally siloed approaches to robotics problems and develop a \emph{foundation robotics model}.
While Section~\ref{sec:learning-imitation} introduced methods for learning \emph{single-task policies} such as ACT or Diffusion Policy, in this section we present advancements in developing \emph{generalist, multi-task, policies}, capable of performing a wide range of tasks across different environments and embodiments, and guided by unstructured instructions typically given in plain, natural language.

\begin{figure}
    \centering
    \includegraphics[width=0.9\textwidth]{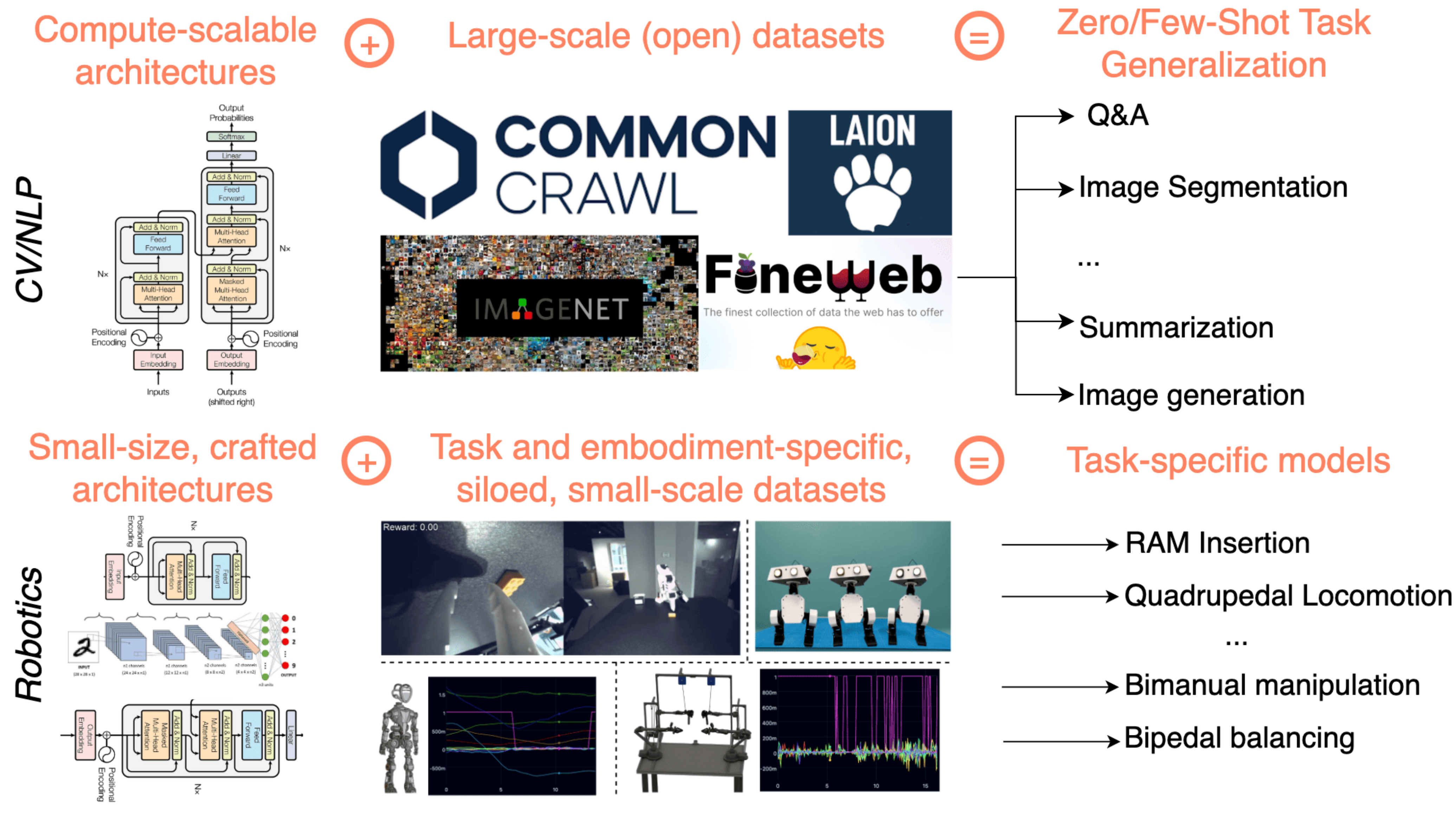}
    \caption{Fields within ML such as Computer Vision and NLP converged on the development of foundation models, trained on a variety of large scale models and capable to perform multiple downstream tasks (top). Conversely, robotics suffered from limited standardization in terms of the architectures used, and siloed, task specific datasets, incurring in a high degree of fragmentation which traditionally hindered the development of generalist models for robotics in favour of task-specific models (bottom).}
    \label{fig:ch5-ml-vs-robotics-foundation}
\end{figure}

\subsection{Preliminaries: Models and Data}
The remarkable success of foundation models in NLP and CV seems to be increasingly predicated on two core principles: architectural innovation and (joint) data-compute scaling.
Indeed, the transformer architecture proved very effective in capturing long-range dependencies in a variety of data formats, and its stability and expressivity made it the \emph{de facto} standard for modern large-scale models trained on internet-scale datasets.
However, in stark contrast with large-scale NLP and CV datasets~\citep{raffelExploringLimitsTransfer2023,ImageNet_VSS09}, robotics has historically developed around small, task-specific datasets. 
In turn, this traditionally hindered scalability across problems as well as results, posing concrete challenges to developing general-purpose robot learning algorithms.
Indeed, differently from the wealth of relatively readily-available task-agnostic text and images datasets on the internet, robotics data is \emph{intrinsically embodied} and thus task-specific: datasets collected for \emph{manipulation} differ significantly from \emph{locomotion}.
In particular, since each expert trajectory is tied to a specific robot platform and the operating conditions of its environment and task, data heterogeneity has long posed a \emph{methodological} challenge for scaling robotics datasets via aggregation.
Further, datasets consisting of expert demonstrations are (1) intrinsically more expensive to collect and (2) notoriously heterogeneous---different human experts may perform the same task in very different.
Beyond this, heterogeneity also raises \emph{conceptual} issues: naively mixing data across embodiments can induce negative transfer, as control strategies developed in isolation for different robot systems in different environments may even conflict when combined.
Thus, the high degree of fragmentation of robotics datasets and tasks has traditionally led to the development of \emph{specialist} policies, trained on small, task-specific datasets, developed to perform well at their designated task but that fail to generalize to new deployment scenarios (Figure~\ref{fig:ch5-ml-vs-robotics-foundation}).

\begin{figure}
    \centering
    \includegraphics[width=0.8\textwidth]{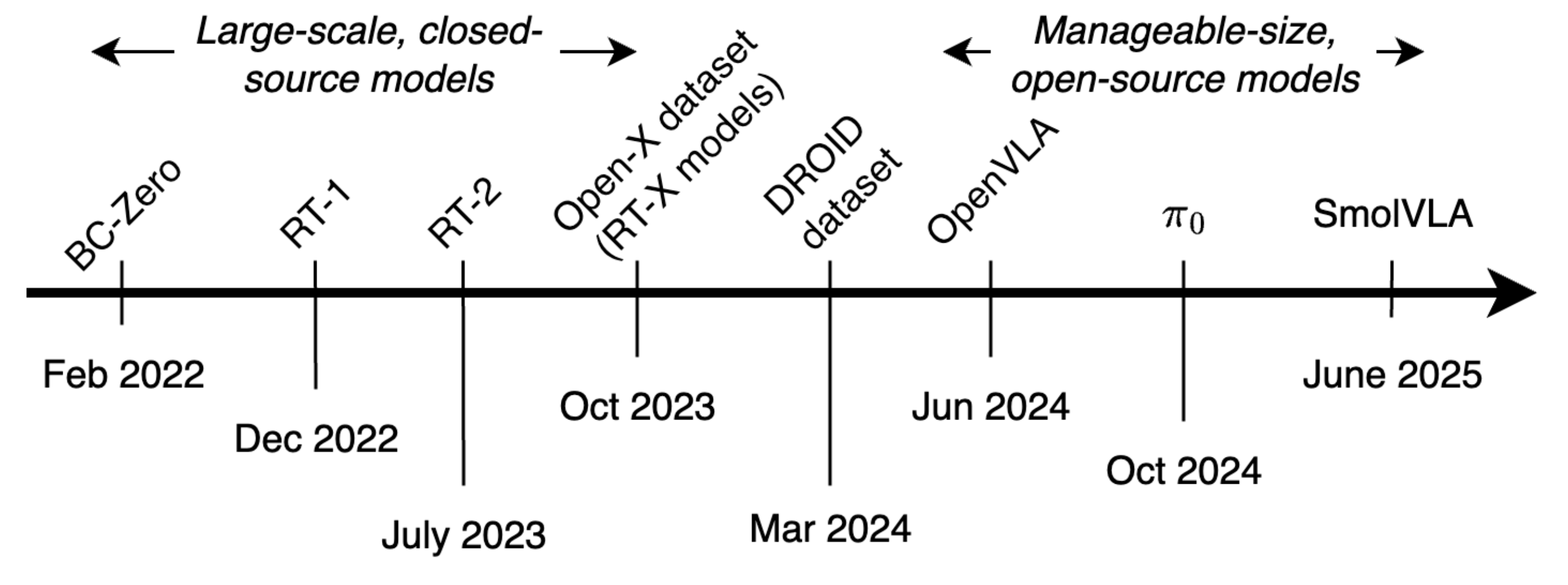}
    \caption{Early efforts in the development of generalist models for robotics include BC-Zero~\citep{jangBCZZeroShotTask2022}, RT-1~\citep{brohanRT1RoboticsTransformer2023}, and RT-2~\citep{brohanRT2VisionLanguageActionModels2023}: large scale models trained on thousands of demonstrations. The open release of the Open-X~\citep{oneillOpenXEmbodimentRobotic2025} and DROID datasets~\citep{khazatskyDROIDLargeScaleInTheWild2025} fostered the development of open source models: OpenVLA~\citep{kimOpenVLAOpenSourceVisionLanguageAction2024}, \pizero~\citep{black$p_0$VisionLanguageActionFlow2024} and SmolVLA~\citep{shukorSmolVLAVisionLanguageActionModel2025}.}
    \label{fig:ch5-generalist-policies-timeline}
\end{figure}

Driven by the goal of developing generalist robot policies, the research community has increasingly explored how insights and techniques from other areas of ML can be integrated into robotics.
Figure~\ref{fig:ch5-generalist-policies-timeline} shows a timeline of some of the most popular contributions attempting at developing generalist policies.
Starting from BC-Zero, a latent variable model trained on 25k+ demonstrations, the field has now evolved into \( \pi_0 \), a transformer-based model trained on 10M+ demonstrations and exhibiting strong few-shot capabilities across tasks and embodiments.
In between, Robotics Transformer 1 (RT-1)~\citep{brohanRT1RoboticsTransformer2023} represented a significant step in the direction of developing a generalist robot policies over prior work including (1) BC-Zero~\citep{jangBCZZeroShotTask2022} and (2) Gato~\citep{reedGeneralistAgent2022}, in that~\citet{brohanRT1RoboticsTransformer2023} use a much larger and diverse set of training tasks compared to both BC-Zero and Gato.
In particular, RT-1 uses a transformer architecture, and is trained on as many as 130k human-recorded trajectories collected over 13 robots and over 17 months.
RT-1 learns to process a history of camera images and a natural language instruction, and feeds the resulting sequence of high-dimensional tokens to a transformer, trained using a \emph{classification loss on a discretized actions space} consisting of six different 256-bins, one for each joint of a 6-dof robotic arm.

In a follow-up work, the same group of authors propose a modified method to learn generalist models, leveraging (1) a more powerful architecture and (2) scaling up the dataset used~\citep[RT-2]{brohanRT2VisionLanguageActionModels2023}. 
In RT-2,~\citet{brohanRT2VisionLanguageActionModels2023} propose inheriting internet-scale semantic knowledge from large-scale multi-modal datasets to learn a single, \emph{unified model} for robotics control.
Such a model, termed \emph{Vision-Language-Action} (VLA) in the original RT-2 paper, effectively casts robot control as a language-modeling problem, and in particular as a Visual Question-Answering (VQ\&A) task, in which the output token space used to represent \emph{textual tokens} is shared with the \emph{8-bits tokens} used to represent the 256 (\( 2^8 \)) actuation levels of a 6-dof robot.
In their work,~\citet{brohanRT2VisionLanguageActionModels2023} propose co-fine-tuning large-scale VLMs such as PaLIX~\citep{chenPaLIXScalingMultilingual2023} or PaLM-E~\citep{driessPaLMEEmbodiedMultimodal2023} on a mix of (1) web and (2) robotics data, complementing VQ\&A training with robotics-specific signal, and learning to directly output robot actions in a shared token space for visual and language inputs.
In their work, the authors claim using large models trained on internet-scale data as backbones for VLAs allows models to tap into the rich semantic knowledge embedded in the VLM's parameters, interpreting instructions and unseen objects by connecting them to concepts acquired while pre-training.
For instance,~\citet{brohanRT2VisionLanguageActionModels2023} show that while RT-2 has never been explicitly trained to repurpose tools for a \emph{hammering} task, it can still combine its semantic understanding of images, so that when asked which object between (1) a piece of paper, (2) a pair of headphones or (3) a rock may be used instead of a hammer, it correctly answers (3).

Traditionally, research efforts revolved around not only training models, but also proposing datasets for the community, a costly and time-consuming process.
Due to the aforementioned embodiment gap, the data used in research efforts in robot learning have traditionally proved rather fragmented, tailored to the specific task considered by the specific group of researchers who collected it, which ultimately hindered integration.
The Open X-Embodiment project~\citep{oneillOpenXEmbodimentRobotic2025} was a landmark collaboration effort to address data fragmentation, by curating the aggregation of 60 \emph{existing} robotics datasets from 22 different robot embodiments and 21 institutions across the world, and resulted in a total 1.4M of cross-embodiments, cross-tasks, openly-available trajectories.
Besides the contribution of an aggregate, large scale dataset,~\citet{oneillOpenXEmbodimentRobotic2025} also demonstrated significant positive transfer \emph{across tasks and embodiments}, showing that \highlight{a single model trained on multi-embodiment data can outperform specialist models} trained on their respective single-embodiment datasets.
The Distributed Robot Interaction Dataset (DROID)~\citep{khazatskyDROIDLargeScaleInTheWild2025} represents another significant step towards addressing the problem of scarse and disaggregated data in robot learning, providing a unique dataset consisting of 75k+ human demonstrations collected in realistic (\emph{in-the-wild}) manipulation settings, providing another cornerstone for building general-purpose robot policies.
Recently, foundational datasets curated through large, centralized efforts, are increasingly complemented by decentralized, community-driven contributions of robotics data.
Software libraries like \lerobot~have been instrumental in enabling decentralized collection of large amounts of data, providing the infrastructure for researchers and practitioners to easily contribute trajectories from a wide range of embodiments, democratizing data access via distributed collection.

Despite these advancements, the success of large, proprietary models like RT-1 and RT-2, highlighted a growing accessibility gap in robotics research, as training and deploying large-scale robotics foundation models requires computational resources simply unattainable for most research institutions. 
The OpenVLA project~\citep{kimOpenVLAOpenSourceVisionLanguageAction2024} emerged in direct contrast to traditionally closed-source efforts to develop VLAs.
In particular,~\citet{kimOpenVLAOpenSourceVisionLanguageAction2024} trained OpenVLA by exclusively leveraging openly available data (970k+ trajectories from the Open-X dataset), and openly shared their training recipes alongside the model weights.
Architecturally, OpenVLA integrates a pre-trained vision encoder to project visual tokens into the embedding space of the Llama2-7B~\citep{touvronLlama2Open2023} language-model backbone.
The language model backbone is then used to predict \emph{discrete action tokens} over 256 activation levels.

\begin{figure}
    \centering
    \includegraphics[width=0.9\textwidth]{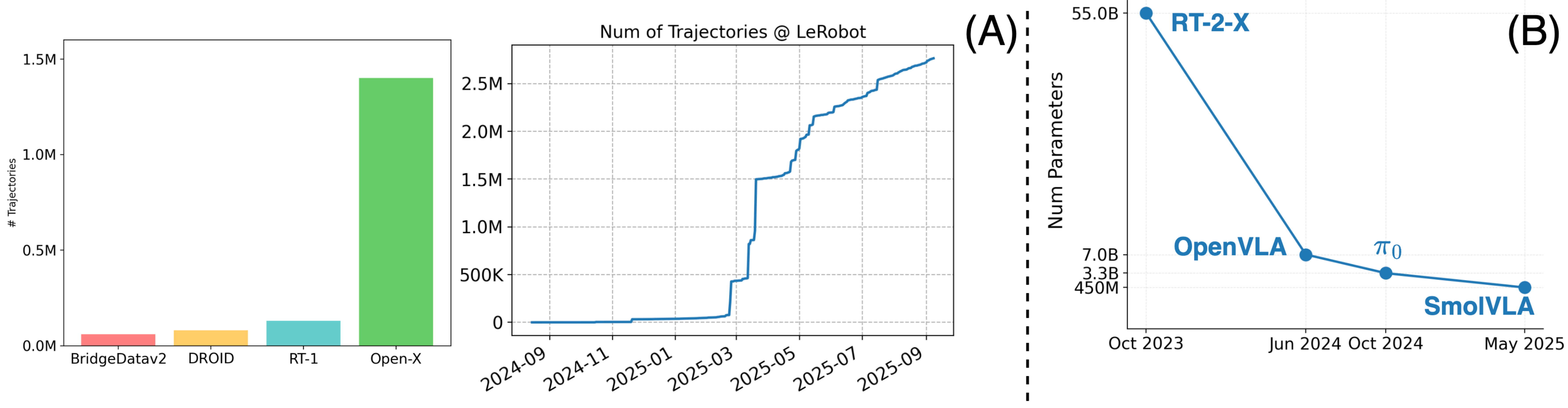}
    \caption{Robot learning is undergoing a paradigmatic shift: centralized data collections (A, left) are increasingly larger, often comprising millions of demonstrations, while (A, right) decentralized data collection efforts are becoming an alternative for large scale data collection. (B) Generalist models are also becoming increasingly smaller and easier to run on limited hardware.}
    \label{fig:ch5-trends}
\end{figure}

Figure~\ref{fig:ch5-trends} shows the current trends in robot learning in terms of size and nature of the robotics datasets contributed, together with the size and accessibility of the available models.
As datasets collected via centralized, cross-institutions cooperation of increasing size are made available for the research community, decentralized datasets collected by individual researchers and practitioners also gained traction, closing the gap with academic benchmarks thanks to community-contributed datasets.
Further, models used across tasks and embodiments are increasingly becoming much more compute-efficient, and as a result the models' size has been consistently reducing over time, with consequent gains for autonomous robots in real-world, resource-constrained environments.

\subsection{VLAs}
Modern recipes to train large scale VLAs extend early efforts to learn foundation models from large amounts of data via BC, introducing significant advancements concerning both architectural and procedural aspects.
From an architectural perspective, modern VLAs such as \pizero~\citep{black$p_0$VisionLanguageActionFlow2024} leverage a \emph{unified transformer model} for efficiency of computation, while maintaining specialized sub-components within the model for visual perception and action prediction, enabling cross-task performance via language conditioning.
Crucially, modern VLAs including\pizero~\citep{black$p_0$VisionLanguageActionFlow2024} and SmolVLA~\citep{shukorSmolVLAVisionLanguageActionModel2025} adopt \emph{unified} transformer models employing disjoint set of weights (\emph{experts}) for both compute-efficient visual-semantic understanding as well as control.
Procedurally, VLAs complement advanced Vision-Language Model (VLM) backbones with action-specific modules (1) adopting mid-sized \emph{action experts} to model continuous actions distributions \( p (a_{t:t+H_a} \vert o_t) \)---avoiding discrete action tokens entirely---and (2) relying on~\emph{action chunking}~\citep[Section~\ref{sec:learning-imitation}]{zhaoLearningFineGrainedBimanual2023} as a strategy to reduce error compounding when predicting multiple actions learning from inherently non-i.i.d. data, such as demonstration data.

These architectural and procedural innovations present three benefits over task-specific methods. 
First, developing architectures that exploit internet-scale pre-trained backbones allows to fully capitalize on the vast world knowledge and skills state-of-the-art VLMs exhibit, preventig models from needing to learn visual, linguistic and semantic concepts from scratch.
Second, using generative models for continuous action distributions allows to learn rich, multimodal data distributions, a much more likely scenario in the big-data regime which is typically tackled while developing generalist policies.
Further, introducing separate components for perception and action planning enable using Mixture of Experts (MoE) architectures~\citep{fedusReviewSparseExpert2022}, which are often more efficient to run---a key feature for models deployed in real-world scenarios.
This new paradigm has been at the core of some of the most capable generalist policies developed to date, capable to few-shot adapt to novel tasks and to perform highly dexterous manipulation tasks ranging from end-to-end folding laundry to bussing tables~\citep{black$p_0$VisionLanguageActionFlow2024}.

\subsubsection{VLMs for VLAs}
VLMs are designed to handle both visual and textual modalities, most commonly by taking both images and text as inputs, generating text conditioned on the visual context.
Recent advances in VLMs have been driven by the success of LLMs, with many approaches building upon pretrained LLMs and adopting similar training paradigms to the ones used in language modeling.
Typically, VLMs~\citep{alayracFlamingoVisualLanguage2022,laurenconWhatMattersWhen2024,linVILAPretrainingVisual2024} are constructed by integrating a pretrained vision encoder~\citep{radfordLearningTransferableVisual2021,zhaiSigmoidLossLanguage2023,finiMultimodalAutoregressivePretraining2024} with a pretrained LLM~\citep{grattafioriLlama3Herd2024,jiangMistral7B2023}.
Training then proceeds in multiple multimodal stages, beginning with a large-scale pretraining on datasets containing image-text pairs~\citep{LAION-COCO,kakaobrain2022coyo700m} and interleaved vision-language corpora~\citep{OBELICS,MMC4}, all followed by a supervised fine-tuning stage on instruction-tuning datasets~\citep{LLaVA-1.5,tong2024cambrian,laurenconWhatMattersWhen2024}.
The inherent multimodal nature of VLMs enables them to jointly reason over vision and language. 
Pre-training on vast internet-scale datasets allows these models to associate visual patterns with textual descriptions, thereby acquiring a rich semantic understanding of the world---knowledge about objects, their properties, and relationships---without explicit supervision for each concept. 
In turn, integrating VLMs as the perceptual backbone for VLAs allows the latter to inherit rich, contextual world knowledge from the VLM, sidestepping the need to re-learn visual and semantic representations.
In principle, this also allows the robot to ground high-level natural language instructions in its visual context, and possibly recognize objects by connecting them to the pre-trained concepts absorbed during pre-training, improving on the possibility to generalize to novel scenarios.

Recently, compute efficiency has also become a central focus in multi-modal research. 
Several works aim to reduce training costs by using smaller, more diverse datasets~\citep{LLaVA-1.5,InstructBLIP,bai2025qwen25vl,zhu2024minigpt,tong2024cambrian}, training smaller-scale models~\citep{marafiotiSmolVLMRedefiningSmall2025, moondream,minicmpv2024}, or by adapting pretrained unimodal models by tuning only a small subset of parameters~\citep{shukor2023epalm,vallaeys2024improveddepalm,MAPL,FROMAGe,tsimpoukelli2021multimodalfrozen,BLIP-2}.
While the majority of VLM research focuses on image and text modalities, recent work has also demonstrated that similar techniques can be extended to integrate additional modalities, such as video and audio~\citep{wang2025internvideo2,liu2024kangaroo,zhang2025videollama,kong2024audioflam}---a particularly promising direction of research for robotics applications, where multiple sensor modalities can be integrated effectively. 
This trend towards efficiency is paramount for robotics applications, where policies must operate under the stringent constraints of real-world deployment. 

\subsection{\( \pi_0 \)}

\pizero~\citep{black$p_0$VisionLanguageActionFlow2024} introduce a VLA consisting of a MoE architecture consisting of (1) a pre-trained VLM backbone (Gemma 2.6B~\citep{teamGemma2Improving2024}) and (2) a dedicated action expert used to generate continuous actions via flow matching.
Images and language are embedded with PaliGemma, a VLM merging independently encoded visual and textual features deep in the network (\emph{late-fusion}), while proprioceptive state and actions chunks are routed to a smaller \emph{action expert}, initialized from scratch.
The two separate experts communicate via self-attention layers, but maintain disjoint weights to obtain query, key and values matrices at each layer, maintaining specialization while efficiently allocating computation.

\begin{figure}
    \centering
    \includegraphics[width=0.9\textwidth]{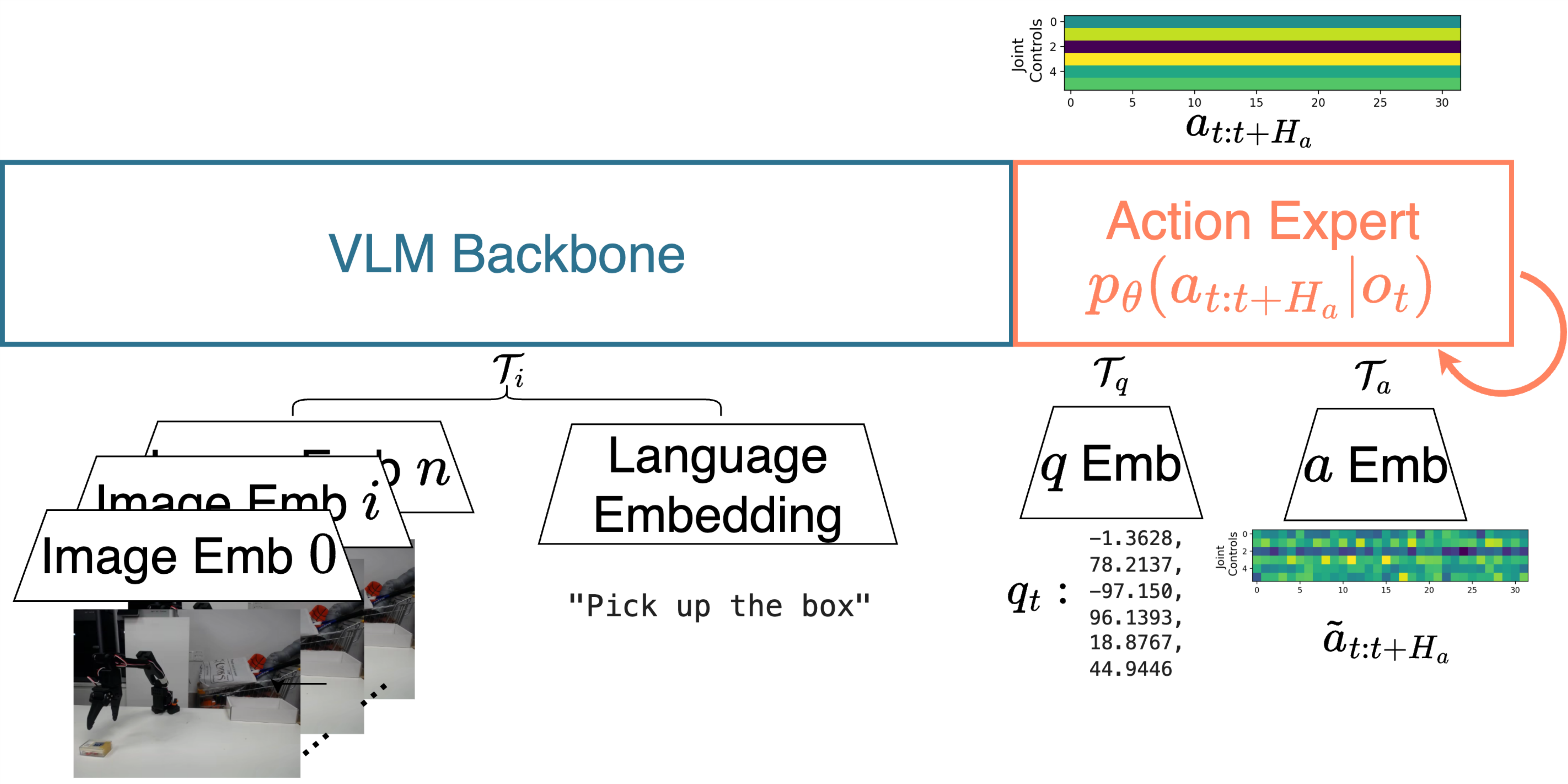}
    \caption{The \pizero~architecture, as in~\citet{black$p_0$VisionLanguageActionFlow2024}. Vision and language tokens are routed to a VLM backbone which is prevented from attending robot proprioperceptive states and action tokens, which are instead routed to a smaller subset of weights within the architecture referred to as "action expert". The architecture is trained with Flow Matching on 10M+ trajectories from a mixture of closed and openly available datasets.}
    \label{fig:ch5-pi0}
\end{figure}

Concretely, \( \pi_0 \) is a single, unified transformer with two disjoint sets of weights \( \phi, \theta\). 
A larger VLM backbone \( f_\phi \) initialized from Gemma 2.6B processes multiple image frames obtained from multiple cameras points \( [\{ I_t \}_{t=1}^n] \), as well as a language instruction \([\ell_t]\) used to describe the task considered.
Concurrently, a 300M-parameter \emph{action expert} based on a similar transformer architecture is used to process both the robot proprioperceptive state \(q_t\) and an action chunk \(a_{t:t+H_a}\) (Figure~\ref{fig:ch5-pi0}).
The different expert networks operate separately in processing the respective inputs and turn them into query, key and value matrices, and only share information between each other via self-attention layers.
The outputs from the VLM backbone are disregarded, while the vector field regressed by the action expert is used to iteratively refine the action process.
In particular, \pizero~uses a \emph{blockwise causal attention mask} over tokens belonging to three separate blocks: (1) image and language tokens \(\mathcal T_i \)  obtained from \([\{ I_t \}_{t=1}^n, \ell_t]\), (2) proprioperceptive tokens \(\mathcal T_q \) obtained from \(q_t\), and (3) the action tokens \( \mathcal T_a \) for items in the chunk \(a^{\tau}_{t:t+H_a}\) at time \( \tau \) in the flow-matching process.
Notably, \emph{within} each block the attention operations are bidirectional, while \emph{across} blocks, future blocks are masked out.
Formally, this corresponds to using an attention mask like:
\begin{equation*}
    \mathbf{A} =
    \bordermatrix{
              & \mathcal{T}_i & \mathcal{T}_q & \mathcal{T}_a \cr
    \mathcal{T}_i & \mathbf{1} & \mathbf{0} & \mathbf{0} \cr
    \mathcal{T}_q & \mathbf{1} & \mathbf{1} & \mathbf{0} \cr
    \mathcal{T}_a & \mathbf{1} & \mathbf{1} & \mathbf{1} \cr
    },
    \quad \mathbf{1}: \text{Bidirectional Attention}, \ \mathbf{0}: \text{Masked Attention} 
\end{equation*}
Note how \emph{intra}-block directional attention allows tokens to communicate freely, while \emph{inter}-block communication is mediated by the attention mask \(\mathbf{A} \).
\emph{Blockwise causal masking} effectively prevents the pre-trained perception-language tokens from attending to robotics-tokens, likely out of distribution for VLM backbones traditionally trained on large corpora of internet, non-robotics, data.
Crucially, because communication is obstructed between image-language tokens, proprioperceptive tokens and action tokens, one can cache keys and values across denoising steps at runtime time, incuring in a reduced computational footprint and faster inference.

In \pizero, both the VLM backbone and action expert are update using a \emph{flow matching} loss, and in particular are updated minimizing:
\begin{align}
    \mathcal{L}(\phi, \theta) &= 
    \mathbb{E}_{\tau, \epsilon, o_t, a_{t:t+H_a}}\Big[
        \big\Vert 
            v_\theta(\underbrace{\tau a_{t:t+H_a} + (1-\tau) \epsilon}_{\tilde a_{t:t+H_a}},\, o_t,\, \tau)
            - (\epsilon - a_{t:t+H_a})
        \big\Vert^2
    \Big], \label{eq:pi0-loss} \\
    &\tau \sim \mathrm{Beta}_{[0,s]}(1.5,1), \quad
    \epsilon \sim \mathcal{N}(\mathbf{0}, \mathbf{I}), \quad
    o_t, a_{t:t+H_a} \sim \mathcal D \notag
\end{align}
where the two experts parametrized by the separate weights \( \phi, \theta \) interact with each other via self-attention layers only, so that the action expert \( v_\theta \) internal computations also depend on the VLM backbone's parameters \( \phi \).
Importantly,~\citet{black$p_0$VisionLanguageActionFlow2024} minimize eq.~\ref{eq:pi0-loss} over both the multimodal backbone and action expert parameters, thus updating both the internal representations of the VLM and action-expert weights using BC-specific gradients.
In contrast,~\citet{driessKnowledgeInsulatingVisionLanguageAction2025} later show that failing to insulate the VLM knowledge from the flow matching gradients actually harms performance.

At runtime, inference is performed iteratively refining action chunks while numerically forward-integrating the vector field predicted by the action expert,
\begin{equation}
    a_{t:t+H_a}^{\tau + \delta} = a_{t:t+H_a}^{\tau } + \delta v_\theta(a_{t:t+H_a}^{\tau }, o_t)
\end{equation}

\begin{wrapfigure}{r}{0.4\textwidth}
    \vspace{-10pt}
    \centering
    \includegraphics[width=\linewidth]{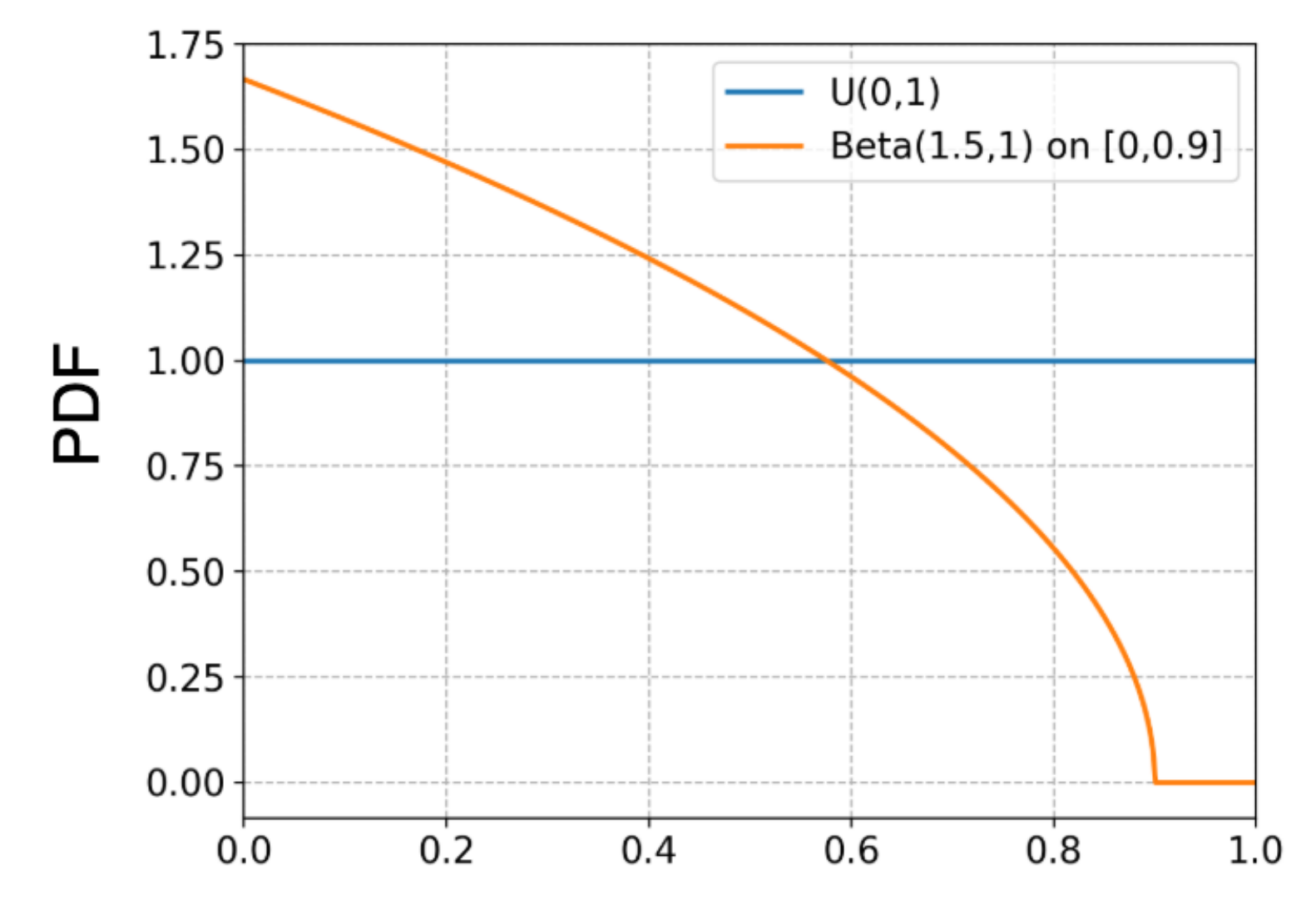}
    \caption{Unlike more traditional flow-matching algorithms, \pizero~uses a modified distribution to sample the timestep \( \tau \) from during training and inference, favouring earlier timestamps corresponding to noisier chunks.}
    \label{fig:ch5-pi0-sampling-timesteps}
\end{wrapfigure}

Flow matching~\citep[Section\ref{sec:ch4-flow-matching}]{lipmanFlowMatchingGenerative2023} can be seen as a continuous time, deterministic generalization of diffusion processes, and has proven effective in modeling highly complex multi-modal distributions, including those over images and video.
In turn, the application of flow matching to large-scale datasets of multiple human behaviors across tasks and embodiments appears rather consequential, particularly considering how it can enable faster inference via a limited number of denoising steps at test time---as few as 10, in \pizero.
In particular, the action expert is implemented as a conditional flow matching model.
Each action token embeds a noisy action \(a_i^{\tau} \in a^\tau_{t:t+H_a}\), alongside a sinusoidal encoding of the \emph{flow process} timestep \(\tau\).
The action expert then leverages full bidirectional attention across the \(H_a\) action tokens provided, and also attends to previous proprioperceptive and image-language tokens.
Interestingly, differently from a standard flow matching pipeline~\citep{lipmanFlowMatchingGenerative2023}, \(\tau\) is \emph{not} sampled from a uniform distribution \(\tau \sim \mathcal U([0,1]) \), but rather obtained from \(\tau \sim \textrm{Beta}(1.5,1) \) defined on the \( [0,s], s<1 \) support (Figure~\ref{fig:ch5-pi0-sampling-timesteps}).

Using such Beta distribution emphasizes higher noise levels during training, a choice~\citet{black$p_0$VisionLanguageActionFlow2024} argue allows \pizero~to focus on learning to reconstruct the mean of the data distribution \( \mathbb E[a_{t:t+H_a} \vert o_t] \) over an identity map during training, in keeping with~\citet{esserScalingRectifiedFlow2024}.
To further optimize performance and reduce inference time,~\citet{black$p_0$VisionLanguageActionFlow2024} propose reducing the support of the timestep distribution to \([0,s], \ s < 1 \), as for any forward-integration step size \( \delta = 1-s \) timesteps above \(s \) are never sampled at inference time.

Besides adopting a MoE architecture with a VLM backbone initialized from a pre-trained model and trained jointly with an action expert via flow matching, \pizero~also relies on a unique pre-training corpus comprising of a mix of proprietary and open data totaling 10M+ trajectories, which in their work~\citet{black$p_0$VisionLanguageActionFlow2024} claim to be the largest dataset used to develop a foundational robotics model to date.
The dataset used to train \pizero---referred to as "the \( \pi \) dataset"---comprises a private, undisclosed portion obtained via expert teleoperation as well as openly available datasets including Open-X and DROID, with only \(\approx 9.1\%\) of the \( \pi \) being openly available.
In the \( \pi \) dataset, open datasets such as DROID and Open-X are complemeneted with expert trajectories consisting of dexterous demonstrations tasks spanning 7 robot configurations and 68 different tasks.
Crucially, \citet{black$p_0$VisionLanguageActionFlow2024} show that pre-training on the \( \pi \) dataset yields a broadly capable base model, which can be adapted via fine-tuning on narrower, higher-quality task data, which induces a fluent multi-stage behavior while retaining robustness.
In particular,~\citet{black$p_0$VisionLanguageActionFlow2024} report that, across a variety of benchmarks, the version of \pizero~pretrained on the \( \pi \) dataset and fine-tuned on extra high-quality data demonstrations \emph{consistently outperforms} a \( \pi_0^{\text{scratch}} \)~baseline trained entirely from scratch for a given specific task, which further underscores the relevance of pretraining on the \( \pi \) dataset.
\citet{black$p_0$VisionLanguageActionFlow2024} do also offer an intuition behind this finding: high-quality demonstrations of a given task tend to omit failure data, which inherently prevents an autonomous agent to learn how to recover from near-failure states.
In turn, robot trained on high-quality data exclusively with BC may as well be entirely incapable to recover from failure.
Conversely, large scale collections of human demonstrations are typically much more diverse (if anything, for their sheer scale), and typically contain rich and diverse information, which may prove suboptimal for any given task when considered in isolation but which proves invaluable in coupling with a small, narrower set of demonstrations.

Lastly,~\citet{black$p_0$VisionLanguageActionFlow2024} present cross-embodiment experiments where they demonstrate \pizero's ability to control both mobile and static manipulator robots with varying arm embodiments.
The emergence of cross-embodiment capabilities is largely to be attributed to the presence of large scale cross-embodiment data in \( \pi \) data mixture, which is in practice handled by \pizero~outputting actions with maximal configuration size across the whole \( \pi \) dataset, and zero-padding robots with fewer dofs.
\pizero~does also rely on exactly three camera views at both training and test time, and uses masked image slots for training and deployment scenarios with fewer cameras.

\subsubsection{Code Example: Using \pizero}
\begin{pbox}[label={ex:using-pizero}]{Using \pizero \\ \url{https://github.com/fracapuano/robot-learning-tutorial/blob/main/snippets/ch5/01_using_pi0.py}}
\begin{lstlisting}[language=python]
import torch

from lerobot.cameras.opencv.configuration_opencv import OpenCVCameraConfig
from lerobot.datasets.utils import hw_to_dataset_features
from lerobot.policies.factory import make_pre_post_processors
from lerobot.policies.pi0.modeling_pi0 import PI0Policy
from lerobot.policies.utils import build_inference_frame, make_robot_action
from lerobot.robots.so100_follower.config_so100_follower import SO100FollowerConfig
from lerobot.robots.so100_follower.so100_follower import SO100Follower

MAX_EPISODES = 5
MAX_STEPS_PER_EPISODE = 20

device = torch.device("mps")  # or "cuda" or "cpu"
model_id = "lerobot/pi0_base"

model = PI0Policy.from_pretrained(model_id)

preprocess, postprocess = make_pre_post_processors(
    model.config,
    model_id,
    # This overrides allows to run on MPS, otherwise defaults to CUDA (if available)
    preprocessor_overrides={"device_processor": {"device": "mps"}},
)

# find ports using lerobot-find-port
follower_port = ...  # something like "/dev/tty.usbmodem58760431631"

# the robot ids are used the load the right calibration files
follower_id = ...  # something like "follower_so100"

# Robot and environment configuration
# Camera keys must match the name and resolutions of the ones used for training!
# You can check the camera keys expected by a model in the info.json card on the Hub
camera_config = {
    "base_0_rgb": OpenCVCameraConfig(index_or_path=0, width=640, height=480, fps=30),
    "left_wrist_0_rgb": OpenCVCameraConfig(index_or_path=1, width=640, height=480, fps=30),
    "right_wrist_0_rgb": OpenCVCameraConfig(index_or_path=2, width=640, height=480, fps=30),
}

robot_cfg = SO100FollowerConfig(port=follower_port, id=follower_id, cameras=camera_config)
robot = SO100Follower(robot_cfg)
robot.connect()

task = ...  # something like "pick the red block"
robot_type = ...  # something like "so100_follower" for multi-embodiment datasets

# This is used to match the raw observation keys to the keys expected by the policy
action_features = hw_to_dataset_features(robot.action_features, "action")
obs_features = hw_to_dataset_features(robot.observation_features, "observation")
dataset_features = {**action_features, **obs_features}

for _ in range(MAX_EPISODES):
    for _ in range(MAX_STEPS_PER_EPISODE):
        obs = robot.get_observation()
        obs_frame = build_inference_frame(
            obs, dataset_features, device, task=task, robot_type=robot_type
        )

        obs = preprocess(obs_frame)

        action = model.select_action(obs)
        action = postprocess(action)
        action = make_robot_action(action, dataset_features)
        robot.send_action(action)

    print("Episode finished! Starting new episode...")
\end{lstlisting}
\end{pbox}

\subsection{SmolVLA}
With VLAs in the early stage of development compared to more mature LLMs and VLMs, much of the progress made on VLAs remains proprietary, with many releases exclusively sharing the weights while withholding the data used, full experimental details and essential methodological components of training.
In constrast with this closed approach, SmolVLA~\citep{shukorSmolVLAVisionLanguageActionModel2025} is an entirely open-source research effort, which aims at democratizing the developments of robotics foundation models by open sourcing the model alongside the data used as well as the training recipes.

\begin{figure}
    \centering
    \includegraphics[width=0.9\textwidth]{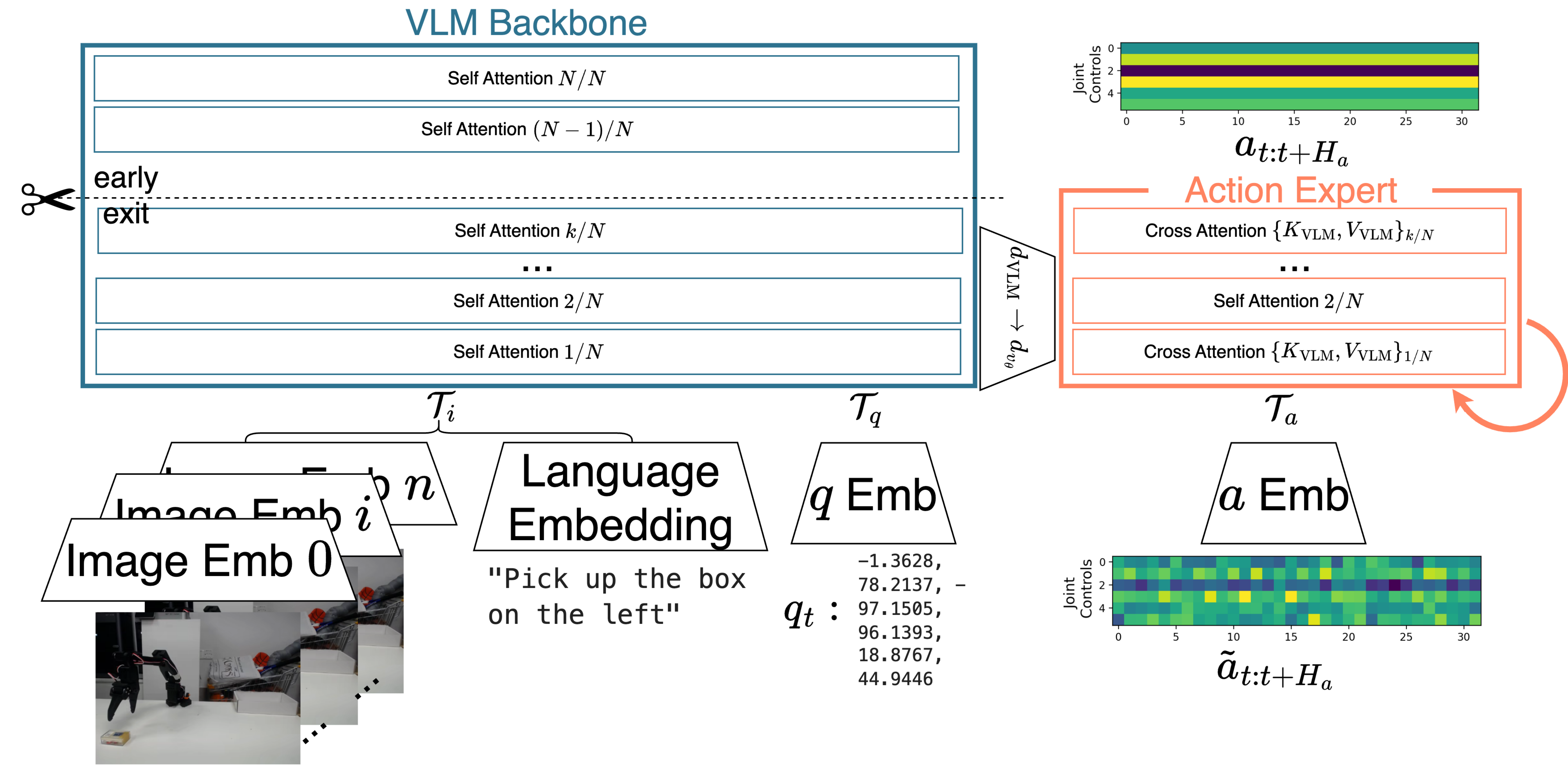}
    \caption{The SmolVLA architecture, as in~\citet{shukorSmolVLAVisionLanguageActionModel2025}. SmolVLA is a compact MoE model trained with flow matching to denoise action chunks. Vision and language tokens are fed to a VLM backbone, and share information with the proprioperceptive and action tokens via the attention mechanism. The attention expert interleaves SA and CA layers for further conditioning on the visual features from the VLM backbone. SmolVLA skips computations and reduces the visual tokens, resulting in 7x less memory usage than \pizero~(450M parameters vs. \pizero's 3.3B).}
    \label{fig:ch5-smolvla}
\end{figure}

While encouraging efforts like \pizero~\citep{black$p_0$VisionLanguageActionFlow2024} demonstrate the feasibility of open VLA systems, they remain (1) large and compute-intensive and (2) dependent on closed datasets collected via centralized efforts on costly robotic platforms, which ultimately hinders the accessibility of the method altogether.
SmolVLA mitigates both these issues by (1) prioritizing a compact, compute-efficient VLA design and (2) targeting community-contributed datasets on accessible robotic platforms such as the SO-100 and SO-101 arms.
Similarly to \pizero, SmolVLA (Figure~\ref{fig:ch5-smolvla}) employs a MoE architecture combining a pretrained VLM backbone with a dedicated action expert, and trains with flow matching.
To ensure efficiency and accessibility, SmolVLA adopts SmolVLM-2~\citep{marafiotiSmolVLMRedefiningSmall2025} as its VLM backbone, considering SmolVLM-2's reduced size and capability to process multiple image inputs alongside text items.
SmolVLM-2 uses SigLIP~\citep{zhaiSigmoidLossLanguage2023} as vision encoder, producing visual features for a SmolLM2 language decoder~\citep{allalSmolLM2WhenSmol2025}.
Further, SmolVLA adopts a smaller action expert consisting of \(\sim\)100M parameters and an interleaved stack of self and cross-attention layers.
To improve efficiency, the action expert adopts a reduced embedding dimension compared to the VLM backbone, resulting in \( d_{v_\theta} = 0.75 d_{\text{VLM}} \).
\citet{shukorSmolVLAVisionLanguageActionModel2025}'s design choices thus result in a much smaller size model compared to \pizero, consisting of ca. 450M parameters versus \pizero's 3.3B parameters.

In practice, SmolVLA consumes multi-view RGB images, a natural-language instruction, and projected sensorimotor state token as inputs, together with the noised \emph{action chunk} \( \tilde{a}_{t:t+H_a} \) the action expert \( v_\theta \) is trained to denoise.
The robot proprioperceptive states are projected to a shared token space with the VLM to match \( d_{\text{VLM}} \), and successively projected into the expert's token space.
Similarily to \pizero, SmolVLA adopts separate experts communicating exclusively through self-attention layers, which however do not employ blockwise causal attention masking and rather favour simple causal masking.

In contrast with \pizero, the action expert interleaves \emph{cross-attention} (CA) and \emph{self-attention} (SA) layers, a choice shown to yield higher success and smoother action chunks in practice.
While in the expert SA layers tokens are used to obtain queries, keys and values, CA layers use action tokens only as queries, and instead project visual, language and proprioperceptive tokens from the VLM backbone to a shared embedding space to then obtain keys and values.
Notably, keys and values can be cached  here as well, resulting in performance gains at inference time.

SmolVLA also trims down both token and layer compute.
First, it \emph{reduces visual tokens} via pixel shuffling to a fixed budget of 64 tokens per frame, foregoing the tiling used during VLM pretraining for the sake of runtime efficiency. 
Second, it \emph{skips upper VLM layers}, as only features from the first \(N\) decoder layers, with \(N=L/2\), are consumed, which provides a good speed-performance trade-off and effectively halves compute needs for the larger part of SmolVLA.
Beyond model compactness, SmolVLA also contributes an inference stack that decouples action prediction from execution for responsiveness on modest hardware (Section~\ref{sec:ch4-async-inference}).

Departing from reliance on proprietary datasets, SmolVLA pretrains exclusively on 450+ \emph{community datasets}, totaling 20k+ trajectories. 
Because instructions in community contributed dataset can be noisy or missing, the authors re-annotate tasks with a small off-the-shelf VLM using frames sampled from the dataset, and standardize camera viewpoints by mapping sources to a consistent top/wrist/side ordering.
At test time, similarily to \pizero, SmolVLA forward-integrates flow over 10 steps, resulting in fast inference.
SmolVLA proves effective across a range of both real-world and simulated environments, rivaling \pizero~while being close to 40\% faster and consuming 6x less memory~\citep{shukorSmolVLAVisionLanguageActionModel2025}.

\subsubsection{Code Example: Using SmolVLA}
\begin{pbox}[label={ex:using-smolvla}]{Using SmolVLA \\ \url{https://github.com/fracapuano/robot-learning-tutorial/blob/main/snippets/ch5/02_using_smolvla.py}}
\begin{lstlisting}[language=python]
import torch

from lerobot.cameras.opencv.configuration_opencv import OpenCVCameraConfig
from lerobot.datasets.utils import hw_to_dataset_features
from lerobot.policies.factory import make_pre_post_processors
from lerobot.policies.smolvla.modeling_smolvla import SmolVLAPolicy
from lerobot.policies.utils import build_inference_frame, make_robot_action
from lerobot.robots.so100_follower.config_so100_follower import SO100FollowerConfig
from lerobot.robots.so100_follower.so100_follower import SO100Follower

MAX_EPISODES = 5
MAX_STEPS_PER_EPISODE = 20

device = torch.device("mps")  # or "cuda" or "cpu"
model_id = "lerobot/smolvla_base"

model = SmolVLAPolicy.from_pretrained(model_id)

preprocess, postprocess = make_pre_post_processors(
    model.config,
    model_id,
    # This overrides allows to run on MPS, otherwise defaults to CUDA (if available)
    preprocessor_overrides={"device_processor": {"device": "mps"}},
)

# find ports using lerobot-find-port
follower_port = ...  # something like "/dev/tty.usbmodem58760431631"

# the robot ids are used the load the right calibration files
follower_id = ...  # something like "follower_so100"

# Robot and environment configuration
# Camera keys must match the name and resolutions of the ones used for training!
# You can check the camera keys expected by a model in the info.json card on the Hub
camera_config = {
    "camera1": OpenCVCameraConfig(index_or_path=0, width=640, height=480, fps=30),
    "camera2": OpenCVCameraConfig(index_or_path=1, width=640, height=480, fps=30),
}

robot_cfg = SO100FollowerConfig(port=follower_port, id=follower_id, cameras=camera_config)
robot = SO100Follower(robot_cfg)
robot.connect()

task = ...  # something like "pick the red block"
robot_type = ...  # something like "so100_follower" for multi-embodiment datasets

# This is used to match the raw observation keys to the keys expected by the policy
action_features = hw_to_dataset_features(robot.action_features, "action")
obs_features = hw_to_dataset_features(robot.observation_features, "observation")
dataset_features = {**action_features, **obs_features}

for _ in range(MAX_EPISODES):
    for _ in range(MAX_STEPS_PER_EPISODE):
        obs = robot.get_observation()
        obs_frame = build_inference_frame(
            obs, dataset_features, device, task=task, robot_type=robot_type
        )

        obs = preprocess(obs_frame)

        action = model.select_action(obs)
        action = postprocess(action)
        action = make_robot_action(action, dataset_features)
        robot.send_action(action)

    print("Episode finished! Starting new episode...")
\end{lstlisting}
\end{pbox}

\newpage
\section{Conclusions}
\label{sec:conclusions}

This tutorial has charted the paradigmatic shift transforming robotics, tracing the \highlight{evolution of robotics from structured, model-based methods to the dynamic, data-driven approaches that define modern robot learning}. We began by examining the limitations of traditional dynamics-based control, namely its brittleness and significant engineering overhead, which motivate the adoption of more flexible, learning-based alternatives. Unlike scalable, data-driven techniques, conventional explicit models demand extensive human expertise, hindering wider accessibility and scalability of robotics.

Our exploration traced a clear trajectory of progress, beginning with Reinforcement Learning (RL). While RL offers a powerful paradigm for learning through interaction, its application in robotics is complicated by challenges such as sample inefficiency, safety concerns in real-world training, and the complexities of reward design. We saw how modern approaches like HIL-SERL make real-world RL more feasible by incorporating training-time human guidance, datasets of previously collected data as well as learned reward classifiers.

Nonetheless, the inherent difficulties of RL increasingly motivate approaches based on imitation learning, capable to safely learns from limited numbers of real-world, reward-free expert demonstrations. In turn, the wider adoption of imitation learning led to the development of single-task policies, where advanced Behavioral Cloning techniques---implemented as state-conditioned generative models like Action Chunking with Transformers and Diffusion Policy---have demonstrated the ability to learn complex, multimodal behaviors from human demonstrations. These advancements laid the groundwork for the current frontier: generalist, language-conditioned Vision-Language-Action models capable to perform few- and zero-shot a variety of different real-world tasks. By leveraging powerful pre-trained backbones and sophisticated generative methods like flow matching, models such as \pizero~and SmolVLA represent a significant leap towards foundational models for robotics capable of generalizing across diverse tasks, and even robot embodiments.

A central theme of this work is the critical role of openness in accelerating this progress. The recent explosion in capability is inseparable from the advent of large-scale, openly available datasets, standardized, stable and accessible model architectures, and accessible, open-source software like \lerobot. We argue this convergence on open-source robotics is not a mere trend but a fundamental enabler, democratizing access to research and unlocking the potential of large, decentralized efforts to advance the field.

The journey detailed in this tutorial, from first principles to the state-of-the-art, aims to equip researchers and practitioners with the context and tools to begin their own explorations in open-source robot learning.

\bibliographystyle{plainnat}
\bibliography{main}

\end{document}